\documentclass[a4paper,10pt]{article}
\usepackage[margin=1in]{geometry}  % For 0.75-inch margins

\usepackage[english]{babel}

\usepackage{xurl}

\usepackage{xurl}
\usepackage[hidelinks]{hyperref}
\usepackage[utf8]{inputenc}
\usepackage[small]{caption}
\usepackage{graphicx}
\usepackage{amsmath}
\usepackage{amsthm}
\usepackage{booktabs}
\usepackage{algorithm}
\usepackage{algpseudocode}
\algrenewcommand\alglinenumber[1]{\tiny #1:}
\usepackage[switch]{lineno}

\usepackage{subfigure}

\usepackage{amsfonts,bm}
\usepackage{mathtools}
\usepackage{wrapfig}
\usepackage{soul}
\usepackage{multirow}
\usepackage[most]{tcolorbox}
\usepackage{xcolor,colortbl}
\definecolor{DPLightGreen}{rgb}{0.307, 0.610, 0.446}
\definecolor{DPLightBlue}{rgb}{0.541, 0.585, 0.951}
\definecolor{DPLightRed}{rgb}{0.941, 0.585, 0.551}
\definecolor{DPLightOrange}{RGB}{147, 140, 122}

\definecolor{RBRed}{rgb}{0.98,0.88,0.85}
\definecolor{RBRedLight}{rgb}{1,0.93,0.90}

\definecolor{RBBlue}{rgb}{0.81,0.80,0.94}
\definecolor{RBBlueLight}{rgb}{0.91,0.90,0.98}

\definecolor{RBBlueL}{rgb}{0.84,0.83,0.97}
\definecolor{RBBlueLL}{rgb}{0.86,0.85,0.99}

\definecolor{RBGreen}{rgb}{0.85,0.91,0.84}
\definecolor{RBGreenLight}{rgb}{0.93,0.98,0.92}

\definecolor{RBOrange}{RGB}{250,240,210}

\definecolor{DPBlueD}{rgb}{0.24,0.43,0.77}
\definecolor{DPPurple}{rgb}{0.46,0.12,0.45}

\definecolor{DPGreen}{rgb}{0,0.45,0.24}
\definecolor{DPLightGreen}{rgb}{0,0.65,0.44}

\newtcbox{\dpboxred}{on line,
  colframe=DPLightRed,colback=DPLightRed!10!white,
  boxrule=0.5pt,arc=2pt,boxsep=0pt,left=2pt,right=2pt,top=2pt,bottom=2pt}
  
\newtcbox{\dpboxorange}{on line,
  colframe=DPLightOrange,colback=RBOrange!10!white,
  boxrule=0.5pt,arc=2pt,boxsep=0pt,left=2pt,right=2pt,top=2pt,bottom=2pt}

\newtcbox{\dpboxgreen}{on line,
  colframe=DPLightGreen,colback=DPLightGreen!10!white,
  boxrule=0.5pt,arc=2pt,boxsep=0pt,left=2pt,right=2pt,top=2pt,bottom=2pt}

\newtcbox{\dpboxblue}{on line,
  colframe=DPLightBlue,colback=DPLightBlue!10!white,
  boxrule=0.5pt,arc=2pt,boxsep=0pt,left=2pt,right=2pt,top=2pt,bottom=2pt}

\definecolor{forestgreen}{rgb}{0.60, 0.27, 0.06}

\newcommand\crule[3][black]{\textcolor{#1}{\rule{#2}{#3}}}
\newcommand{\from}[3]{\noindent {\bf {\color{red}[{\sc From #1 to #2:} {\small #3}]}}}
\newcommand{\eg}{\text{e.g.}}

\newcommand{\CADD}{\textit{CADD}}

\newcommand\blfootnote[1]{%
  \begingroup
  \renewcommand\thefootnote{}\footnote{#1}%
  \addtocounter{footnote}{-1}%
  \endgroup
}

\usepackage{tikz}

\definecolor{primarybox}{RGB}{240,248,255}    % Light Azure
\definecolor{primaryline}{RGB}{70,130,180}    % Steel Blue
\definecolor{secondarybox}{RGB}{255,250,240}  % Cornsilk
\definecolor{secondaryline}{RGB}{210,105,30}  % Chocolate

\newenvironment{runningexample}[1][]{%
    \begin{tcolorbox}[
        colback=primarybox,
        colframe=primaryline,
        boxrule=0.5mm,
        sharp corners=all,
        before skip=10pt,
        after skip=10pt,
        left=4mm,
        right=4mm,
        top=2mm,
        bottom=2mm,
        breakable,
        title={Running Example: #1}
    ]%
}{%
    \end{tcolorbox}%
}

\def\scititle{
	Context and Transcripts Improve Detection of Deepfake Audios of Public Figures
}
\title{\scititle}

\author{
Chongyang Gao$^{1\dagger}$,
Marco Postiglione$^{1\dagger}$, 
Julian Baldwin$^{1}$, 
Natalia Denisenko$^1$, \\
Isabel Gortner$^1$,
Luke Fosdick$^1$, 
Chiara Pulice$^1$,
Sarit Kraus$^2$,
V. S. Subrahmanian$^{1 *}$\\
$^1$Northwestern University\\
$^2$Bar-Ilan University\\
}
\begin{document} 
% Insert the title and author list
\maketitle

\blfootnote{$\dagger$ Equal Contribution. $*$ Correspondence to: V. S. Subrahmanian (vss@northwestern.edu)}
% Abstract, in bold
% There are strict length limits, and not all formats have abstracts.
% Consult the journal instructions to authors for details.
% Do not cite any references in the abstract.
\begin{abstract} % \bfseries \boldmath
Humans use context to assess the veracity of information. However, current audio deepfake detectors only analyze the audio file without considering either context or transcripts. We create and analyze a Journalist-provided Deepfake Dataset (JDD) of 255 public deepfakes which were primarily contributed by over 70 journalists since early 2024. We also generate a synthetic audio dataset (SYN) of dead public figures and propose a novel Context-based Audio Deepfake Detector (CADD) architecture. In addition, we evaluate performance on two large-scale datasets: ITW and P$^2$V. We show that sufficient context and/or the transcript can significantly improve the efficacy of audio deepfake detectors. Performance (measured via F1 score, AUC, and EER) of multiple baseline audio deepfake detectors and traditional classifiers can be improved by 
5\%-37.58\% in F1-score, 3.77\%-42.79\% in AUC, and 6.17\%-47.83\% in EER. We additionally show that CADD, via its use of context and/or transcripts, is more robust to 5 adversarial evasion strategies, limiting performance degradation to an average of just -0.71\% across all experiments. Code, models, and datasets are available at our project page: \url{https://sites.northwestern.edu/nsail/cadd-context-based-audio-deepfake-detection/}(access restricted during review).
\end{abstract}

% The first paragraph of any Science paper does NOT have a heading
% Nor is it indented
\section{Introduction}
%why audio deepfakes are important
Audio deepfakes of public figures are proliferating at an alarming rate \cite{pei2024deepfake,dhs_deepfake_threats,byman2023deepfakes,linna2024deepfakes,dash2023chatgpt}. They are used to tarnish politicians, influence elections, and perpetrate financial scams. Infamous audio deepfakes include the phone call by former US President Joe Biden allegedly telling New Hampshire voters to stay away from the polls\footnote{\url{https://www.politifact.com/factchecks/2024/jan/22/robocaller/fake-joe-biden-robocall-in-new-hampshire-tells-dem/}},  Elon Musk allegedly promoting a cryptocurrency deal\footnote{\url{https://www.nytimes.com/interactive/2024/08/14/technology/elon-musk-ai-deepfake-scam.html}}, and Indian billionaire Mukesh Ambani allegedly promoting investments\footnote{\url{https://timesofindia.indiatimes.com/city/mumbai/retired-professional-falls-victim-to-deepfake-scam-featuring-mukesh-ambani/articleshow/111667464.cms}}. Deepfakes involving public figures pose a clear and present danger because of their potential to sway or deceive large populations. 

%why context can be helpful
These incidents have spurred interest in developing audio deepfake detection algorithms \cite{kawa2023improved,kawa2022specrnet,wu2018light,afchar2018mesonet,jung2022pushing}. Past work has focused on solely studying the audio files presented to such detectors \cite{DBLP:conf/icassp/WangHY0Z024,DBLP:conf/icassp/DengRZZS24,DBLP:conf/icassp/CuccovilloGA24}. However, in the case of audio samples of public figures, there is rich context that can be leveraged. We propose CADD (Context-based Audio Deepfake Detection), which automatically adds contextual information to an audio. This can include the person's demographics, occupation, past news and social media posts about the person, and more. The audio with rich context-enhanced can then be analyzed by existing audio deepfake detectors and/or classical machine-learning models. Context-enhanced audio deepfake detectors can potentially also help detect deepfake videos by analyzing the audio channel of a video to check if it is fake.

\begin{comment}
For example, a deepfake audio impersonating Indian billionaire Mukesh Ambani providing unsolicited investment advice\footnote{\url{https://www.asiafinancial.com/deepfake-of-asias-richest-man-used-in-india-stock-market-scam}} was released around June 24, 2024. The fabricated audio claims: \textit{``Do you really trust your own judgment? Do you really think you can make money with your stocks? Follow me and my student, Mr. Venit. We will provide you with free stock diagnosis and investment plan customization and recommend stocks worth buying in the future stock market. Time is limited. The number of people is limited. Paid resources are now shared for free. Everyone is welcome to join."}
This deepfake enabled a financial scam. In this example, the context automatically gathered by CADD included news articles and Reddit posts published before June 24 including one talking about past scams in which Mr. Ambani was impersonated via deepfakes\footnote{\href{https://www.dnaindia.com/mumbai/report-mukesh-ambani-deepfake-video-mumbai-doctor-falls-prey-to-frauds-loses-rs-7-lakh-in-3094299}{[Mukesh Ambani deepfake video: Mumbai doctor falls prey to frauds, loses Rs 7 lakh in... (DNA India, 2024-06-22)]}}. Wikidata about Mr. Ambani also formed part of the context gathered by CADD. The history of impersonation involving fake investments might be relevant to the assessment of the audio in question. Yet, no prior audio deepfake detector considers such context.
\end{comment}

%the data issue
Adding to the challenge of automatically deciding if an audio clip is real or fake is the fact that existing data sets (\eg ASVSpoof2021 \cite{yamagishi2021asvspoof}, ADD2023 \cite{yi2023add}, WaveFake~\cite{frank2021wavefake}) are limited. In most related work~\cite{yamagishi2021asvspoof,yi2023add,frank2021wavefake,zhang2022partialspoof,yi2021half,bird2023real,ma2024cfad,muller2024mlaad,xie2024codecfake}, researchers use existing audio deepfake creation tools such as text-to-speech \cite{DBLP:conf/nips/RenRTQZZL19}, voice cloning \cite{DBLP:conf/nips/KimSBSBDVYC23} and voice conversion \cite{DBLP:conf/icassp/RibeiroRCHGL22} to generate deepfake audio and then train machine learning models on these datasets. We call these ``synthetic'' datasets. Such datasets assume that adversaries will use generative models blindly to produce deepfakes. Following along these lines, we also create a synthetic dataset SYN using a process described in materials and methods \cite{methods}. SYN contains a total of 771 audios (468 deepfake audios of dead subjects together with 303 publicly available real audios), along with relevant context.

However, our experience in developing the Global Online Deepfake Detection System (GODDS)~\cite{postiglione2024godds} that currently serves a community of 70+ journalists, shows that real-world audio deepfakes may be more complex, as savvy adversaries try to evade detection. While they may use off-the-shelf tools to generate deepfakes, additional types of manipulation may be used~\cite{wu2024clad}. In contrast to these past efforts, we present a \emph{Journalist-provided Deepfake Dataset} (JDD), composed of publicly available audio samples of public figures that were primarily contributed by over 70 journalists. While several datasets attempt to capture real-world characteristics, they differ from JDD in important ways. The In-The-Wild (ITW) dataset~\cite{muller2022does} includes public figures but features highly unrealistic content. The creators of ITW themselves state that ``Since the speakers talk absurdly and out-of-character (`Donald Trump reads Star Wars'), it is easy to verify that the audio files are really spoofed''\cite{muller2022does}. Our experiments will show that this dataset is a very easy one for deepfake detectors and the deepfakes in the dataset may easily be determined as false by any journalist without the application of a deepfake detector. The P$^2$V dataset~\cite{gao2025perturbed}\footnote{\url{https://sites.northwestern.edu/nsail/p2v-perturbed-public-voices/}} takes a different approach by generating realistic deepfakes based on contextually-relevant LLM-generated transcripts, incorporating state-of-the-art audio generative techniques and adversarial manipulations. While P$^2$V represents a significant advancement in synthetic realism, it still relies on simulated scenarios rather than actual deepfakes encountered in journalistic practice. {Hence, JDD is fundamentally different from SYN, ITW, and P$^2$V, as it comprises actual deepfakes that journalists have encountered or identified in real-world scenarios, representing the true threat landscape faced by media professionals.}
% OLD TEXT:
%A related ``real world'' dataset called In-The-Wild (ITW) \cite{muller2022does} also includes public figures, but almost all the audios in ITW are highly unlikely to have occurred. In fact, the creators of ITW state that ``Since the speakers talk absurdly and out-of-character (`Donald Trump reads Star Wars'), it is easy to verify that the audio files are really spoofed''\cite{muller2022does}. Our experiments will show that this dataset is a very easy one for deepfake detectors and the deepfakes in the dataset may easily be determined as false by any journalist without the application of a deepfake detector. {Hence, JDD is fundamentally different from both SYN and ITW.}

Our comprehensive experiments on these diverse datasets demonstrate that CADD can significantly improve detection accuracy across 71 state-of-the-art baselines and traditional machine learning models, enabled by its ability to leverage rich contextual information. Moreover, CADD substantially enhances the robustness of these detectors against various audio manipulations that typically cause significant performance degradation in traditional approaches \cite{gao2025perturbed}.

\section{Results}

CADD analyzes an audio clip $a$ of public figure $f_a$ by analyzing features automatically extracted from context (recent news pieces and social media posts mentioning $f_a$ and Wikidata information), transcript, and the audio itself. We compared CADD against 71 baselines: 22 state-of-the-art (SOTA) audio deepfake detectors including RawNet3~\cite{jung2022pushing}, LCNN~\cite{wu2018light}, MesoNet~\cite{afchar2018mesonet}, and SpecRNet~\cite{kawa2022specrnet} with various feature extractors (LFCC, MFCC, Whisper~\cite{kawa2023improved}), and 49 traditional machine learning (TML) models. Three CADD variants were tested: CADD(T) using transcripts, CADD(C) using context, and CADD(T+C) combining both. More details on CADD and selected baselines are provided in materials and methods \cite{methods}. We evaluate performance using precision (P), recall (R), and F1-score for both the real class, fake class and weighted average across classes, along with the Area Under the ROC Curve (AUC) and Equal Error Rate (EER). To provide a concise yet comprehensive summary, we report an overall Avg score, defined as the mean of AUC, F1-score on the fake class, and 1-EER.

\subsection{Performance on the JDD Dataset}
\begin{figure}
    \centering
    \subfigure[CADD vs. State-of-the-art (SOTA) baselines]{
        \includegraphics[width=\linewidth]{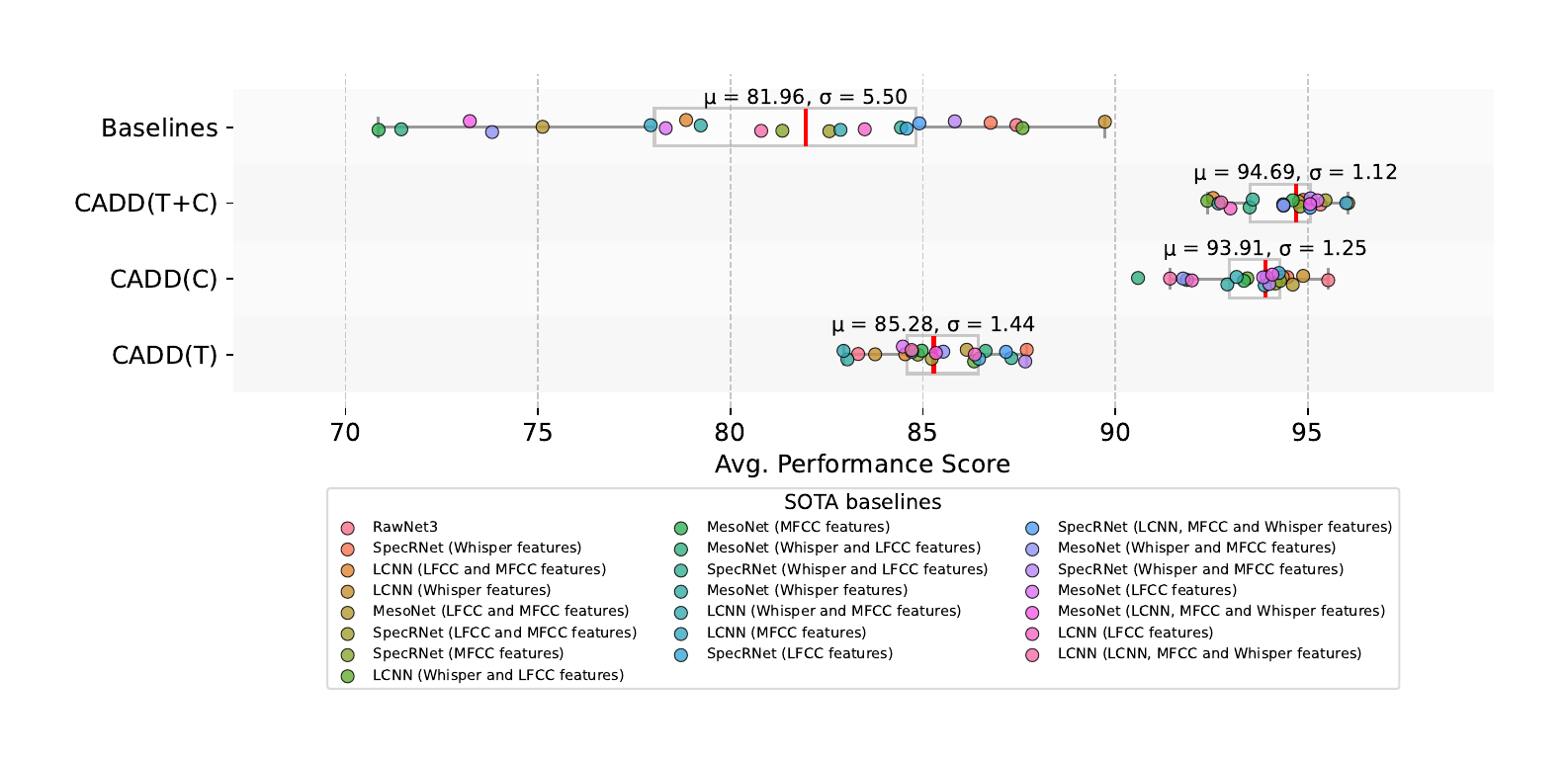}
        \label{fig:sota-baselines_JDD}
    }
    \subfigure[CADD vs. Traditional Machine Learning (TML) baselines]{
        \includegraphics[width=\linewidth]{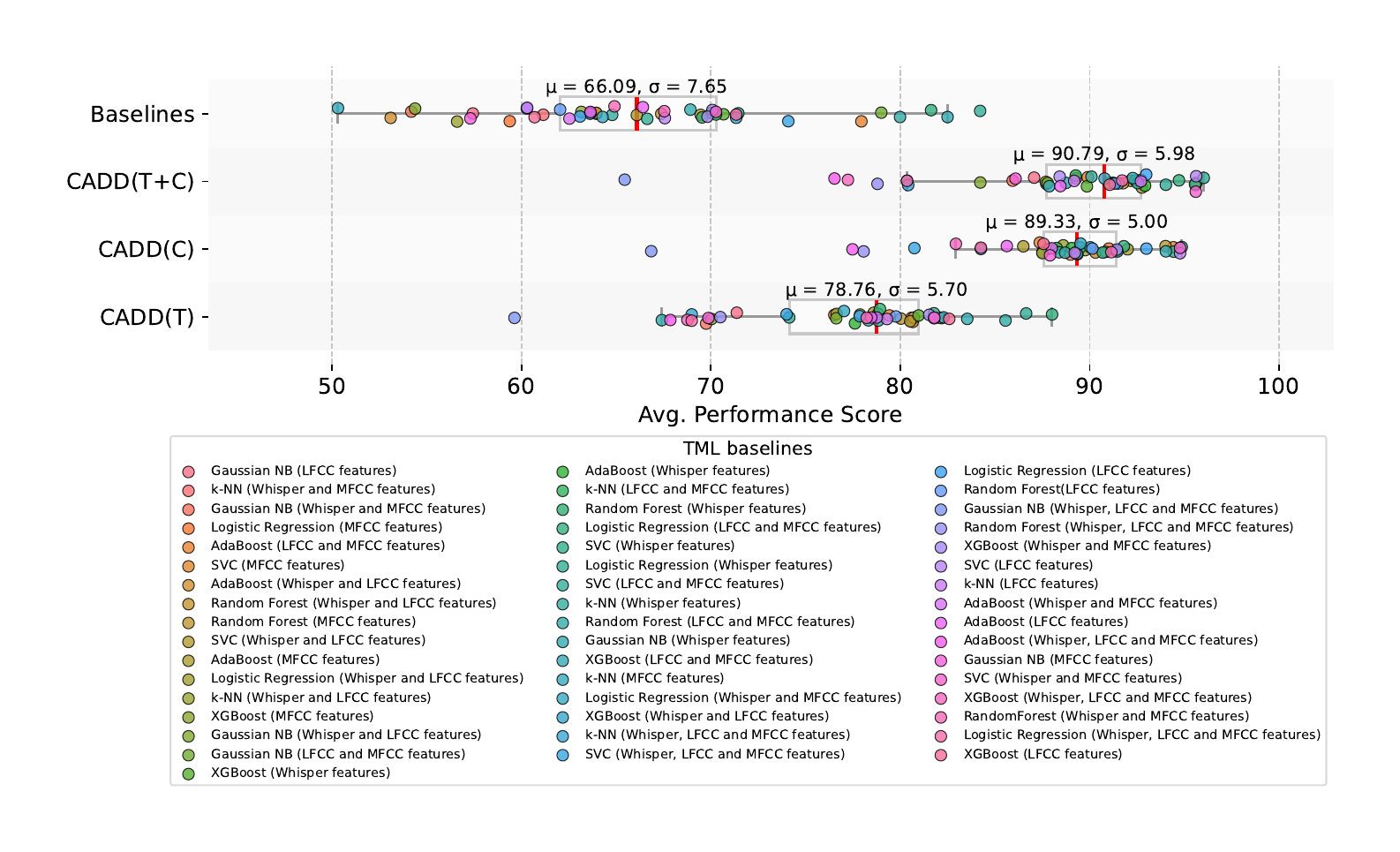}
        \label{fig:tml-baselines_JDD}
    }
    \caption{\textbf{Performance comparison of state-of-the-art (a) and traditional machine learning (b) baselines and our CADD configurations (CADD(T), CADD(C), and CADD(T+C)).} Each point represents a model's average performance score computed on our Journalist-provided Deepfake Dataset (JDD), with the same color denoting the same baseline across different configurations.}
    \label{fig:comparison}
\end{figure}

Figure~\ref{fig:comparison} shows performance of the baselines and CADD algorithms on the JDD dataset of all the detectors, split by SOTA baselines (in Figure~\ref{fig:sota-baselines_JDD}) and TML baselines in Figure~\ref{fig:tml-baselines_JDD}. Four ``rows'' of results are shown corresponding to the baseline algorithms, our CADD(T+C), CADD(C), and CADD(T). The x-axis shows the Avg score. We see that CADD(T+C) consistently delivers the best results w.r.t. the SOTA baselines. In the case of TML baselines, CADD(C) and CADD(T+C) jointly deliver the best results. Results focusing on the individual metrics (F1-score, AUC, and EER) are reported in figs. S1-S3.

Table~\ref{tab:main_rw} %Table 1 of main.tex
shows a detailed performance breakdown (precision, recall, F1 score, AUC, EER) for the top 4 state-of-the-art (SOTA) models and the top 4 traditional machine learning (TML) baselines out of the 71 evaluated, along with the 3 CADD algorithm configurations applied to JDD. These results illustrate the effectiveness of our approach. For consistency, we apply the same model selection across all other benchmarks. Complete results for all evaluated models are provided in the Appendix (tables S1-S35). % For example, RawNet3 \cite{jung2022pushing} is a well-known audio deepfake detection architecture. The next 3 rows show results when CADD(T), CADD(C), and CADD(T+C) are built on top of RawNet3. The last three columns show the average metric (Avg), the improvement in Avg over the baseline, and the p-value indicating statistical significance of the improvement.

To establish statistical significance in our performance comparisons, we conducted hypothesis testing on the absolute error distributions of each model. For each test sample, we computed the absolute error ($|y-\hat{y}|$; $y$ and $\hat{y}$ denote the ground truth and the prediction, respectively) for both the baseline detector and its CADD variant, yielding two error distributions per comparison. We then applied a Mann-Whitney U test \cite{mann1947test} to determine if the CADD variant produced significantly lower error rates than the baseline. All significance claims (denoted by $P<0.05$) indicate that the CADD variant demonstrated statistically lower error distributions compared to its corresponding baseline. To account for multiple hypothesis testing, we applied the False Discovery Rate (FDR) correction \cite{benjamini1995controlling}. 

Across the entire set of 22 SOTA detectors (see tables S1-S4), CADD(T+C) yields the best results for 18 out of 22 SOTA detectors, with 15 of these improvements demonstrating statistical significance ($P<0.05$). CADD(C) yields the best result in the other 4 cases, with 3 out of 4 showing statistical significance ($P<0.05$). Across all 22 cases, CADD(T+C) leads to an average increase of 13.9\% in the Avg score compared to the baseline.

Similarly, when considering the full set of experiments involving TML models (see tables S5-S11),  CADD(T+C) exhibits the best performance in 31 out of 49 baselines, with 27 of these improvements showing statistical significance ($P<0.05$). In the other 18 cases, using CADD(C) yields optimal results, with 11 demonstrating statistical significance ($P<0.05$). Across all 49 cases, CADD(T+C) leads to an average increase of 22.57\% in the average score compared to the corresponding baseline, representing a substantial and significant improvement. 

\subsection{Performance on the SYN dataset}
We ran the same experiments using the SYN dataset. As with the previous JDD analysis, we present a subset of the results in Table~\ref{tab:main_syn} (the overall summary can be found in figs. S4-S7 and the full results are in tables S13-S23). Here too, CADD improves most of the performance metrics of the majority of feature/classifier combinations. The only exception is RawNet3 when considering the AUC and EER metrics. Even in the case of RawNet3, CADD(T+C) has a higher F1 score on both the Real and Fake classes and its average score is also slightly higher. 

Considering the full set of experiment on the SYN dataset (see tables S13-S23), on average across the 4 SOTA detectors and 7 feature type combinations, CADD(T+C) exhibits a 3.99\% increase in Avg scores, with improvements being statistically significant ($P<0.05$) in 17 out of 21 cases. However, when we consider the TML baselines and the associated feature types, we find an increase of 9.5\% using CADD(T+C) on SYN compared to the machine learning classifier without context and transcript, with improvements being statistically significant ($P<0.05$) in 33 out of 49 cases. 

\subsection{Performance on the ITW dataset}
We also conducted the same experiments on the ITW dataset \cite{muller2022does}, simulating contextual information as ITW does not contain publication dates. The results are presented in detail in figs. S8-S11 and tables S24-S34. We found that SOTA detectors achieve near-perfect performance (i.e., Avg score is higher than 0.99 in most cases) but even here, CADD (with simulated context) always improved performance. On average, CADD(T+C) exhibits a 0.89\% increase across the 22 SOTA baselines and a 1.48\% increase across the 49 TML detectors in terms of Avg scores, with improvements being statistically significant ($P<0.05$) in 20 out of 21 cases and 49 out of 49 cases for SOTA and TML baselines, respectively. The smaller performance improvement is attributed to the fact that SOTA baselines already achieve an average score above 0.99 in most cases.

\subsection{Performance on the P$^{2}$V dataset}
We evaluate our method on the P$^{2}$V dataset~\cite{gao2025perturbed}, an IRB-approved benchmark designed to reflect realistic malicious deepfake scenarios by incorporating identity-aligned transcripts generated by LLMs and deepfakes generated by state-of-the-art voice cloning pipelines. Using the provided identity labels, we collect contextual information from recent news articles, social media posts, and Wikidata entries, and remove any samples for which no contextual data can be retrieved. P$^{2}$V contains more than 250,000 samples, which is over eight times larger than the widely used ITW dataset, thereby enabling robust large-scale training and evaluation. Due to the large dataset size and the resulting computation cost, we conduct experiments using the best-performing Whisper features identified from prior evaluations on other datasets. As shown in Table~3, CADD consistently achieves the best performance among all evaluated SOTA baselines. In particular, relative to SpecRNet, CADD with both contextual knowledge and transcript information yields a 6.15 point improvement in F1-score and a 52.26\% reduction in EER, highlighting its superior capability. All observed improvements are statistically significant ($p < 0.001$), further demonstrating the effectiveness of incorporating contextual awareness for deepfake detection. {Among the TML baselines, only XGBoost achieves statistically significant improvements when incorporating contextual information ($p<0.05$ for CADD(C) and $p < 0.001$ for CADD(T+C)), while GaussianNB, SVC, RandomForest and other TML baselines in table \ref{tab:p2v_tml} show no significant gains after FDR correction. This likely stems from P$^{2}$V's inherent class imbalance, which challenges simpler algorithms lacking sophisticated mechanisms for skewed distributions. This underscores a key advantage of CADD's \textit{Neural Processing} module: its deep learning architecture handles class imbalance more effectively through learned representations and adaptive weighting, enabling it to extract nuanced patterns from contextual features where simpler TML baselines cannot.}

\subsection{Performance Summary}
To conclude, first, CADD improves performance of SOTA and TML detectors/features in all cases. Second, SOTA and TML baselines performed very poorly on JDD data,with the highest Avg score reaching only 87.43, compared to 96.05 achieved by CADD. This may be due to the fact that the JDD dataset was contributed by journalists who, perhaps, found assessing the veracity of these audios particularly challenging. Third,  CADD improves 22 SOTA baselines' Avg score by 5.84 to 23.75\% and improves 49 TML baselines' Avg scores by 5.09-39.02\% on the JDD dataset. 

The performance improvements are consistent but vary across datasets: most substantial on JDD (13.9\% average improvement for SOTA baselines and 22.57\% for TML baselines), moderate on SYN (3.99\% for SOTA and 9.5\% for TML baselines), and smallest yet still statistically significant on ITW (0.89\% for SOTA and 1.48\% for TML baselines) where SOTA baselines already achieved high performance. 
% (0.74/95.12 + 1.27/96.72 + 5.46/91.03 + 1.68/92.68)/3
On P$^{2}$V, CADD achieves a 3.3\% average improvement over state-of-the-art baselines, demonstrating its superior performance when enriched with contextual information.
This pattern correlates with dataset complexity and the sufficiency of the context---CADD delivers the greatest performance gains on challenging real-world examples (JDD), moderate gains on synthetic data (SYN and P$^{2}$V), and incremental improvements even on datasets where baselines already achieve near-perfect performance (ITW).

\subsection{Robustness of CADD Under Evasive Audio Manipulations} 
Audio deepfake detectors are often unreliable when the input audio is subject to manipulation, even with simple alterations such as time stretching \cite{DBLP:journals/corr/abs-2404-15854,DBLP:conf/mm/WangJHGXML20}. As the 22 SOTA baselines/feature combinations outperform the TML baselines, we compared their robustness with that of CADD under 5 adversarial manipulations to the audio: air absorption, which simulates the natural attenuation of sound as it travels through air (including frequency-dependent attenuation effects), background noise injection, Gaussian noise injection, compression via MP3, and stretching the playout time of the audio \cite{DBLP:journals/corr/abs-2404-15854}. For each type of manipulation, we tested multiple intensity levels, resulting in a total of 23 distinct manipulations. Further details are provided in the supplementary text.  We also release perturbed versions of the three datasets used in our experiments(JDD, SYN, and ITW) to support further research on robustness under real-world audio degradations.

\begin{figure}
    \centering
    \includegraphics[width=\textwidth]{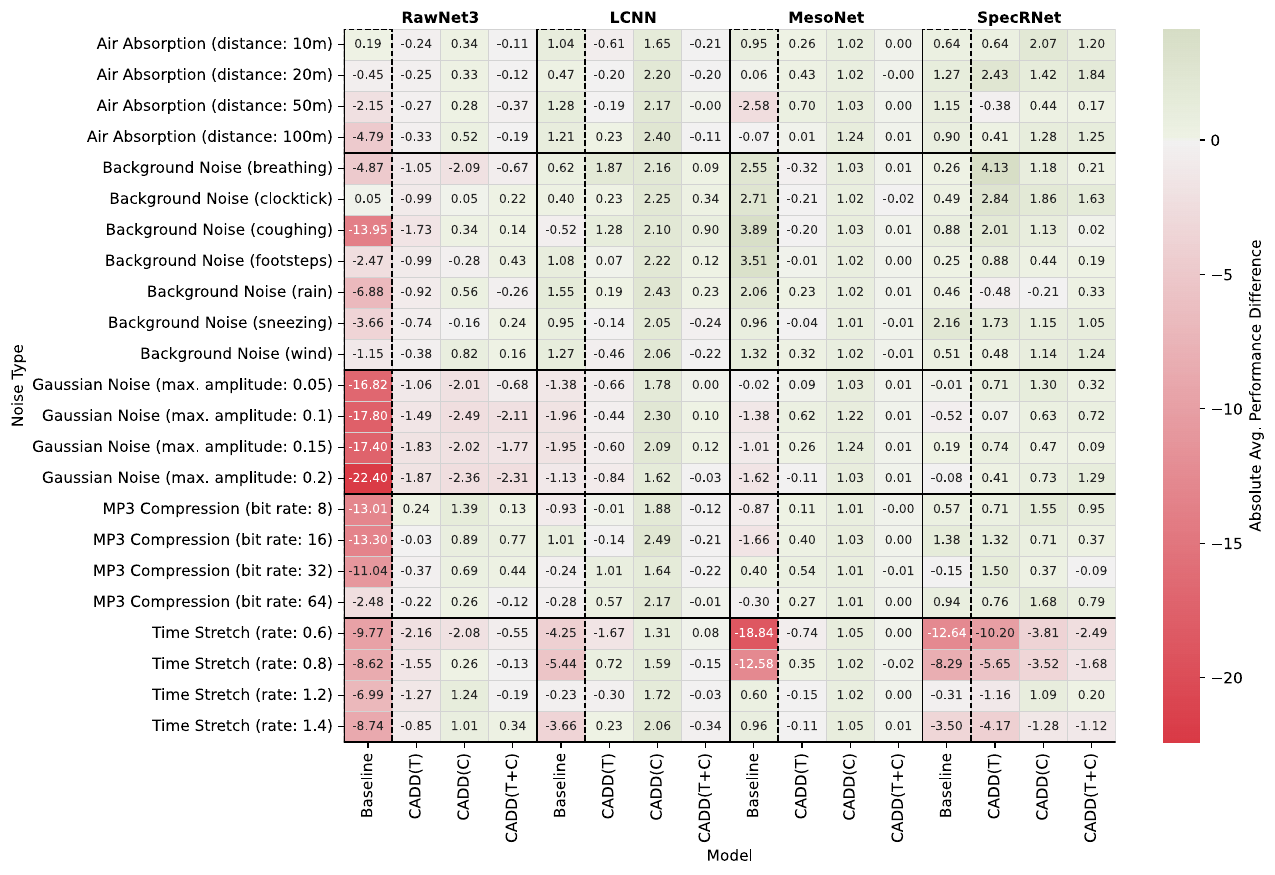}
    \caption{\textbf{Performance Robustness Under Audio Manipulations (JDD dataset)} The figure shows the absolute percentage difference between the results (in terms of Avg scores) obtained on the perturbed JDD test set and those obtained on the original, unperturbed test set for the subset of 4 best-performing SOTA baselines and related CADD configurations. SOTA models are separated by black vertical lines, with baseline results highlighted in a dashed rectangle.}    \label{fig:robustness_experiments_ALL}
\end{figure}

% \paragraph{Robustness on JDD}
Figure~\ref{fig:robustness_experiments_ALL} presents a subset of results including the 4 best SOTA baselines: RawNet3, LCNN with Whisper features, MesoNet with Whisper features, and SpecRNet with Whisper features. The table rows correspond to the 5 types of attacks and their respective intensity levels. To provide a concise yet comprehensive summary, we report an overall Avg score, defined as the mean of AUC, F1-score on the fake class, and 1-EER. The complete results are shown in the figures S12-S14. On JDD, baselines showed an average performance degradation of 8.2\% in terms of Avg score (worst case: -22.4\% under Gaussian noise), while CADD(T+C) maintained an average degradation of only -0.71\%. Given the similarity between P$^{2}$V and SYN and the substantial computational cost of large-scale training, we conduct synthetic-data experiments only on SYN. Similar patterns emerged on SYN, where CADD reduced average degradation from -5.74\% to -0.91\%. On ITW, CADD showed less improvement, likely because the dataset's simplicity allowed models to achieve near-perfect results by focusing on audio features alone, and the simulated context proved less effective than real-world contextual information.

\subsection{Error Rates in Detecting Deepfakes about Political and Entertainment Figures}

We compared deepfake detection performance between entertainment and political figures in our JDD dataset by analyzing absolute detection errors $|\hat{y}-y|$, $y$ and $\hat{y}$ denoting ground truth and model's prediction, respectively. Using Mann-Whitney U tests with Benjamini-Hochberg correction \cite{benjamini1995controlling}, we examined error distributions across our CADD configurations with sample sizes distributed as follows: entertainment category ($n_E = 71$) and non-entertainment category ($n_{\bar{E}} = 101$), political category ($n_P = 68$) and non-political category ($n_{\bar{P}} = 104$).

Results in Table S35 show that CADD models using contextual information (CADD(C) and CADD(T+C)) exhibit significant differences in error distribution across occupational categories ($P < 0.001$). For entertainment figures, CADD models achieved lower (i.e. better) absolute errors (0.0007--0.1691) compared to other figures (0.0025--0.3370). For political figures, error rates were significantly higher (0.0036--0.4836) compared to other figures (0.0018--0.4092). This pattern was consistent across architectures, though least pronounced with LCNN, where CADD(C) and CADD(T+C) showed statistical significance for entertainment ($P < 0.01$) but not for political figures.

The linguistic analysis presented in table S36 revealed political content featured higher linguistic complexity than entertainment content when measured by Flesch Reading Ease Scores ($72.83$ vs. $79.29$; $P < 0.01$) and longer words (4.38 vs. 4.05 characters; $P < 0.01$). Topic diversity, calculated using Latent Dirichlet allocation (LDA) \cite{blei2003latent}, was approximately 45\% higher in political transcripts (0.84 vs. 0.58; $P < 0.001$). News context followed similar patterns, while Reddit discussions showed reversed topic diversity, with entertainment discussions being more diverse than political ones (0.94 vs. 0.80; $P < 0.001$).

These findings suggest two explanations: (1) higher linguistic complexity in political discourse creates a more challenging detection environment, and (2) deepfake creation patterns differ between domains. Based on our experience on our GODDS platform \cite{postiglione2024godds}, entertainment deepfakes typically follow consistent content patterns (often for scams or promotions), yielding more recognizable detection signatures. Political deepfakes demonstrate greater technical and contextual variety, serving purposes from satire to disinformation and presenting more complex detection challenges.

\section{Method}

\begin{figure}
% \vskip -0.2in
\begin{center}
\centerline{\includegraphics[width=\textwidth]{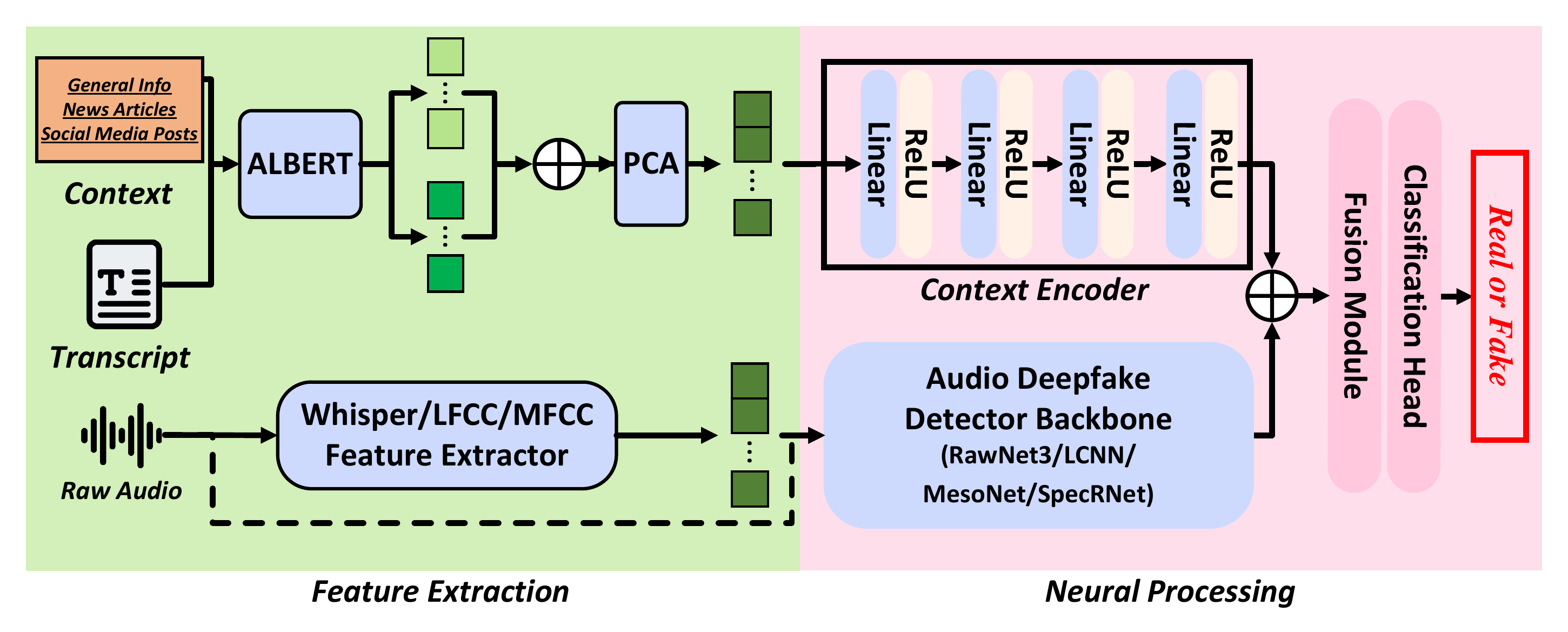}}
\caption{The architecture of our \CADD\ context fusion module has three components: 1) the context encoder, 2) the deepfake audio detection backbone, and 3) the fusion module and classification head.}
\label{fig:network}
\end{center}
% \vskip -0.3in
\end{figure}

Figure~\ref{fig:network} shows the CADD architecture, which includes two main components: Feature Extraction and Neural Processing.

Feature processing is different for audio vs. context/transcript. When an audio file is presented to CADD, we extract the transcript automatically. We also identify prior news articles and social media posts, as well as Wikidata information about the speaker. Embeddings are extracted from these textual artifacts using ALBERT \cite{DBLP:conf/iclr/LanCGGSS20}. Our experiments consider context embeddings alone (CADD(C)), the transcript embeddings alone (CADD(T)) or both (CADD(T+C)). A PCA module subsequently reduces these embeddings to a 100-dimensional vector. On the audio clip side, we extract features using well-known techniques such as linear frequency cepstral coefficients (LFCCs) \cite{zhang2022robust}, mel-frequency cepstral coefficients (MFCCs) \cite{gupta2013feature} and the SOTA Whisper framework \cite{radford2023robust}. These feature vectors may be of varying length depending on the feature extraction framework used.

Neural processing involves three parts (right half of Figure~\ref{fig:network}). A context encoder takes the 100-dimensional vector produced by PCA and generates another embedding. The audio feature vector is run through an audio deepfake detection system --- but with one difference. The last layer of such a system is typically a fully connected layer. The audio deepfake detector backbone does not include this last layer. The output generated by the other layers is then concatenated with the embedding generated by the context encoder and is run through a fusion module and classification head that produces the final output.

As a separate baseline not shown in Figure~\ref{fig:network}, we also concatenated the embeddings produced by PCA and by the audio feature extractor and learned traditional machine learning predictive models. 

“Materials and methods” in the Supplementary Materials provide additional details on the CADD architecture.

\section{Conclusion}
All past work on detecting audio deepfakes portraying public figures focus on the audio clips alone. 
The only publicly available audio deepfake dataset of public figures, the InTheWild (ITW) dataset, includes a note from its authors stating that ``Since the speakers talk absurdly and out-of-character (`Donald Trump reads Star Wars'), it is easy to verify that the audio files are really spoofed'' \cite{muller2022does}, which raises concerns about the realism of the dataset. Other audio deepfake datasets\cite{yamagishi2021asvspoof,yi2023add,frank2021wavefake} take real audio of non-public figures and apply off-the-shelf audio deepfake generation tools. Hence, these past datasets are flawed. We see these flaws come dramatically to light when we tested deepfake detectors that performed well on these flawed datasets on our Journalist-provided Deepfake Dataset (JDD) which was collected with inputs from over 70 journalists. 

In the case of public figures, our results show that using both the context and the transcript of an audio can play a significant role. Our CADD architecture is specifically designed to build on top of existing deepfake detectors and additionally leverage context and transcript features. On the real-world JDD dataset, our best performing CADD system beats baselines by 5.09--39.02\% in Avg score. On the SYN dataset, our best performing CADD system beats baselines by 0.06--34.83\% in Avg score. On the flawed ITW dataset, where all algorithms achieve near-perfect performance due to its limitations, CADD still beats state-of-the-art results. On P$^{2}$V, a large-scale synthetic benchmark, CADD achieves a 3.3\% average improvement, closely matching the gains observed on SYN. These results demonstrate that the proposed model operates reliably across both small and large datasets when provided with sufficiently rich contextual information.

Furthermore, CADD is, on average, much more robust than past baselines. When tested against 23 well-known audio manipulations on the JDD dataset, baselines experience an average performance degradation of -3.67\%, with a maximum drop of -22.4\%. In contrast, CADD reduces the average degradation to just -0.71\%, with the worst-case drop limited to -7.94\%.

Simply put, the conclusion is inescapable: at least in the case of public figures where context can be easily extracted from open sources such as news and social platforms (\eg, Reddit), the use of context to identify deepfake audios yields compelling improvements over analyzing the audio clip alone.

Future work might focus on the use of context and transcripts in the case of videos of public figures, and context alone in the case of images of public figures.

% If your text is very short you might need to uncomment the following line to avoid
% layout problems with the figures and tables.
%\newpage

% \clearpage % Clear all remaining figures and tables then start a new page

% The list of references goes after the main text and before the acknowledgements
% When preparing an initial submission, we recommend you use BibTeX, like this:
%
\bibliographystyle{unsrt}
\bibliography{bib} % for a file named science_template.bib

%%%%%%%%%%%%%%%% ACKNOWLEDGEMENTS %%%%%%%%%%%%%%%

\noindent\textbf{IRB Approvals.} Out of an abundance of caution, we sought IRB approvals from both institutions that the authors belong to. The Northwestern University IRB “determined that the proposed activity is not research involving human subjects.” (letter dated Jan 6 2025, study number STU00223255). The Bar-Ilan University IRB approved our research on Dec 31 2024 (approval number 311224407).

\section*{Acknowledgments}
We are grateful to Microsoft for supporting this research through free computing credits on the Azure computer cluster.

\paragraph*{Funding:}
N/A

%For example: F.~A. was funded by the Generous Science Agency grant~2372.
\paragraph*{Author contributions:}
CG: Conceptualization, Methodology, Software, Validation, Formal analysis, Investigation, Data Curation, Writing - Original Draft, Writing - Review \& Editing, Visualization;
MP: Conceptualization, Methodology, Software, Validation, Formal analysis, Investigation, Data Curation, Writing - Original Draft, Writing - Review \& Editing, Visualization; 
JB: Software, Investigation;
ND: Software; 
IG: Data Curation; 
LF: Data Curation; 
CP: Project administration; 
SK: Conceptualization, Writing - Original Draft, Writing - Review \& Editing, Supervision;
VSS: Conceptualization, Methodology, Writing - Original Draft, Writing - Review \& Editing, Supervision, Project administration.

\paragraph*{Competing interests:}
There are no competing interests to declare.

\paragraph*{Data and materials availability:}
We publicly release our models \cite{models} and data \cite{data}.

\newpage
%%%%%%%%%%%%%%%% MAIN TEXT FIGURES %%%%%%%%%%%%%%%

%%%%%%%%%%%%%%%% MAIN TEXT TABLES %%%%%%%%%%%%%%%
\clearpage

\begin{table}
\centering

\caption{\textbf{Performance comparison of 4 state-of-the-art (SOTA) models and 4 traditional machine learning (TML) baselines for deepfake detection on JDD data, with varying contextual information used by CADD.}
We evaluate CADD models trained under three conditions: with transcript-only (CADD(T)), with context-only (CADD(C)), and with both transcript and context knowledge (CADD(T+C)). The best performance values for each metric are highlighted in bold and marked with different background colors (\crule[RBRed]{0.2cm}{0.2cm} \crule[RBGreen]{0.2cm}{0.2cm} \crule[RBBlue]{0.2cm}{0.2cm}), while the second-best performance is underlined. The last column shows the FDR-corrected P-values (Benjamini-Hochberg procedure) from statistical significance testing (indicated by * for $P<0.05$, ** for $P<0.01$ and *** for $P<0.001$). See tables S1-S12 for full results.}
\label{tab:main_rw}
\scalebox{0.64}{
\small
\renewcommand{\arraystretch}{0.8} % Values less than 1 reduce spacing
% Updated column definition: removed 2 'c' columns
\begin{tabular}{@{}l|l|ccccccccccc|c@{}}
\toprule
 & & 
  \multicolumn{3}{c}{\cellcolor{RBGreen}Real} &
  \multicolumn{3}{c}{\cellcolor{RBRed}Fake} &
  \multicolumn{5}{c}{\cellcolor{RBBlue}Overall} & \\ \cmidrule(l){3-14} 
\textbf{Base Model} & \textbf{CADD Enhanced Model} & 
  P $(\uparrow)$ &
  R $(\uparrow)$ &
  F1 $(\uparrow)$ &
  P $(\uparrow)$ &
  R $(\uparrow)$ &
  F1 $(\uparrow)$ &
  P $(\uparrow)$ &
  R $(\uparrow)$ &
  F1 $(\uparrow)$ &
  AUC $(\uparrow)$ &
  EER  $(\downarrow)$ & 
  P-Value \\ \midrule

RawNet3 
& Baseline & 80.52 & 89.13 & 84.49 & 90.06 & 81.48 & 85.44 & 85.67 & 85.00 & 85.00 & 92.91 & 16.05 & --- \\

& Ours CADD(T) & 74.99 & 81.88 & 78.23 & 83.22 & 76.54 & 79.68 & 79.43 & 79.00 & 79.02 & 88.18 & 17.90 & $0.375$ \\

& Ours CADD(C) & \cellcolor{RBGreenLight}\textbf{90.58} & \underline{94.20} & \cellcolor{RBGreenLight}\textbf{92.27} & 94.88 & \cellcolor{RBRedLight}\textbf{91.36} & \cellcolor{RBRedLight}\textbf{93.02} & \cellcolor{RBBlueLight}\textbf{92.90} & \cellcolor{RBBlueLight}\textbf{92.67} & \cellcolor{RBBlueLight}\textbf{92.67} & \underline{98.51} & \cellcolor{RBBlueLight}\textbf{4.94}  & $0.001$** \\

& Ours CADD(T+C) & \underline{88.63} & \cellcolor{RBGreenLight}\textbf{95.65} & \underline{91.99} & \cellcolor{RBRedLight}\textbf{96.04} & \underline{89.51} & \underline{92.64} & \underline{92.63} & \underline{92.33} & \underline{92.34} & \cellcolor{RBBlueLight}\textbf{98.91} & \underline{5.56}  & $0.101$ \\

\midrule 
LCNN
& Baseline & 87.29 & 79.71 & 83.23 & 84.09 & 90.12 & 86.95 & 85.56 & 85.33 & 85.24 & 95.21 & 12.96 & --- \\
 & Ours CADD(T) & 74.03 & 84.06 & 78.63 & 84.77 & 74.69 & 79.30 & 79.83 & 79.00 & 78.99 & 89.88 & 17.90 & $10^{-12}$*** \\
 & Ours CADD(C) & \cellcolor{RBGreenLight}\textbf{94.65} & \underline{89.13} & \underline{91.77} & \underline{91.23} & \cellcolor{RBRedLight}\textbf{95.68} & \underline{93.38} & \underline{92.80} & \underline{92.67} & \underline{92.64} & \underline{98.05} & \underline{6.79}  & $10^{-9}$*** \\
 & Ours CADD(T+C) & \underline{92.81} &  \cellcolor{RBGreenLight}\textbf{93.48} &  \cellcolor{RBGreenLight}\textbf{93.13} & \cellcolor{RBRedLight}\textbf{94.43} & \underline{93.83} & \cellcolor{RBRedLight}\textbf{94.12} & \cellcolor{RBBlueLight}\textbf{93.69} & \cellcolor{RBBlueLight}\textbf{93.67} & \cellcolor{RBBlueLight}\textbf{93.67} & \cellcolor{RBBlueLight}\textbf{98.98} & \cellcolor{RBBlueLight}\textbf{4.94}  & $10^{-5}$*** \\ \midrule

MesoNet 
& Baseline &  68.23 & 88.41 & 76.83 & 86.81 & 64.20 & 73.46 & 78.26 & 75.33 & 75.01 & 85.84 & 21.60 & --- \\
& Ours CADD(T) &  78.95 & 77.54 & 77.07 & 81.49 & 79.63 & 79.52 & 80.32 & 78.67 & 78.39 & 88.12 & 18.52 & $0.002$** \\
& Ours CADD(C) & \underline{89.32} & \underline{92.03} & \underline{90.49} & \underline{93.06} & \underline{90.12} & \underline{91.43} & \underline{91.34} & \underline{91.00} & \underline{91.00} & \underline{96.55} & \underline{9.26}  & $10^{-14}$*** \\
& Ours CADD(T+C) &  \cellcolor{RBGreenLight}\textbf{90.97} &  \cellcolor{RBGreenLight}\textbf{92.75} &  \cellcolor{RBGreenLight}\textbf{91.76} & \cellcolor{RBRedLight}\textbf{93.81} & \cellcolor{RBRedLight}\textbf{91.98} & \cellcolor{RBRedLight}\textbf{92.82} & \cellcolor{RBBlueLight}\textbf{92.50} & \cellcolor{RBBlueLight}\textbf{92.33} & \cellcolor{RBBlueLight}\textbf{92.33} & \cellcolor{RBBlueLight}\textbf{98.32} & \cellcolor{RBBlueLight}\textbf{8.02}  & $10^{-14}$*** \\ \midrule

SpecRNet 
& Baseline &  82.37 & 79.71 & 80.51 & 84.19 & 85.19 & 84.27 & 83.35 & 82.67 & 82.54 & 92.07 & 16.05 & --- \\
& Ours CADD(T) &  82.83 & 86.96 & 84.83 & 88.37 & 84.57 & 86.41 & 85.82 & 85.67 & 85.68 & 92.12 & 15.43 & $10^{-6}$*** \\
& Ours CADD(C) & \underline{89.05} &  \cellcolor{RBGreenLight}\textbf{94.20} & \underline{91.52} & \cellcolor{RBRedLight}\textbf{94.93} & \underline{90.12} & \underline{92.43} & \underline{92.22} & \underline{92.00} & \underline{92.01} & \underline{97.79} & \cellcolor{RBBlueLight}\textbf{6.79}  & $10^{-13}$*** \\
& Ours CADD(T+C) &  \cellcolor{RBGreenLight}\textbf{90.95} &  \cellcolor{RBGreenLight}\textbf{94.20} &  \cellcolor{RBGreenLight}\textbf{92.53} & \underline{94.92} & \cellcolor{RBRedLight}\textbf{91.98} & \cellcolor{RBRedLight}\textbf{93.41} & \cellcolor{RBBlueLight}\textbf{93.09} & \cellcolor{RBBlueLight}\textbf{93.00} & \cellcolor{RBBlueLight}\textbf{93.01} & \cellcolor{RBBlueLight}\textbf{98.03} & \cellcolor{RBBlueLight}\textbf{6.79}  & $10^{-13}$*** \\ \midrule

\midrule

LR & Baseline & 78.26 & 78.26 & 78.26 & 78.26 & 78.26 & 78.26 & 78.26 & 78.26 & 78.26 & 88.37 & 21.74 & --- \\
 & Ours CADD(T) & 85.11 & 86.96 & 86.02 & 86.67 & 84.78 & 85.71 & 85.89 & 85.87 & 85.87 & 91.40 & 13.04 & $10^{-5}$***\\
 & Ours CADD(C) & \underline{87.76} & \underline{93.48} & \underline{90.53} & \underline{93.02} & \underline{86.96} & \underline{89.89} & \underline{90.39} & \underline{90.22} & \underline{90.21} & \underline{96.41} & \underline{10.87} & $10^{-10}$***\\
 & Ours CADD(T+C) & \cellcolor{RBGreenLight}\textbf{89.80} & \cellcolor{RBGreenLight}\textbf{95.65} & \cellcolor{RBGreenLight}\textbf{92.63} & \cellcolor{RBRedLight}\textbf{95.35} & \cellcolor{RBRedLight}\textbf{89.13} & \cellcolor{RBRedLight}\textbf{92.13} & \cellcolor{RBBlueLight}\textbf{92.57} & \cellcolor{RBBlueLight}\textbf{92.39} & \cellcolor{RBBlueLight}\textbf{92.38} & \cellcolor{RBBlueLight}\textbf{98.53} & \cellcolor{RBBlueLight}\textbf{6.52} & $10^{-10}$***\\ \midrule

SVC 
& Baseline & 80.00 & 86.96 & 83.33 & 85.71 & 78.26 & 81.82 & 82.86 & 82.61 & 82.58 & 88.23 & 17.39 & --- \\
 & Ours CADD(T) & 83.33 & 86.96 & 85.11 & 86.36 & 82.61 & 84.44 & 84.85 & 84.78 & 84.78 & 92.91 & 17.39 & $10^{-6}$***\\
 & Ours CADD(C) & \underline{91.49} & \underline{93.48} & \underline{92.47} & \underline{93.33} & \cellcolor{RBRedLight}\textbf{91.30} & \underline{92.31} & \underline{92.41} & \underline{92.39} & \underline{92.39} & \underline{98.82} & \underline{6.52} & $10^{-12}$***\\
 & Ours CADD(T+C) & \cellcolor{RBGreenLight}\textbf{91.67} & \cellcolor{RBGreenLight}\textbf{95.65} & \cellcolor{RBGreenLight}\textbf{93.62} & \cellcolor{RBRedLight}\textbf{95.45} & \underline{91.30} & \cellcolor{RBRedLight}\textbf{93.33} & \cellcolor{RBBlueLight}\textbf{93.56} & \cellcolor{RBBlueLight}\textbf{93.48} & \cellcolor{RBBlueLight}\textbf{93.48} & \cellcolor{RBBlueLight}\textbf{99.10} & \cellcolor{RBBlueLight}\textbf{4.35} & $10^{-13}$***\\ \midrule

RF 
& Baseline & 71.73 & 89.86 & 79.76 & 86.42 & 64.49 & 73.82 & 79.07 & 77.17 & 76.79 & 87.88 & 21.74 & --- \\
 & Ours CADD(T) & 77.23 & \underline{92.03} & 83.80 & 90.72 & 72.46 & 80.23 & 83.97 & 82.25 & 82.02 & 90.94 & 14.49 & $0.047$*\\
 & Ours CADD(C) & \underline{87.05} & \cellcolor{RBGreenLight}\textbf{92.75} & \underline{89.79} & \cellcolor{RBRedLight}\textbf{92.37} & \underline{86.23} & \underline{89.17} & \underline{89.71} & \underline{89.49} & \underline{89.48} & \underline{96.83} & \underline{11.59} & $10^{-11}$***\\
 & Ours CADD(T+C) & \cellcolor{RBGreenLight}\textbf{88.18} & 92.03 & \cellcolor{RBGreenLight}\textbf{90.06} & \underline{91.72} & \cellcolor{RBRedLight}\textbf{87.68} & \cellcolor{RBRedLight}\textbf{89.65} & \cellcolor{RBBlueLight}\textbf{89.95} & \cellcolor{RBBlueLight}\textbf{89.86} & \cellcolor{RBBlueLight}\textbf{89.85} & \cellcolor{RBBlueLight}\textbf{97.39} & \cellcolor{RBBlueLight}\textbf{9.42} & $10^{-11}$***\\ \midrule

XGBoost
& Baseline & 76.00 & 82.61 & 79.17 & 80.95 & 73.91 & 77.27 & 78.48 & 78.26 & 78.22 & 87.62 & 17.39 & --- \\
 & Ours CADD(T) & 84.62 & 71.74 & 77.65 & 75.47 & \underline{86.96} & 80.81 & 80.04 & 79.35 & 79.23 & 89.37 & 19.57 & $0.003$**\\
 & Ours CADD(C) & \cellcolor{RBGreenLight}\textbf{91.49} & \cellcolor{RBGreenLight}\textbf{93.48} & \cellcolor{RBGreenLight}\textbf{92.47} & \cellcolor{RBRedLight}\textbf{93.33} & \cellcolor{RBRedLight}\textbf{91.30} & \cellcolor{RBRedLight}\textbf{92.31} & \cellcolor{RBBlueLight}\textbf{92.41} & \cellcolor{RBBlueLight}\textbf{92.39} & \cellcolor{RBBlueLight}\textbf{92.39} & \underline{96.31} & \cellcolor{RBBlueLight}\textbf{6.52} & $10^{-8}$***\\
 & Ours CADD(T+C) & \underline{87.50} & \underline{91.30} & \underline{89.36} & \underline{90.91} & 86.96 & \underline{88.89} & \underline{89.20} & \underline{89.13} & \underline{89.13} & \cellcolor{RBBlueLight}\textbf{96.64} & \underline{8.70} & $10^{-9}$***\\ 
\bottomrule

\end{tabular}%
}
\renewcommand{\arraystretch}{1} % Values less than 1 reduce spacing
\end{table}

\begin{table}
\centering
\caption{\textbf{Performance comparison of 4 state-of-the-art (SOTA) models and 4 traditional machine learning (TML) baselines for deepfake detection on SYN data with varying the contextual information used by CADD.}
We evaluate CADD models trained under three conditions: with transcript-only (CADD(T)), with context-only (CADD(C)), and with both transcript and context knowledge (CADD(T+C)). The best performance values for each metric are highlighted in bold and marked with different background colors (\crule[RBRed]{0.2cm}{0.2cm} \crule[RBGreen]{0.2cm}{0.2cm} \crule[RBBlue]{0.2cm}{0.2cm}), while the second-best performance is underlined. The last column shows the FDR-corrected P-values (Benjamini-Hochberg procedure) from statistical significance testing (indicated by * for $P<0.05$, ** for $P<0.01$ and *** for $P<0.001$). See tables S13-S23 for full results.}
\label{tab:main_syn}
\scalebox{0.62}{
\small
\renewcommand{\arraystretch}{0.8} % Values less than 1 reduce spacing
% Updated column definition: removed 2 'c' columns
\begin{tabular}{@{}l|l|ccccccccccc|c@{}}
\toprule
 & & 
  \multicolumn{3}{c}{\cellcolor{RBGreen}Real} &
  \multicolumn{3}{c}{\cellcolor{RBRed}Fake} &
  \multicolumn{5}{c}{\cellcolor{RBBlue}Overall} & \\ \cmidrule(l){3-14} 
\textbf{Base Model} & \textbf{CADD Enhanced Model} & 
  P $(\uparrow)$ &
  R $(\uparrow)$ &
  F1 $(\uparrow)$ &
  P $(\uparrow)$ &
  R $(\uparrow)$ &
  F1 $(\uparrow)$ &
  P $(\uparrow)$ &
  R $(\uparrow)$ &
  F1 $(\uparrow)$ &
  AUC $(\uparrow)$ &
  EER  $(\downarrow)$ & 
  P-Value \\ \midrule
RawNet3 
& Baseline & \underline{94.07} & \underline{93.44} & \underline{93.71} & \underline{95.78} & \underline{96.10} & \underline{95.92} & \underline{95.11} & \underline{95.05} & \underline{95.05} & \cellcolor{RBBlueLight}\textbf{98.74} & \cellcolor{RBBlueLight}\textbf{3.55}  & --- \\
& Ours CADD(T) &
  87.35 & \cellcolor{RBGreenLight}\textbf{93.99} & 90.53 & \cellcolor{RBRedLight}\textbf{95.90} & 91.13 & 93.45 & 92.54 & 92.26 & 92.30 & \underline{98.37} & 6.74  & $0.013$* \\
& Ours CADD(C) & 88.09 & 91.80 & 89.86 & 94.55 & 91.84 & 93.15 & 92.01 & 91.83 & 91.86 & 96.48 & 6.38  & $0.018$* \\
& Ours CADD(T+C) & \cellcolor{RBGreenLight}\textbf{98.85} & 92.90 & \cellcolor{RBGreenLight}\textbf{95.77} & 95.57 &  \cellcolor{RBRedLight}\textbf{99.29} &  \cellcolor{RBRedLight}\textbf{97.39} &  \cellcolor{RBBlueLight}\textbf{96.86} & \cellcolor{RBBlueLight}\textbf{96.77} & \cellcolor{RBBlueLight}\textbf{96.75} & 98.17 & \underline{4.26}  & $0.943$ \\ 

\midrule 
LCNN 
& Baseline &  \underline{96.62} & 92.35 & 94.40 & 95.21 & \underline{97.87} & \underline{96.51} & \underline{95.77} & \underline{95.70} & \underline{95.68} & 99.15 & 4.61  & --- \\
& Ours CADD(T) &  91.35 & \cellcolor{RBGreenLight}\textbf{97.81} & \underline{94.47} & \cellcolor{RBRedLight}\textbf{98.51} & 93.97 & 96.18 & 95.69 & 95.48 & 95.51 & \cellcolor{RBBlueLight}\textbf{99.26} & \underline{3.19}  & $10^{-23}$*** \\
& Ours CADD(C) & 91.37 & \underline{96.72} & 93.94 & 97.78 & 93.97 & 95.82 & 95.26 & 95.05 & 95.08 & 98.37 & 3.55  & $10^{-12}$*** \\
& Ours CADD(T+C) & \cellcolor{RBGreenLight}\textbf{98.39} & \underline{96.72} & \cellcolor{RBGreenLight}\textbf{97.53} & \underline{97.89} & \cellcolor{RBRedLight}\textbf{98.94} & \cellcolor{RBRedLight}\textbf{98.41} & \cellcolor{RBBlueLight}\textbf{98.09} & \cellcolor{RBBlueLight}\textbf{98.06} & \cellcolor{RBBlueLight}\textbf{98.06} & \underline{99.23} & \cellcolor{RBBlueLight}\textbf{0.71}  & $10^{-26}$*** \\ \midrule

MesoNet
& Baseline & \underline{88.02}  & 91.80 & \underline{89.86} & 94.51 & \underline{91.84}  & \underline{93.15} & \underline{91.96} & \underline{91.83} & \underline{91.86} & 97.19 & \underline{6.74} & --- \\
& Ours CADD(T) & 80.34  & \cellcolor{RBGreenLight}\textbf{95.08} & 86.15 & \cellcolor{RBRedLight}\textbf{96.88} & 82.27  & 88.04 & 90.37 & 87.31 & 87.30 & \underline{98.54} & 7.45 & $0.167$ \\
& Ours CADD(C) &  87.34  & 87.98 & 87.50 & 92.01 & 91.13  & 91.50 & 90.17 & 89.89 & 89.93 & 94.80 & 9.93 & $10^{-5}$*** \\
& Ours CADD(T+C) &  \cellcolor{RBGreenLight}\textbf{93.06}  & \underline{93.99} & \cellcolor{RBGreenLight}\textbf{93.32} & \underline{96.14} & \cellcolor{RBRedLight}\textbf{95.04}  & \cellcolor{RBRedLight}\textbf{95.49} & \cellcolor{RBBlueLight}\textbf{94.93} & \cellcolor{RBBlueLight}\textbf{94.62} & \cellcolor{RBBlueLight}\textbf{94.64} & \cellcolor{RBBlueLight}\textbf{98.45} & \cellcolor{RBBlueLight}\textbf{6.03} & $10^{-17}$*** \\ \midrule 

SpecRNet 
& Baseline & 98.30 & 92.90 & 95.51 & 95.55 & \underline{98.94} & 97.21 & \underline{96.63} & \underline{96.56} & \underline{96.54} & 99.05 & 3.55  & --- \\ 
 & Ours CADD(T) & 93.68 & \cellcolor{RBGreenLight}\textbf{97.27} & 95.44 & \cellcolor{RBRedLight}\textbf{98.18} & 95.74 & 96.95 & 96.41 & 96.34 & 96.36 & 99.47 & 3.90  & $0.021$* \\ 
 & Ours CADD(C) & \cellcolor{RBGreenLight}\textbf{98.87} & 96.17 & \underline{97.50} & 97.57 & \cellcolor{RBRedLight}\textbf{99.29} & \cellcolor{RBRedLight}\textbf{98.42} & \cellcolor{RBBlueLight}\textbf{98.08} & \cellcolor{RBBlueLight}\textbf{98.06} & \cellcolor{RBBlueLight}\textbf{98.06} & \underline{99.78} & \cellcolor{RBBlueLight}\textbf{2.13}  & $10^{-5}$*** \\ 
 & Ours CADD(T+C) & \underline{98.36} & \underline{96.72} & \cellcolor{RBGreenLight}\textbf{97.52} & \underline{97.91} & \underline{98.94} & \underline{98.41} & \cellcolor{RBBlueLight}\textbf{98.08} & \cellcolor{RBBlueLight}\textbf{98.06} & \cellcolor{RBBlueLight}\textbf{98.06} & \cellcolor{RBBlueLight}\textbf{99.87} & \underline{2.48}  & $0.478$ \\ 
 
 \midrule \midrule
LR 
& Baseline & 98.31 & \cellcolor{RBGreenLight}\textbf{95.08} & \underline{96.67} & \underline{96.88} & 98.94 & \underline{97.89} & \underline{97.44} & \underline{97.42} & \underline{97.41} & \underline{99.35} & 4.92 & --- \\
 & Ours CADD(T) & \cellcolor{RBGreenLight}\textbf{100.00} & \underline{95.08} & \cellcolor{RBGreenLight}\textbf{97.48} & \cellcolor{RBRedLight}\textbf{96.91} & \cellcolor{RBRedLight}\textbf{100.00} & \cellcolor{RBRedLight}\textbf{98.43} & \cellcolor{RBBlueLight}\textbf{98.12} & \cellcolor{RBBlueLight}\textbf{98.06} & \cellcolor{RBBlueLight}\textbf{98.06} & \cellcolor{RBBlueLight}\textbf{99.90} & \cellcolor{RBBlueLight}\textbf{1.64} & $0.748$\\
 & Ours CADD(C) & \underline{100.00} & 91.80 & 95.73 & 94.95 & \underline{100.00} & 97.41 & 96.94 & 96.77 & 96.75 & 96.53 & 6.56 & $10^{-6}$***\\
 & Ours CADD(T+C) & 96.67 & 95.08 & 95.87 & 96.84 & 97.87 & 97.35 & 96.77 & 96.77 & 96.77 & 98.43 & \underline{3.28} & $10^{-5}$***\\ \midrule
SVC 
& Baseline & 93.22 & 90.16 & 91.67 & 93.75 & 95.74 & 94.74 & 93.54 & 93.55 & 93.53 & \underline{99.39} & \underline{6.56} & --- \\
 & Ours CADD(T) & 98.36 & \cellcolor{RBGreenLight}\textbf{98.36} & \cellcolor{RBGreenLight}\textbf{98.36} & \cellcolor{RBRedLight}\textbf{98.94} & 98.94 & \cellcolor{RBRedLight}\textbf{98.94} & \cellcolor{RBBlueLight}\textbf{98.71} & \cellcolor{RBBlueLight}\textbf{98.71} & \cellcolor{RBBlueLight}\textbf{98.71} & \cellcolor{RBBlueLight}\textbf{99.88} & \cellcolor{RBBlueLight}\textbf{1.64} & $0.051$\\
 & Ours CADD(C) & \cellcolor{RBGreenLight}\textbf{100.00} & \underline{91.80} & \underline{95.73} & \underline{94.95} & \cellcolor{RBRedLight}\textbf{100.00} & \underline{97.41} & \underline{96.94} & \underline{96.77} & \underline{96.75} & 97.59 & 6.56 & $10^{-4}$***\\
 & Ours CADD(T+C) & \underline{100.00} & 91.80 & 95.73 & 94.95 & \underline{100.00} & 97.41 & 96.94 & 96.77 & 96.75 & 98.85 & 6.56 & $0.040$*\\ \midrule
\multirow{4}{*}{RF} 
 & Baseline & 97.07 & 83.61 & 89.67 & 90.33 & 98.23 & 94.07 & 92.98 & 92.47 & 92.34 & \underline{98.51} & \underline{7.65} & --- \\
 & Ours CADD(T) & 98.32 & \cellcolor{RBGreenLight}\textbf{96.17} & \cellcolor{RBGreenLight}\textbf{97.23} & \cellcolor{RBRedLight}\textbf{97.57} & 98.94 & \cellcolor{RBRedLight}\textbf{98.24} & \cellcolor{RBBlueLight}\textbf{97.87} & \cellcolor{RBBlueLight}\textbf{97.85} & \cellcolor{RBBlueLight}\textbf{97.84} & \cellcolor{RBBlueLight}\textbf{99.80} & \cellcolor{RBBlueLight}\textbf{3.28} & $10^{-7}$***\\
 & Ours CADD(C) & \cellcolor{RBGreenLight}\textbf{100.00} & \underline{91.80} & \underline{95.73} & \underline{94.95} & \cellcolor{RBRedLight}\textbf{100.00} & \underline{97.41} & \underline{96.94} & \underline{96.77} & \underline{96.75} & 96.77 & 7.65 & $10^{-14}$***\\
 & Ours CADD(T+C) & \underline{100.00} & 89.07 & 94.22 & 93.38 & \underline{100.00} & 96.58 & 95.99 & 95.70 & 95.65 & 97.46 & 7.65 & $10^{-13}$***\\ \midrule

XGBoost & Baseline & 91.94 & \underline{93.44} & 92.68 & 95.70 & 94.68 & 95.19 & 94.22 & 94.19 & 94.20 & 98.57 & 4.92 & --- \\
 & Ours CADD(T) & \cellcolor{RBGreenLight}\textbf{96.83} & \cellcolor{RBGreenLight}\textbf{100.00} & \cellcolor{RBGreenLight}\textbf{98.39} & \cellcolor{RBRedLight}\textbf{100.00} & \cellcolor{RBRedLight}\textbf{97.87} & \cellcolor{RBRedLight}\textbf{98.92} & \cellcolor{RBBlueLight}\textbf{98.75} & \cellcolor{RBBlueLight}\textbf{98.71} & \cellcolor{RBBlueLight}\textbf{98.71} & \cellcolor{RBBlueLight}\textbf{99.79} & \cellcolor{RBBlueLight}\textbf{0.00} & $0.138$\\
 & Ours CADD(C) & 93.33 & 91.80 & 92.56 & 94.74 & 95.74 & 95.24 & 94.18 & 94.19 & 94.18 & 98.95 & 8.20 & $0.155$\\
 & Ours CADD(T+C) & \underline{95.00} & \underline{93.44} & \underline{94.21} & \underline{95.79} & \underline{96.81} & \underline{96.30} & \underline{95.48} & \underline{95.48} & \underline{95.48} & \underline{99.20} & \underline{4.92} & $0.943$\\ 
\bottomrule
\end{tabular}%
}
\renewcommand{\arraystretch}{1} % Values less than 1 reduce spacing
\end{table}

\begin{table}
\centering
\caption{\textbf{Performance comparison of 4 state-of-the-art (SOTA) models and 4 traditional machine learning (TML) baselines for deepfake detection on P$^2$V data with varying the contextual information used by CADD.}
We evaluate CADD models trained under three conditions: with transcript-only (CADD(T)), with context-only (CADD(C)), and with both transcript and context knowledge (CADD(T+C)). The best performance values for each metric (within each model group) are highlighted in bold and marked with different background colors (\crule[RBRed]{0.2cm}{0.2cm} \crule[RBGreen]{0.2cm}{0.2cm} \crule[RBBlue]{0.2cm}{0.2cm}), while the second-best performance is underlined. The last column shows the FDR-corrected P-values (Benjamini-Hochberg procedure) from statistical significance testing (indicated by * for $P<0.05$, ** for $P<0.01$ and *** for $P<0.001$). See table S35 for full TML results.}
\label{tab:p2v}
\scalebox{0.64}{
\small
\renewcommand{\arraystretch}{0.8} % Values less than 1 reduce spacing
\begin{tabular}{@{}l|l|ccccccccccc|c@{}}
\toprule
 & & 
  \multicolumn{3}{c}{\cellcolor{RBGreen}Real} &
  \multicolumn{3}{c}{\cellcolor{RBRed}Fake} &
  \multicolumn{6}{c}{\cellcolor{RBBlue}Overall} \\ \cmidrule(l){3-14} 
\textbf{Base Model} & \textbf{CADD Enhanced Model} & 
  P $(\uparrow)$ &
  R $(\uparrow)$ &
  F1 $(\uparrow)$ &
  P $(\uparrow)$ &
  R $(\uparrow)$ &
  F1 $(\uparrow)$ &
  P $(\uparrow)$ &
  R $(\uparrow)$ &
  F1 $(\uparrow)$ &
  AUC $(\uparrow)$ &
  EER  $(\downarrow)$ & 
  P-Value \\ \midrule

\multirow{4}{*}{RawNet3} 
& Baseline  & 46.64 & 70.34 & \underline{56.09} & 98.64 & 96.40 & 97.51 & 96.42 & 95.28 & \underline{95.73} & 92.72 & 12.18 & --- \\

& Ours CADD(T) & 32.01 & \cellcolor{RBGreenLight}\textbf{88.68} & 47.04 & \cellcolor{RBRedLight}\textbf{99.45} & 91.57 & 95.35 & \underline{96.56} & 91.45 & 93.28 & \underline{95.84} & \cellcolor{RBBlueLight}\textbf{10.00} & $10^{-7843}$*** \\

& Ours CADD(C) & \underline{48.64} & 53.43 & 50.92 & 97.91 & \cellcolor{RBRedLight}\textbf{97.48} & \underline{97.69} & 95.80 & \underline{95.59} & 95.69 & 93.32 & 14.61 & $10^{-8217}$*** \\

& Ours CADD(T+C) & \cellcolor{RBGreenLight}\textbf{51.60} & \underline{73.38} & \cellcolor{RBGreenLight}\textbf{60.60} & \underline{98.79} & 96.92 & \cellcolor{RBRedLight}\textbf{97.84} & \cellcolor{RBBlueLight}\textbf{96.77} & \cellcolor{RBBlueLight}\textbf{95.91} & \cellcolor{RBBlueLight}\textbf{96.25} & \cellcolor{RBBlueLight}\textbf{96.13} & \underline{10.89} & $10^{-5689}$*** \\ 

\midrule 

\multirow{4}{*}{LCNN}
& Baseline & 68.85 & \underline{71.52} & \cellcolor{RBGreenLight}\textbf{70.16} & \underline{98.72} & 98.55 & 98.64 & \underline{97.44} & 97.40 & \underline{97.42} & 96.35 & 9.62 & --- \\

& Ours CADD(T) & 27.87 & \cellcolor{RBGreenLight}\textbf{79.80} & 41.31 & \cellcolor{RBRedLight}\textbf{99.01} & 90.76 & 94.71 & 95.97 & 90.29 & 92.42 & 93.60 & 13.15 & $10^{-1339}$*** \\

& Ours CADD(C)  & \underline{83.60} & 53.48 & 65.23 & 97.95 & \underline{99.53} & \underline{98.73} & 97.34 & \underline{97.56} & 97.30 & \cellcolor{RBBlueLight}\textbf{97.38} & \cellcolor{RBBlueLight}\textbf{8.54} & $10^{-1002}$*** \\

& Ours CADD(T+C) & \cellcolor{RBGreenLight}\textbf{88.89} & 56.08 & \underline{68.77} & 98.07 & \cellcolor{RBRedLight}\textbf{99.69} & \cellcolor{RBRedLight}\textbf{98.87} & \cellcolor{RBBlueLight}\textbf{97.67} & \cellcolor{RBBlueLight}\textbf{97.82} & \cellcolor{RBBlueLight}\textbf{97.58} & \underline{97.27} & \underline{8.55} & $10^{-1108}$*** \\ 

\midrule

\multirow{4}{*}{MesoNet} 
& Baseline & \cellcolor{RBGreenLight}\textbf{95.38} & 9.12 & 16.64 & 96.09 & \cellcolor{RBRedLight}\textbf{99.98} & 98.00 & 96.06 & 96.09 & 94.51 & 98.08 & 5.93 & --- \\

& Ours CADD(T) & 19.67 & 1.18 & 2.22 & 95.76 & \underline{99.79} & 97.73 & 92.50 & 95.56 & 93.64 & 74.71 & 31.13 & $10^{-13024}$*** \\

& Ours CADD(C)  & \underline{82.54} & \underline{83.87} & \cellcolor{RBGreenLight}\textbf{83.20} & 99.28 & 99.21 & \cellcolor{RBRedLight}\textbf{99.24} & \cellcolor{RBBlueLight}\textbf{98.56} & \cellcolor{RBBlueLight}\textbf{98.55} & \cellcolor{RBBlueLight}\textbf{98.56} & \underline{98.86} & \underline{4.86} & $10^{-4463}$*** \\

& Ours CADD(T+C)  & 67.53 & \cellcolor{RBGreenLight}\textbf{92.06} & \underline{77.91} & \cellcolor{RBRedLight}\textbf{99.64} & 98.02 & \underline{98.82} & \underline{98.26} & \underline{97.76} & \underline{97.93} & \cellcolor{RBBlueLight}\textbf{99.20} & \cellcolor{RBBlueLight}\textbf{4.05} & $10^{-1757}$*** \\ 

\midrule

\multirow{4}{*}{SpecRNet} 
& Baseline & 24.04 & 83.58 & 37.34 & 99.17 & 88.18 & 93.36 & 95.96 & 87.99 & 90.96 & 93.23 & 13.49 & --- \\

& Ours CADD(T)  & 13.69 & \cellcolor{RBGreenLight}\textbf{91.76} & 23.83 & \cellcolor{RBRedLight}\textbf{99.51} & 74.12 & 84.96 & 95.83 & 74.87 & 82.34 & 92.32 & 15.72 & $10^{-2496}$***   \\

& Ours CADD(C) & \underline{45.14} & \underline{86.23} & \underline{59.26} & \underline{99.36} & \underline{95.31} & \underline{97.29} & \underline{97.04} & \underline{94.92} & \underline{95.66} & \underline{94.89} & \underline{10.66} & $10^{-847}$***  \\

& Ours CADD(T+C) & \cellcolor{RBGreenLight}\textbf{59.70} & 82.84 & \cellcolor{RBGreenLight}\textbf{69.39} & 99.22 & \cellcolor{RBRedLight}\textbf{97.50} & \cellcolor{RBRedLight}\textbf{98.35} & \cellcolor{RBBlueLight}\textbf{97.53} & \cellcolor{RBBlueLight}\textbf{96.87} & \cellcolor{RBBlueLight}\textbf{97.11} & \cellcolor{RBBlueLight}\textbf{97.84} & \cellcolor{RBBlueLight}\textbf{6.44} & $10^{-2501}$***  \\ \midrule \midrule 

\multirow{4}{*}{LR} 
& Baseline  & 5.77 & \cellcolor{RBGreenLight}\textbf{45.34} & 10.23 & \underline{97.88} & 77.34 & 86.40 & \cellcolor{RBBlueLight}\textbf{95.15} & 76.39 & 84.14 & 63.48 & \underline{42.83} & --- \\
& Ours CADD(T)  & \underline{5.79} & \cellcolor{RBGreenLight}\textbf{45.34} & \underline{10.27} & \cellcolor{RBRedLight}\textbf{97.89} & \underline{77.45} & \underline{86.47} & \cellcolor{RBBlueLight}\textbf{95.15} & \underline{76.49} & \underline{84.21} & 63.53 & \underline{42.83} & 0.985 \\
& Ours CADD(C)  & \cellcolor{RBGreenLight}\textbf{5.81} & \cellcolor{RBGreenLight}\textbf{45.34} & \cellcolor{RBGreenLight}\textbf{10.30} & \cellcolor{RBRedLight}\textbf{97.89} & \cellcolor{RBRedLight}\textbf{77.51} & \cellcolor{RBRedLight}\textbf{86.52} & \cellcolor{RBBlueLight}\textbf{95.15} & \cellcolor{RBBlueLight}\textbf{76.56} & \cellcolor{RBBlueLight}\textbf{84.25} & \cellcolor{RBBlueLight}\textbf{63.73} & \cellcolor{RBBlueLight}\textbf{42.65} & 0.985 \\
& Ours CADD(T+C)  & \cellcolor{RBGreenLight}\textbf{5.81} & \cellcolor{RBGreenLight}\textbf{45.34} & \cellcolor{RBGreenLight}\textbf{10.30} & \cellcolor{RBRedLight}\textbf{97.89} & \cellcolor{RBRedLight}\textbf{77.51} & \cellcolor{RBRedLight}\textbf{86.52} & \cellcolor{RBBlueLight}\textbf{95.15} & \cellcolor{RBBlueLight}\textbf{76.56} & \cellcolor{RBBlueLight}\textbf{84.25} & \underline{63.68} & \cellcolor{RBBlueLight}\textbf{42.65} & 0.985 \\ \midrule 

\multirow{4}{*}{SVC} 
& Baseline  & \underline{23.06} & \underline{21.33} & 22.16 & \underline{97.60} & \cellcolor{RBRedLight}\textbf{97.82} & \cellcolor{RBRedLight}\textbf{97.71} & 95.39 & \cellcolor{RBBlueLight}\textbf{95.55} & \cellcolor{RBBlueLight}\textbf{95.47} & 73.42 & \cellcolor{RBBlueLight}\textbf{34.77} & --- \\
& Ours CADD(T)  & 22.82 & 21.15 & 21.95 & 97.59 & \underline{97.81} & \underline{97.70} & 95.37 & \underline{95.54} & 95.45 & 73.42 & \underline{35.13} & 0.985 \\
& Ours CADD(C)  & \underline{23.06} & \cellcolor{RBGreenLight}\textbf{21.86} & \underline{22.45} & \cellcolor{RBRedLight}\textbf{97.61} & 97.77 & 97.69 & \underline{95.40} & 95.52 & \underline{95.46} & \cellcolor{RBBlueLight}\textbf{73.53} & \cellcolor{RBBlueLight}\textbf{34.77} & 0.985 \\
& Ours CADD(T+C)  & \cellcolor{RBGreenLight}\textbf{23.24} & \cellcolor{RBGreenLight}\textbf{21.86} & \cellcolor{RBGreenLight}\textbf{22.53} & \cellcolor{RBRedLight}\textbf{97.61} & 97.79 & \underline{97.70} & \cellcolor{RBBlueLight}\textbf{95.41} & \underline{95.54} & \cellcolor{RBBlueLight}\textbf{95.47} & \underline{73.50} & \cellcolor{RBBlueLight}\textbf{34.77} & 0.985 \\ \midrule 

\multirow{4}{*}{RandomForest} 
& Baseline  & \underline{53.63} & \underline{1.73} & \underline{3.36} & \underline{97.08} & \cellcolor{RBRedLight}\textbf{99.95} & \cellcolor{RBRedLight}\textbf{98.50} & \underline{95.79} & \cellcolor{RBBlueLight}\textbf{97.04} & \underline{95.67} & \cellcolor{RBBlueLight}\textbf{64.85} & 40.08 & --- \\
& Ours CADD(T)  & \cellcolor{RBGreenLight}\textbf{55.81} & \cellcolor{RBGreenLight}\textbf{2.03} & \cellcolor{RBGreenLight}\textbf{3.92} & \cellcolor{RBRedLight}\textbf{97.09} & \cellcolor{RBRedLight}\textbf{99.95} & \cellcolor{RBRedLight}\textbf{98.50} & \cellcolor{RBBlueLight}\textbf{95.86} & \cellcolor{RBBlueLight}\textbf{97.04} & \cellcolor{RBBlueLight}\textbf{95.69} & \underline{64.81} & \underline{40.02} & 0.985 \\
& Ours CADD(C)  & 50.00 & \underline{1.73} & 3.35 & \underline{97.08} & \cellcolor{RBRedLight}\textbf{99.95} & \underline{98.49} & 95.68 & \underline{97.03} & \underline{95.67} & 64.46 & \underline{40.02} & 0.985 \\
& Ours CADD(T+C)  & 43.80 & 1.49 & 2.89 & 97.07 & \underline{99.94} & \underline{98.49} & 95.49 & 97.02 & 95.65 & 64.56 & \cellcolor{RBBlueLight}\textbf{39.73} & 0.985 \\ \midrule 

\multirow{4}{*}{XGBoost} 
& Baseline  & 20.63 & \cellcolor{RBGreenLight}\textbf{22.22} & 21.40 & \underline{97.61} & 97.38 & 97.50 & 95.33 & 95.15 & \underline{95.24} & \cellcolor{RBBlueLight}\textbf{69.98} & \underline{36.02} & --- \\
& Ours CADD(T)  & \cellcolor{RBGreenLight}\textbf{22.08} & \underline{22.04} & \cellcolor{RBGreenLight}\textbf{22.06} & \cellcolor{RBRedLight}\textbf{97.62} & \underline{97.62} & \underline{97.62} & \cellcolor{RBBlueLight}\textbf{95.37} & \underline{95.38} & \cellcolor{RBBlueLight}\textbf{95.38} & \underline{69.93} & 36.38 & 0.708 \\
& Ours CADD(C)  & 18.62 & 18.82 & 18.72 & 97.52 & 97.48 & 97.50 & 95.17 & 95.15 & 95.16 & 69.89 & \cellcolor{RBBlueLight}\textbf{35.30} & 0.016* \\
& Ours CADD(T+C)  & \underline{21.88} & 21.33 & \underline{21.60} & 97.60 & \cellcolor{RBRedLight}\textbf{97.67} & \cellcolor{RBRedLight}\textbf{97.63} & \underline{95.35} & \cellcolor{RBBlueLight}\textbf{95.40} & \cellcolor{RBBlueLight}\textbf{95.38} & 69.48 & 37.10 & $10^{-7}$*** \\  

\bottomrule
\end{tabular}%
}
\renewcommand{\arraystretch}{1} % Reset spacing
\end{table}

%%%%%%%%%%%%%%%% REFERENCES %%%%%%%%%%%%%%%

% After the paper has completed peer review and been revised ready for acceptance,
% you should comment out the lines above and copy-paste the contents of your .bbl
% file here instead. This will help ensure that our conversion software works correctly.
% Remember to re-run BibTeX first - check the timestamp!
%
% Example of the first three entries copy-pasted from science_template.bbl:
%
%\begin{thebibliography}{1}
%
%\bibitem{example}
%A.~N. {Author}, An example reference. \emph{Journal of Improbable Research}
%  \textbf{1}, 67 (2020).
%
%\bibitem{example2}
%F.~M. {Surname}, S.~{Author}, A second example. \emph{Interesting Research
%  Letters} \textbf{32}, 897 (2019).
%
%\bibitem{example_preprint}
%P.~{One}, P.~{Two}, P.~{Three}, {An unpublished preprint}. \emph{preprint}
%  (2021), arXiv:2101.12345.
%
%\end{thebibliography}

%%%%%%%%%%%%%%%% SUPPLEMENT LIST %%%%%%%%%%%%%%%

% List the contents of your Supplementary Materials, including the numbers of any
% supplementary figures, tables, external data files etc. and any references that are
% cited only in the supplement. In this example, refs. 7-8 are cited only in the supplement.
% Fill out your numbers accordingly and delete any lines that aren't applicable.
% \newpage \hphantom\newpage\newpage
\clearpage %\hphantom
\appendix
% References \textit{(7-\arabic{enumiv})}\\ % automatically fills out the last reference number
% (filling out the other numbers automatically is possible but fiddly and liable to break)
% Movie S1\\
% Data S1

%%%%%%%%%%%%%%%% END OF MAIN TEXT %%%%%%%%%%%%%%%

\newpage

%%%%%%%%%%%%%%%% START OF SUPPLEMENT %%%%%%%%%%%%%%%

% Figures, tables, equations and pages in the supplement are numbered S1, S2 etc.
\renewcommand{\thefigure}{S\arabic{figure}}
\renewcommand{\thetable}{S\arabic{table}}
\renewcommand{\theequation}{S\arabic{equation}}
\renewcommand{\thepage}{S\arabic{page}}
\setcounter{figure}{0}
\setcounter{table}{0}
\setcounter{equation}{0}
\setcounter{page}{1} % not 0 as \newpage already started a supplementary page
% References continue the numbering from the main text.

%%%%%%%%%%%%%%%% SUPPLEMENT TITLE PAGE %%%%%%%%%%%%%%%

\begin{center}
\section*{Supplementary Materials for\\ \scititle}

% Author list for the supplement
% Indicate the corresponding authors, but do NOT include institutions here
% It would be nice if the template auto-generated this, but doing so is complicated...
Chongyang Gao$^{1\dagger}$,
Marco Postiglione$^{1\dagger}$, 
Julian Baldwin$^{1}$,
Natalia Denisenko$^1$, \\
Isabel Gortner$^1$,
Luke Fosdick$^1$,
Chiara Pulice$^1$,
Sarit Kraus$^2$,
V. S. Subrahmanian$^{1\ast}$ \\
\small$^\ast$Corresponding author. Email: vss@northwestern.edu\\
\small$^\dagger$These authors contributed equally to this work.
\end{center}

% Fill out the numbers for each type of supplementary material,
% and delete any lines that aren't applicable.
% These are just example numbers that don't match the rest of this template.
\subsubsection*{This PDF file includes:}
Materials and Methods \\
Supplementary Text \\
Figs. S1 to S16 \\
Tables S1 to S38 \\
Machine Learning Checklist

\newpage

%%%%%%%%%%%%%%%% MATERIALS AND METHODS %%%%%%%%%%%%%%%
\subsection*{Materials and Methods}

Throughout this section, we will use the following example as a running case study to concretely demonstrate each step of the CADD methodology. 

\begin{runningexample}[Financial Fraud Deepfake]
Around June 24, 2024, a sophisticated deepfake audio emerged impersonating Mr. Mukesh Ambani, one of India's most prominent business leaders, and providing unsolicited investment advice\footnote{\url{https://www.asiafinancial.com/deepfake-of-asias-richest-man-used-in-india-stock-market-scam}}. The fabricated audio claims: \textit{``Do you really trust your own judgment? Do you really think you can make money with your stocks? Follow me and my student, Mr. Venit. We will provide you with free stock diagnosis and investment plan customization and recommend stocks worth buying in the future stock market. Time is limited. The number of people is limited. Paid resources are now shared for free. Everyone is welcome to join."}\newline 

This deepfake highlights how malicious actors exploit public figures for financial fraud. The audio was designed to mimic Mr. Ambani's speech patterns and leverage his credibility to lure potential investors into an investment scam.\newline 

\end{runningexample}

\subsubsection*{Journalist-provided Deepfake Dataset (JDD)}
We first explain our methodology for collecting our Journalist-provided Deepfake Dataset (JDD), featuring real-world deepfake audios of public figures, followed by a detailed description of how we gather contextual information for each sample. Our dataset includes both real and deepfake publicly-available audio sourced from various platforms. Over 70 journalists from renowned news organizations worldwide (\eg, New York Times, Wall Street Journal, CNN, Agence France Presse, and others) contributed to the dataset by submitting deepfake artifacts they encountered during their investigations. These submissions enabled us to collect deepfake audio of public figures that could pose societal risks. Our data sources are outlined below:

\begin{itemize}
    \item \textit{Google Alerts.} Since February 17, 2024, we have been collecting news articles, blog posts, and other media flagged by Google Alerts for the keyword ``deepfake''.
    \item \textit{Fact-Checkers.} We also continuously monitor deepfake content publicly shared by accredited fact-checking organizations, including Africa Check, AFP Fact Check, Incident Database, Boomlive, FactCheck.org, OpenSecrets, PolitiFact, and Snopes\footnote{Fact-checking websites: \href{https://africacheck.org}{Africa Check}, \href{https://factcheck.afp.com}{AFP Fact Check}, \href{https://incidentdatabase.ai}{Incident Database}, \href{https://www.boomlive.in}{Boomlive}, \href{https://www.factcheck.org}{FactCheck.org}, \href{https://www.opensecrets.org}{OpenSecrets}, \href{https://www.politifact.com}{PolitiFact}, \href{https://www.snopes.com}{Snopes}.}. These sources provide reliable identification of media containing manipulated content. Some of these sources have also asked us for help in determining whether certain content is likely real or fake.
    \item \textit{GODDS Platform \cite{postiglione2024godds}. } We have made the Global Online Deepfake Detection System (GODDS) platform\footnote{GODDS platform: \href{https://godds.ads.northwestern.edu}{godds.ads.northwestern.edu}} publicly available \emph{to journalists only} (this restriction was imposed by our university lawyers) on July 8, 2024. GODDS allows users to submit suspected deepfake content. To date, our platform has supported over 70 journalists globally who wish to know if an artifact is likely to be real or fake. 
\end{itemize}

Authentic audio content is sourced from videos of well-known deceased public figures that are publicly available on YouTube. Our experiments results (see Tables S1-S12) show that past state-of-the-art models show significantly reduced performance on the collected real-world dataset.

To enhance the performance of our CADD algorithm, we incorporate contextual information about the subject portrayed in the alleged deepfake audios. This context is essential to provide a more comprehensive understanding of the circumstances in which the media was generated or circulated. We consider three main sources of context, which are detailed below.

\paragraph{General Information} We extract general biographical and factual data about the subject using the WikiData API \footnote{\href{https://www.wikidata.org/wiki/Wikidata:REST_API}{wikidata.org}}. This source provides structured data on a wide range of entities, including individuals, organizations, and places. For each subject, we retrieve key attributes such as a brief description, occupations, and demographic information. We also gather publicly available personal data where available, such as spouse, children, and gender, as well as broader social presence metrics like the number of social media followers. This helps in profiling the subject with detailed background information that might be relevant to understanding the intent or nature of the alleged deepfake.

\begin{runningexample}[General Information]
    Key details extracted from WikiData about Mukesh Ambani include:
    \begin{itemize}
        \item \textbf{Description: } Indian businessman
        \item \textbf{Gender: } Male
        \item \textbf{Birth Date: } April 19, 1957
        \item \textbf{Nationality: } India
        \item \textbf{Occupation:} Entrepreneur, businessperson, graphic designer
        \item \textbf{Education: } University of Mumbai, Stanford University, Stanford Graduate School of Business, Institute of Chemical Technology, Hill Grange High School
        \item \textbf{Spouse:} Nita Ambani 
        \item \textbf{Children: } Akash Ambani, Isha Ambani, Anant Ambani
        \item \textbf{Followers: } n.a.
    \end{itemize}

    These contextual details are crucial parts of our audio deepfake detection framework, as they provide a rich background against which anomalous audio content can be cross-referenced and validated by CADD models.
\end{runningexample}    

\paragraph{News Articles} We extract news articles available up to the date the deepfake was published to gather the latest public narratives involving the subject at that time. Using WorldNewsAPI\footnote{\href{https://worldnewsapi.com}{worldnewsapi.com}}, we retrieve the 10 most recent news articles published before the deepfake's release that mention the subject. From these articles, we extract both the title and the full text, which serve as an indicator of how the subject has been portrayed in recent media. This information helps identify discrepancies between the portrayal in the alleged deepfake and current real-world events or perceptions.

\begin{runningexample}[News Articles]
We retrieved 10 recent news articles mentioning Mukesh Ambani before the deepfake's release, including a highly relevant piece from DNA India\footnote{\url{https://www.dnaindia.com/mumbai/report-mukesh-ambani-deepfake-video-mumbai-doctor-falls-prey-to-frauds-loses-rs-7-lakh-in-3094299}}. The article describes a scam where a doctor lost 710,000 Indian Rupees (approximately \$ 8,500) after being deceived by a deepfake video of Mukesh Ambani that misrepresents him as promoting a  different fraudulent share trading academy. This prior video portrays valuable context suggesting that scammers have previously impersonated public figures like Mukesh Ambani---a pattern our CADD system is designed to identify. Details of the other news articles are included in supplementary text.
\end{runningexample}

\paragraph{Social Media Posts} We also consider social media activity to capture real-time discussions about the subject at the time the deepfake was published. We used Reddit's PRAW (Python Reddit API Wrapper)  API\footnote{\href{https://praw.readthedocs.io/en/stable/}{praw.readthedocs.io}} to gather up to 10 Reddit posts mentioning the subject. We followed a similar approach to the one used for news articles by retrieving posts available up to the date the deepfake was published. For each post, we extract the title, main body of the post, and the first 10 comments. This allows us to assess public sentiment, debate, and commentary surrounding the subject, providing additional social context that could influence the impact or plausibility of the deepfake content.

\begin{runningexample}[Social Media Posts]
We retrieved Reddit posts about Mukesh Ambani prior
to the deepfake incident, revealing nuanced contextual insights that could aid deepfake detection. One particularly informative post\footnote{\url{https://www.reddit.com/r/Maharashtra/comments/1d99962/gujju_mukesh_ambani_to_invest_13400_crores_to_buy/}} discussed a potential land investment of 13,400 crores in Navi Mumbai, underscoring Ambani's ongoing business activities and investment patterns. This context enables cross-referencing of deepfake content against current, verifiable information about his business ventures. Another post\footnote{\url{https://www.reddit.com/r/updatenewdailyweekly/comments/1d1ja2i/rils_mukesh_ambani_set_for_african_safari_with_5g/}} highlighted Mukesh Ambani's international business expansion plans, specifically his initiative to bring 5G tech solutions to Africa. This detailed portrayal of Ambani's strategic vision and technological innovation provides a baseline for comparing the authenticity of potential deepfake content. Details of all the Reddit posts are included in supplementary text.
\end{runningexample}

\paragraph{Dataset Statistics}

The JDD dataset comprises 255 real-world deepfake audio samples, spanning from November 18, 2015, to July 31, 2024. Due to our data-gathering methodology and the recent acceleration in deepfake generation techniques, the majority of samples were obtained in 2023 and 2024, as shown in Figure~\ref{fig:godds-rw-timetrend}. 

For model training, we augmented the deepfake samples with an equal number (255) of authentic audio samples, derived from public speeches featuring well-known figures and published between 2009 and 2024. This results in a balanced dataset of 510 audio samples, evenly split between real and deepfake instances. To facilitate model evaluation, we stratified the dataset into training (70\%), validation (10\%), and test (20\%) sets. 

For each authentic and deepfake sample in the JDD dataset, contextual information is gathered using the methodology outlined previously. Figure~\ref{fig:godds-rw-subjectcategories} presents the categories to which the subjects in the dataset belong. Entertainment dominates with approximately 150 subjects, followed by Politics with around 60 subjects. Other categories including Music, Media, Sports, Writing, Fashion, Business, Law, Academia \& Research, and Activism show progressively smaller representations. These categories were derived by grouping occupations retrieved from WikiData based on the names of the subjects. For each individual, we consider only the first occupation listed by WikiData, which reflects the most prominent or widely recognized role associated with that person. Please refer to supplementary text for more details about the mapping between occupations and categories.

\paragraph{Limitations}
The dataset is designed to capture deepfake audio manipulation primarily involving entertainment and political figures, reflecting the significant societal impact of deepfakes targeting high-profile individuals. While this focus does not encompass all public personalities, it aligns with the broader concern that deepfakes of prominent figures can shape public discourse and influence trust in media. The dataset primarily draws from English-language sources such as Western news organizations and platforms like Reddit, ensuring coverage of widely circulated deepfake incidents. Additionally, the dataset predominantly consists of samples from 2023-2024, offering insights into the most recent advancements in deepfake technology. Data sources---including Google Alerts, fact-checking websites, and the GODDS platform---were selected for their reliability in identifying verified deepfake cases, though deepfakes circulating in less monitored spaces may be underrepresented. Finally, the dataset's focus on well-known public figures might not accurately reflect the risks and characteristics of deepfakes targeting less prominent individuals. However, it provides a valuable foundation for studying deepfake detection in high-impact scenarios, recognizing that the broader implications extend to individuals beyond those directly represented.

\subsubsection*{Synthetic Audio Dataset (SYN)} 
\label{sec:SYN_dataset}

To explore the effects of context information, we also created the Synthetic Audio Dataset (SYN) following a two-phase process applied to a set of deceased public figures, as detailed in Algorithm~\ref{alg:transcript-generation} and Algorithm~\ref{alg:audio-generation}.

\paragraph{Phase 1: Fake Transcript Generation}

The first phase, detailed in Algorithm 1, generates a diverse set of speech transcripts for deceased public figures. For each subject in our dataset, we use a Large Language Model (LLM)\footnote{GPT-4o was used in our experiments.} to produce four distinct transcripts. Specifically, based on the assumption that the LLM has acquired contextual knowledge about the public figure during its pre-training, our process begins by generating two base transcripts: one representing what the subject \emph{would say}, maintaining their typical speaking style and beliefs, and another containing content they \emph{would not say}, which introduces deliberate out-of-character deviations. Including both types allows us to build a more balanced dataset and avoid biasing detection models toward either stylistic mimicry or unrealistic speech patterns. If we relied only on in-character speech, detectors might focus narrowly on stylistic cues. Conversely, using only out-of-character content could result in systematic artifacts that models might exploit. Refer to supplementary text for more details about the prompts used for this phase. 

To simulate common deepfake scenarios in which fabricated speech is tied to real-world events, we generate two additional context-aware transcripts. These are created by conditioning on external information retrieved from the WorldNewsAPI\footnote{\url{http://worldnewsapi.com}}. When available, we use the publication date of the authentic audio associated with the subject. If no such date exists—especially for individuals who passed away long before recent news coverage—we randomly sample a date from January 1, 2023, onward\footnote{This start date is imposed by the coverage limits of WorldNewsAPI, which provides more reliable results for recent content.}. Given the selected date, we retrieve related news articles and feed them to the LLM along with instructions to generate two further transcripts: one aligned with the subject’s persona and views in the context of the news, and one misaligned or implausible given their public character. This contextualization ensures that the fake content reflects realistic deepfake use cases, such as misinformation around major events or political discourse involving well-known figures.

\begin{algorithm}[t]
\caption{Generation of fake transcripts for the Synthetic Voice Cloning dataset}
\label{alg:transcript-generation}
\begin{algorithmic}[1]
\Require $\textit{AuthenticSamples}$ \Comment{Set of (subject, audio) pairs}
\Require $\textit{startDate}$

\Procedure{GenerateFakeTranscripts}{$\textit{AuthenticSamples}$, $\textit{startDate}$}
    \State $\textit{FakeTranscripts} \gets \emptyset$
    \ForAll{$(\textit{subject}, \textit{audio}) \in \textit{AuthenticSamples}$}
        \ForAll{$\textit{intent} \in \{\text{would say, would not say}$\}}
            \State $\textit{transcript} \gets \textsc{LLM}(\textit{subject}, \textit{intent})$
            \State \textit{FakeTranscripts}.\textsc{add}(\textit{subject}, \textit{audio}, \textit{transcript})
        \EndFor
        \State $\textit{date} \gets \textsc{SelectDate}(\textit{startDate}, \text{now})$
        \State $\textit{context} \gets \textsc{ContextExtraction}(\textit{subject}, \textit{date})$
        \ForAll{$\textit{intent} \in \{\text{would say, would not say}$\}}
            \State $\textit{transcript} \gets \textsc{LLM}(\textit{subject}, \textit{intent}, \textit{context})$
            \State \textit{FakeTranscripts}.\textsc{add}(\textit{subject}, \textit{audio}, \textit{transcript})
        \EndFor
    \EndFor
    \Return $\textit{FakeTranscripts}$
\EndProcedure
\end{algorithmic}
\end{algorithm}

\paragraph{Phase 2: Synthetic Audio Generation}

The second phase, outlined in Algorithm 2, transforms the generated transcripts into synthetic audio samples using various voice cloning methods, specifically XTTS-v2 \cite{casanova2024xtts}, OpenVoice-v2 \cite{qin2023openvoice}, MetaVoice \cite{metavoice2024}, and WhisperSpeech \cite{whisperspeech2024}. To ensure a fair evaluation of different voice cloning techniques, we implement a balanced assignment strategy. This strategy distributes the voice cloning methods evenly across the dataset, ensuring each method is represented equally in the final collection.

For each transcript, we assign a voice cloning method according to our balanced distribution and generate synthetic audio using the assigned method. The generation process uses the original audio as a reference voice and the generated transcript as content. 

\begin{algorithm}[t]
\caption{Generation of fake audio for the Synthetic Voice Cloning dataset}
\label{alg:audio-generation}
\begin{algorithmic}[1]
\Require $\textit{FakeTranscripts}$ \Comment{Set of (subject, audio, transcript)}
\Require $\textit{VoiceCloningMethods}$ \Comment{Set of available methods}
\Procedure{GenerateFakeDataset}{$\textit{FakeTranscripts}$, $\textit{VoiceCloningMethods}$}
    \State $\textit{FakeDataset} \gets \emptyset$
    \State $\textit{methodAssignments} \gets \textsc{BalancedAssignments}(\textit{VoiceCloningMethods}, \textit{FakeDataset}.\textsc{len})$
    
    \ForAll{$(\textit{subject}, \textit{audio}, \textit{transcripts}) \in \textit{FakeTranscripts}$}
            \State $\textsc{VoiceCloner} \gets \textsc{GetNextBalancedMethod}(\textit{methodAssignments})$
            \State $\textit{fakeAudio} \gets \textsc{VoiceCloner}(\textit{audio}, \textit{transcript})$
            \State $\textit{FakeDataset}.\textsc{add}(\textit{subject}, \textit{transcript}, \textit{method}, \textit{fakeAudio})$
    \EndFor
    
    \Return $\textit{FakeDataset}$
\EndProcedure
\end{algorithmic}
\end{algorithm}

\paragraph{Dataset Statistics}
The SYN dataset comprises 468 synthetic deepfake audio samples from December 12, 2014 to July 21, 2024. As in the case of the JDD dataset, we supplemented the dataset with 202 publicly available authentic audio samples, which led to a total of 771 instances. The dataset was stratified into training (70\%), validation (10\%) and test (20\%) sets.

\paragraph{Limitations}
By focusing exclusively on deceased public figures, the dataset enables a controlled investigation of synthetic voice cloning technologies while mitigating ethical concerns associated with generating artificial speech for living individuals. However, this focus limits the dataset’s applicability to real-world scenarios where deepfakes target living individuals, potentially overlooking nuances in detection performance for such cases. The dataset includes four specific voice cloning methods (XTTS-v2, OpenVoice-v2, MetaVoice, and WhisperSpeech), allowing for a detailed comparative analysis of state-of-the-art techniques. While this approach ensures rigorous evaluation, it does not encompass the full range of available voice cloning technologies, meaning certain manipulation techniques may not be represented. Additionally, while the dataset systematically balances voice cloning methods and includes both in-character and out-of-character speech, its structured design may not fully capture the unpredictable nature of deepfake audio in-the-wild. Although the incorporation of contextual information from recent news helps simulate realistic scenarios, it does not account for all possible use cases, such as spontaneous or highly deceptive manipulations that might emerge outside of controlled settings.

\subsubsection*{In-The-Wild Dataset (ITW)}
The publicly-available In-The-Wild dataset \cite{muller2022does} consists of 37.9 hours of audio segmented into 31,779 samples, with an average duration of 4.3 seconds per clip. The collection contains both authentic (19,963 samples, 20.7 hours) and synthetic (11,816 samples, 17.2 hours) audio, all uniformly downsampled to 16 kHz. For our experiments, we divided the dataset into training (70\%; 22,245 samples), validation (10\%; 3,178 samples), and testing (20\%; 6,356 samples) splits in a stratified manner to maintain class distribution.

While the dataset provides valuable variations, its construction methodology introduces significant limitations for real-world deepfake detection applications. The synthetic samples were deliberately collected from 219 publicly available audio or video files that explicitly advertise themselves as deepfakes, typically featuring speakers in conspicuously out-of-character scenarios (\eg, "Donald Trump reads Star Wars"). This selection criterion creates an artificial distinction that rarely exists in malicious deepfakes deployed in real environments, where creators attempt to maximize plausibility rather than absurdity. In addition, the genuine samples were retrieved to match the characteristics of their synthetic counterparts, controlling for speaker emotion, background noise, and duration. While this pairing approach facilitates comparative analysis, it further distances the dataset from reflecting authentic detection challenges encountered in uncontrolled settings. These limitations motivated our collection and contribution of the Journalist-provided Deepfake Dataset (JDD), which addresses these shortcomings by incorporating samples that better represent real-world malicious deepfakes.

Another notable limitation of the original dataset is the absence of contextual metadata, including publication dates for the synthetic content. To apply our CADD framework, we assigned a uniform publication date (July 24, 2024) across all news pieces and social media posts providing context. This artificial synchronization limits our ability to fully leverage temporal context that would typically be available when detecting deepfakes in natural environments. Despite this constraint, our experiments demonstrated meaningful performance improvements when incorporating even this simulated contextual information.

\subsubsection*{Perturbed Public Voices (P$^{2}$V)}
We additionally evaluate our method on \textbf{Perturbed Public Voices (P$^{2}$V)}~\cite{gao2025perturbed}, a large-scale, IRB-approved dataset designed to reflect realistic and adversarial audio deepfake scenarios. P$^{2}$V consists of 257,440 audio samples drawn from 206 deceased public figures, including 247,200 synthetic and 10,240 authentic recordings. All audio is uniformly processed at 16~kHz and partitioned at the subject level into training (60\%), validation (20\%), and test (20\%) splits to prevent speaker leakage across splits.

Unlike ITW, where synthetic samples are often semantically implausible or deliberately absurd, P$^{2}$V explicitly enforces \emph{identity-consistent and contextually plausible} speech content. Synthetic transcripts are generated using large language models conditioned on real news articles published during the subject’s lifetime, ensuring that the generated speech aligns with the public persona, historical context, and linguistic style of the impersonated figure. This design better approximates malicious deepfakes, where attackers aim to maximize plausibility rather than conspicuousness. 

It is important to note that P$^{2}$V does not provide the rich, externally verified contextual information leveraged by CADD, compared to our JDD. In particular, its samples are not curated or validated by professional journalists, making it difficult to reliably distinguish subtle real-world impersonation intent and narrative consistency. Moreover, P$^{2}$V remains a controlled synthetic benchmark rather than a truly in-the-wild dataset, and its voice generation pipelines cover only a limited subset of real-world deepfake production practices. As a result, P$^{2}$V should be viewed as a complementary benchmark that captures challenging synthetic conditions, rather than a complete proxy for real-world, context-rich deepfake scenarios such as those provided by the JDD dataset.

\subsubsection*{Cross-Validation}
To further assess robustness and mitigate sampling bias, we conduct 10-fold cross-validation over the four strongest baselines and our CADD variants on JDD. As summarized in Table S12, CADD consistently outperforms RawNet3, LCNN, MesoNet, and SpecRNet baselines across all folds and evaluation metrics. When both information sources are combined (CADD(T+C)), the model achieves the highest precision, recall, F1-score, and AUC across real and fake classes, while also reducing EER to the lowest levels observed. These results demonstrate that CADD’s improvements are stable across data splits and that contextual signals provide complementary benefits for deepfake detection, reinforcing the effectiveness and generalizability of the proposed framework.

\subsubsection*{Context-based Audio Deepfake Detection}

In this section, we present an in-depth description of our Context-based Audio Deepfake Detection (CADD) system. CADD focuses on detecting deepfakes targeting public figures. Deepfakes frequently feature such individuals, and detection accuracy can be enhanced by incorporating contextual information surrounding the artifact. 

\paragraph{Features Extraction}

We first introduce the method used to embed text artifacts and then describe how this method is applied to obtain representations of context and transcript. 

\smallskip\noindent\textit{Text Feature Extraction.} The text data $t$ --- which may come from various sources such as audio transcripts, news articles, or Reddit posts --- is first tokenized into a sequence of tokens $\mathbf{x} = [x_{1}, \dots, x_{T}]$ using a pre-trained tokenizer. Given that text has different lengths, we apply padding to the maximum sequence length in the batch and truncation when necessary. The tokenizer also generates an attention mask $\mathbf{M} = [m_{1}, ..., m_{T}]$ where:
\begin{equation}
m_{i} = \begin{cases}
1 & \text{if token } t \text{ is not a padding token} \\
0 & \text{otherwise}
\end{cases}
\end{equation}

The tokenized input is processed through a SentenceBERT \cite{reimers-2019-sentence-bert} transformer model $f_\theta(\cdot)$ with parameters $\theta$, producing contextual token embeddings:
\begin{equation}
\mathbf{H}i = f_\theta(\mathbf{x}) \in \mathbb{R}^{T \times d}
\end{equation}
where $T$ is the sequence length and $d$ is the embedding dimension. In our experiments, we set $d=100$ for simplicity and generalizability.

We use attention-weighted mean pooling over the token embeddings to obtain a fixed-size text representation $\mathbf{e}_i \in \mathbb{R}^d$. This accounts for padding tokens by using the attention mask in the averaging operation:
\begin{equation}
\mathbf{e} = \frac{\sum_{i=1}^T m_{i}\mathbf{h}_{i}}{\max(\sum_{i=1}^T m_{i}, \epsilon)}
\end{equation}
where $\mathbf{h}{_i}$ is the $i$-th token embedding, $m_{i}$ is the corresponding attention mask value, and $\epsilon = 10^{-9}$ is a small constant added for numerical stability.

This pooling strategy ensures that: (1) only valid tokens contribute to the final sentence representation, (2) the contribution of each token is properly normalized, and (3) the resulting embedding preserves the dimensionality of the original token representations.

\smallskip\noindent\textit{Context Features.} We used WikiData to extract a comprehensive profile of the subject including gender, occupations, details about their spouse and children (if applicable, including the number of children), and their social media follower count. The subject’s description and occupation are embedded, and these embeddings are averaged to produce a single representation. Gender and relationships with spouses and children are represented as binary values. Other numerical information required no additional processing.

We generate embedded representations of news article titles retrieved via the WorldNewsAPI\footnote{\href{https://worldnewsapi.com}{worldnewsapi.com}} by individually processing each text and then averaging the resulting embeddings. This approach is similarly applied to the bodies of the news articles. Titles and bodies of posts extracted from Reddit are processed in the same manner. Additionally, we compute embeddings of comments.

We first concatenate the various feature vectors derived from general information, news articles, and social media posts to prepare the context features for input into our model. This concatenated feature vector is then normalized to ensure that each feature contributes equally to the final representation. To mitigate the curse of dimensionality and enhance computational efficiency, we apply Principal Component Analysis (PCA)~\cite{abdi2010principal}, reducing the feature space to 100 dimensions. This dimensionality reduction step preserves the most significant variance in the data while discarding redundant and less informative components.

\smallskip\noindent\textit{Transcript Features.} The input audio is transcribed using an automated speech recognition (ASR) system. Specifically, we used the Whisper model developed by Radford et al. \cite{radford2022whisper}\footnote{\url{https://huggingface.co/openai/whisper-large-v2}}. Additionally, we aggregate the transcript features with context features using the previous method for experiments.

\begin{runningexample}[Feature Visualization]

The 2D plot in the figure below shows the structure of the embedding space, created by concatenating context and transcript representations and applying dimensionality reduction via t-SNE~\cite{van2008visualizing}. 

\vspace{0.3cm}
\begin{center}
\includegraphics[width=0.9\linewidth]{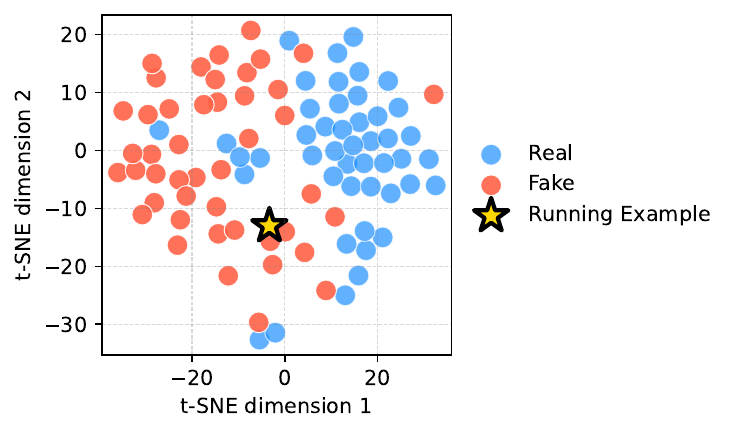}
\end{center}

\vspace{0.3cm}

The Mukesh Ambani deepfake case, our running example, is used to show how our methodology identifies and situates this data point within the broader feature space. The distribution of samples reveals meaningful separations in the data, with the running example positioned near other deepfake samples that share similar transcript and contextual characteristics. 
\end{runningexample}

\paragraph*{Neural Processing}
\label{sec:model_framework}

Our CADD Neural Processing framework integrates contextual information through the architecture illustrated in the right part of Figure 3 of the main manuscript. It comprises three key modules: (1) a context encoder, (2) an audio deepfake detector backbone~\cite{jung2022pushing,wu2018light,afchar2018mesonet,kawa2022specrnet}, and (3) a fusion module and classification head. The framework processes two input streams—context features for the context encoder and raw audio for the RawNet3 backbone or extracted audio features for other backbones, respectively. Their outputs are concatenated and fed into the fusion module, which refines the representation before final classification. Below, we detail each module’s design and function

\smallskip \noindent \textit{Context Encoder.}
The context encoder, $E_{c}$, consists of four blocks, each comprising linear layers and a Leaky Rectified Linear Unit (LeakyReLU) activation layer with a negative slope of 0.1 to mitigate overfitting. To accommodate the dimension of the context features, the input feature dimension for the first block matches the dimension of the extracted context features, and its output dimension is set to 128 for generalizability and robustness, which can be tuned for better performance. For simplicity and to show that context knowledge can improve the performance robustly, the input and output dimensions for the second and third blocks are set to 128 as well. The input dimension remains 128 for the final block, while the output dimensions vary according to the detection backbone: 384 for RawNet3, 128 for LCNN, 6 for MesoNet, and 64 for SpecRNet based on the original baseline implementation. We set these output dimensions to be lower than the feature dimensions of the detection backbones to ensure that the proposed CADD framework prioritizes audio features, with context knowledge serving as supplementary information to support the final prediction.

% if difference explain 
% followed the original papers
\smallskip \noindent \textit{Detection Backbone.}
We selected the detection backbone, $B$, for the two deepfake audio detection architectures. The first is an end-to-end architecture, where the backbone processes raw audio as input and is trained end-to-end. The second type involves a feature extraction model followed by classifiers. Our framework includes four types of architectures. Following~\cite{kawa2023improved}, we incorporate RawNet3~\cite{jung2022pushing} as the end-to-end architecture. For the feature extraction approach, we utilize LCNN~\cite{wu2018light}, MesoNet~\cite{afchar2018mesonet}, and SpecRNet~\cite{kawa2022specrnet}.

RawNet3 is a hybrid architecture that combines elements of ECAPA-TDNN~\cite{desplanques2020ecapa} and RawNet2~\cite{tak2021end}, with added logarithmic and normalization features as part of the backbone. The foundational block of RawNet3 consists of sequential 1D convolutional layers, ReLU activation, and batch normalization layers. The resulting output and the residual output are fed into a max-pooling layer and a feature map scaling module. The feature map scaling module uses global average pooling, a fully connected layer, and a sigmoid layer. LCNN, which has demonstrated high performance on the ASVspoof dataset is a convolutional neural network tailored to capture low-level acoustic features. It includes blocks containing a convolutional layer, a Max-Feature Map (MFM) layer, and a batch normalization layer. Based on the MaxOut activation function, the MFM layer enables the model to focus on task-relevant features. MesoNet, another CNN-based architecture, is structured with blocks that include a convolutional layer followed by ReLU activation, batch normalization, and max pooling. The feature dimension before the classification head for MesoNet is $16$. SpecRNet comprises ResBlock and FMS Attention blocks. The ResBlock includes a convolutional layer, batch normalization, and a Leaky Rectified Linear Unit (LeakyReLU)\cite{xu2015empirical} activation layer. The FMS Attention block, first introduced in RawNet2\cite{jung2020improved}, consists of two max-pooling layers with a feature map scaling module between them. 

% LCNN~\cite{wu2018light}, MesoNet~\cite{afchar2018mesonet}, SpecRNet~\cite{kawa2022specrnet} rawnet2~\cite{tak2021end} {jung2020improved} 
% aasist~\cite{jung2022aasist} wav2vec2.0~\cite{tak2022automatic} (baevski2020wav2vec) rawnet3~\cite{jung2022pushing} whisper~\cite{kawa2023improved} 
% prompt~\cite{oiso2024prompt}

\smallskip \noindent \textit{Fusion Module \& Classification Head.}
Our Fusion Module has two blocks, each containing fully connected layers and a Leaky Rectified Linear Unit (LeakyReLU) activation layer with a negative slope of 0.1 following the original implementation. The input to the fusion module is the concatenated output of the audio deepfake detector backbone and the context encoder, resulting in an input dimension that matches the size of these combined embeddings. The output dimension of the first layer in the fusion module is set to the dimension of the output of the deepfake detection backbone. For simplicity, we then set the input and output dimensions of the second layer the same as the first layer's output dimension. The classification head consists of a fully connected layer. Following~\cite{kawa2023improved}, the feature dimension before reaching the classification head are $3072$, $768$, $16$ and  $128$, for RawNet3, LCNN, MesoNet and SpecRNet, respectively.

\paragraph{Model Training}
Our task is to predict whether the audio is a deepfake or not, making it a binary classification problem. The input to our CADD framework consists of a tuple containing the raw audio and extracted context features, $(x_{a}, x_{c})$. The deepfake audio detection backbone processes the raw audio, $x_{a}$, and outputs audio features $B(x_{a})$. Simultaneously, the context encoder takes the context features, $x_{c}$, as input and produces context features $E_c(x_{c})$. The resulting audio and context features are concatenated and passed through the fusion and classification module, $f_{cls}$. Since this is a binary classification task, we apply binary cross-entropy loss, formulated as follows:
\begin{equation}
\label{equ:le}
\mathcal{L} = -{(y \cdot log(p) + (1 - y)log(1 - p))}
\end{equation}
where $y$ in Equation \ref{equ:le} refers to the label of the input audio and $p=f_{cls}(Concate(E_c(x_{c}), B(x_{a})))$.

We follow the settings from the neural network-based methods from~\cite{kawa2023improved} for the optimizer, learning rate, and weight decay. All models were trained with the Adam optimizer. The learning rate and weight decay for RawNet3 are set to 0.001 and $5e-5$, respectively. For LCNN, MesoNet, and SpecRNet, the learning rate is set to 0.0001, and weight decay is set to 0.0001. We used a batch size of 16 and performed a grid search over training epochs, testing 10, 20, and 30 epochs, and found that the models converge at 30 epochs. Each experiment is run with random seeds 0, 1, and 2, and we report the average results. The Whisper encoder was kept frozen during training. All experiments are conducted on three NVIDIA RTX A6000 GPUs.

\begin{runningexample}[CADD Predictions]
We analyze the predictions provided by the four best-performing SOTA baselines (i.e. RawNet3, LCNN with Whisper features, MesoNet with Whisper features, and SpecRNet with Whisper features) on the Mukesh Ambani example. The table below shows the probability of the Mukesh Ambani audio being a deepfake as predicted by the different SOTA baselines and related CADD configurations:

\vspace{0.3cm}
\begin{center}
\small
\begin{tabular}{@{}lcccc@{}}
\toprule
\textbf{Backbone} &
  \textbf{Baseline} & \textbf{CADD(T)} & \textbf{CADD(C)} & \textbf{CADD(T+C)} \\ \midrule
RawNet3 & 19.39 & 90.47 & 97.63 & 97.04 \\
LCNN (with Whisper features) & 51.61 & 82.34 & 98.89 & 99.53 \\
MesoNet (with Whisper features) & 38.23 & 61.03 & 70.28 & 68.98 \\
SpecRNet (with Whisper features) & 14.58 & 0.78 & 33.58 & 35.04 \\
\bottomrule
\end{tabular}
\end{center}
\vspace{0.3cm}

Each baseline model identifies the sample as authentic in most cases, with LCNN being the only exception at 51.61\% probability of deepfake detection. When our CADD framework is applied, detection accuracy improves substantially across nearly all backbones. Most impressive is the {CADD(T+C)} implementation---integrating both transcript and contextual information---which achieves 99.53\% confidence in correctly identifying the deepfake when trained on the LCNN backbone with Whisper features. \newline

To address potential concerns about methodological bias, specifically that CADD's performance might depend heavily on the presence of a specific news article about the deepfake within the contextual information, we conducted an additional experiment. After removing the targeted news article about the financial fraud from the context, our analysis (shown below) reveals that the CADD framework maintains robust detection capability, with the CADD(T+C) approach still achieving 99.31\% confidence with the LCNN backbone. This confirms that our approach leverages diverse contextual signals rather than relying on a single explicitly fact-checked source.

\vspace{0.3cm}
\begin{center}
\small
\begin{tabular}{@{}lcccc@{}}
\toprule
\textbf{Backbone} &
  \textbf{Baseline} & \textbf{CADD(T)} & \textbf{CADD(C)} & \textbf{CADD(T+C)} \\ \midrule
RawNet3 & 19.39 & 92.75 & 96.40 & 96.31 \\
LCNN (with Whisper features) & 51.61 & 78.37 & 98.27 & 99.31 \\
MesoNet (with Whisper features) & 38.23 & 59.91 & 70.62 & 68.96 \\
SpecRNet (with Whisper features) & 14.58 & 0.74 & 30.63 & 33.10 \\
\bottomrule
\end{tabular}
\end{center}
\vspace{0.2cm}
\end{runningexample}

%%%%%%%%%%%%%%%% SUPPLEMENTARY TEXT %%%%%%%%%%%%%%%

\subsubsection*{Baselines}
Our extensive experiments looked at our novel CADD pipeline as well as several baselines. 

\paragraph{Baselines based on Recent Audio Deepfake Detectors} Four state-of-the-art (SOTA) audio deepfake detectors were tested as baselines: RawNet3~\cite{jung2022pushing}, LCNN~\cite{wu2018light}, MesoNet~\cite{afchar2018mesonet}, and SpecRNet~\cite{kawa2022specrnet}. Since RawNet3 operates directly on raw audio, it was used without additional feature extraction. The other three detectors were evaluated with three feature extractors: linear frequency cepstral coefficients (LFCC), mel-frequency cepstral coefficients (MFCC), and the state-of-the-art Whisper features~\cite{kawa2023improved}.  This resulted in a total of 22 baselines: one RawNet3 model and 21 configurations derived from the three detectors combined with 7 different non-empty subsets of LFCC, MFCC, and Whisper features.  

\paragraph{Baselines based on Traditional Machine Learning} We also fed the 7 above-mentioned LFCC, MFCC and Whisper feature subsets into 7 widely-used traditional machine learning (TML) models: Logistic Regression, Random Forest, SVM, AdaBoost, XGBoost, Gaussian Naive Bayes, K-Nearest Neighbors. This led to a total of 49 TML baselines.

Thus, we have a total of 71 baselines (SOTA plus TML) which were tested on the three datasets: JDD, SYN, and ITW. We additionally studied how three versions of our CADD framework can build on these 71 baselines and improve them: CADD(T) which uses transcripts, CADD(C) which uses context, and CADD(T+C) which combines both transcripts and context. In all, 284 classification models were tested to assess predictive quality.

\subsection*{Supplementary Text}
\subsubsection*{Mapping between subject categories and occupations}
\label{sec:appendix_mapping}
We have classified occupations of the public figures in our Journalist-provided Deepfake Dataset (JDD) and Synthetic Audio Dataset (SYN) into several categories. The details of this categorization are outlined below:

\begin{itemize}
    \item \textit{Entertainment}: actor, television actor, film actor, film director, film producer, stage actor, voice actor, stand-up comedian, comedian
    \item \textit{Politics}: politician, diplomat, monarch, statesperson, lobbyist, First Lady
    \item \textit{Music}: singer, singer-songwriter, guitarist, drummer, rapper, composer, opera singer, dancer
    \item \textit{Media}: radio personality, YouTuber, journalist, opinion journalist, pundit, sports journalist, television presenter, television producer, publisher
    \item \textit{Sports}: association football player, cricketer, Formula One driver, American football player, basketball player, mixed martial arts fighter, ice hockey player, professional wrestler, tennis player
    \item \textit{Writing}: poet, writer, screenwriter, novelist
    \item \textit{Fashion}: model, fashion model, fashion designer, beauty pageant contestant
    \item \textit{Business}: entrepreneur, business magnate, economist, businessperson, business executive
    \item \textit{Law}: lawyer, barrister, jurist, judge
    \item \textit{Academia \& Research}: academic, researcher, teacher
    \item \textit{Activism}: disability rights activist, activist, environmentalist, civil rights advocate
    \item \textit{Other}: all the other occupations that do not fit in any of the categories above.
\end{itemize}

\subsubsection*{Prompt Template for Fake Transcript Generation}
\label{sec:appendix_syn_prompt}
The template used to create prompts that generate fake transcripts for our Synthetic Audio Dataset (SYN) dataset without taking contextual information into account is provided as follows:

\begin{tcolorbox}[
    % enhanced,
    colback=blue!5!white, 
    colframe=black, 
    width=\textwidth,
    coltitle=white,
    title=\textbf{Prompt Template for Fake Transcripts Generation, without Context},
    top=3mm, bottom=2mm,
    % attach boxed title to top left={yshift=-2.8mm, xshift=4mm},
    % boxed title style={colback=black, colframe=black, boxrule=0mm, 
    % toptitle=1mm, bottomtitle=1mm}
]

Write something that {\color{blue}\textit{Subject's Name}} would definitely {\color{blue}\textit{(not)}} say. Format your output as a json with the field "text".
\end{tcolorbox}

Here, {\color{blue}\textit{Subject's Name}} is replaced with the name of the public figure to which the fake transcript refers, and {\color{blue}\textit{not}} is included when we aim to generate content that the person would likely not say.

Before generating context-aware fake transcripts, we first collect relevant news articles for the specified date, with a maximum limit of 10 articles. We then construct the prompt as follows:

\begin{tcolorbox}[
    % enhanced,
    colback=blue!5!white, 
    colframe=black, 
    width=\textwidth,
    coltitle=white,
    title=\textbf{Prompt Template for Fake Transcripts Generation, with Context},
    top=3mm, bottom=2mm,
    % attach boxed title to top left={yshift=-2.8mm, xshift=4mm},
    % boxed title style={colback=black, colframe=black, boxrule=0mm, 
    % toptitle=1mm, bottomtitle=1mm}
]
Please read the following list of news titles along with their corresponding summaries:\newline 

ARTICLE 1\newline
Title: {\color{blue}\textit{news title}}\newline
Summary: {\color{blue}\textit{news summary}}

...

ARTICLE {\color{blue}\textit{N}}\newline
Title: {\color{blue}\textit{news title}}\newline
Summary: {\color{blue}\textit{news summary}} \newline 

Given this context, write something that {\color{blue}\textit{Subject's Name}} would definitely {\color{blue}\textit{(not)}} say. Format your output as a JSON with the field "text".
\end{tcolorbox}

\subsubsection*{Audio Manipulations}
The audio manipulations used in our experiments are detailed below.

\paragraph{Air Absorption} Text-to-Speech (TTS) and voice-cloning systems are typically trained on high-quality audio datasets, such as audiobooks, resulting in generated audio that resembles studio-quality recordings. These recordings often sound as if the speaker is positioned close to a microphone in a controlled environment. Air absorption simulates the natural attenuation of sound as it travels through air, including frequency-dependent attenuation effects. Adversarial users may exploit this by altering audio to mimic scenarios where the speaker is at varying distances from the microphone, thereby modifying its acoustic properties to evade detection. In our experiments, we simulated distances of 10, 20, 50 and 100 meters.

\paragraph{Background Noise} Real-world audio recordings often contain background noise, such as breathing, clock ticks, footsteps, or ambient sounds like rain and wind. These noises can enhance the perceived authenticity of the recording. Malicious actors may exploit this by deliberately adding or simulating background noise to align synthetic audio with real-world environmental characteristics, making it more challenging to distinguish as artificially generated. Following Wu et al. \cite{DBLP:journals/corr/abs-2404-15854}, we have evaluated perturbations including rain, wind, breathing, clock ticks, footsteps, sneezing, and coughing.

\paragraph{Gaussian Noise} Gaussian noise, a common type of statistical noise characterized by a normal distribution, frequently appears in real-world audio recordings due to electronic devices or environmental interference. While TTS and voice-cloning methods typically generate clean, noise-free audio, real-world recordings are rarely perfect. To mimic these imperfections and evade detection, malicious actors may introduce Gaussian noise into their synthetic audio. We have studied results obtained by applying a noise amplification factor randomly chosen between the range [0.001, $x$], with $x$ being set either to 0.05, 0.1, 0.15 or 0.2.

\paragraph{MP3 Compression} MP3 compression is widely used for audio storage and transmission due to its ability to significantly reduce file sizes while maintaining acceptable audio quality, a common scenario for social media or messaging platforms. However, the compression process introduces artifacts and removes subtle details, altering the original audio's spectral characteristics. Since TTS and voice-cloning methods often produce high-quality, uncompressed audio, malicious actors may apply MP3 compression to their synthetic audio to mimic the imperfections commonly found in real-world recordings. We present results with compression to a bitrate of 8, 16, 32 and 64 kbps. 

\paragraph{Time Stretch} Time-stretching is a technique used to alter the duration of an audio signal without changing its pitch. This effect can modify the playback speed, either accelerating or slowing it down, and is commonly used in applications like speech editing. While TTS and voice-cloning systems generate audio at a natural tempo, malicious actors may apply time-stretching to synthetic audio to distort its rhythm or pacing. Such manipulation can make the audio less detectable by algorithms that are tuned to recognize specific temporal patterns. In our experiments, we have applied slowdowns at a rate of 0.6 and 0.8, and speed-ups at rates of 1.2 and 1.4.

\subsubsection*{Running Example (Additional Material)}

\paragraph{News Articles}

The news articles collected for the Mukesh Ambani's running example are listed below:

\begin{itemize}
    \item \href{https://dailytimes.com.pk/1203195/ambanis-served-us-gold-with-roti-at-pre-wedding-sara-ali-khan/}{Ambanis served us gold with roti at pre-wedding: Sara Ali Khan (DailyTimes, 2024-06-23)}
    \item \href{https://indianexpress.com/article/india/gautam-adani-drew-rs-9-26-crore-salary-in-fy24-lower-than-his-executives-industry-peers-9410503/}{Gautam Adani drew Rs 9.26 crore salary in FY24 --- lower than his executives, industry peers (IndianExpress, 2024-06-23)}
    \item \href{https://thenorthlines.com/gautam-adani-drew-rs-9-26-cr-salary-in-fy24-lower-than-his-executives-industry-peers/}{Gautam Adani drew Rs 9.26 cr salary in FY24 \--- lower than his executives, industry peers (The North Lines, 2024-06-23)}
    \item \href{https://timesofindia.indiatimes.com/business/india-business/gautam-adanis-salary-was-less-than-his-peers-and-executives-in-fy24-heres-how-much-he-got/articleshow/111206059.cms}{Gautam Adani's salary was less than his peers and executives in FY24. Here's how much he got (Times of India, 2024-06-23)}
    \item \href{https://timesofindia.indiatimes.com/life-style/fashion/celeb-style/isha-ambanis-unforgettable-fashion-moments-at-anant-and-radhikas-pre-wedding-cruise-bash/articleshow/111203561.cms}{Isha Ambani's unforgettable fashion moments at Anant and Radhika's pre-wedding cruise bash (Times of India, 2024-06-23)}
    \item \href{https://www.business-standard.com/companies/news/adani-draws-rs-9-26-cr-salary-in-fy24-lower-than-his-executives-peers-124062300114_1.html}{Adani draws Rs 9.26 cr salary in FY24, lower than his executives, peers (Business Standard, 2024-06-23)}
    \item \href{https://thenorthlines.com/sara-ali-khan-jokes-about-gold-served-with-roti-at-anant-ambanis-pre-wedding-celebrations/}{Sara Ali Khan jokes about gold served with roti at Anant Ambani's pre-wedding celebrations (The North Lines, 2024-06-22)}
    \item \href{https://www.dnaindia.com/mumbai/report-mukesh-ambani-deepfake-video-mumbai-doctor-falls-prey-to-frauds-loses-rs-7-lakh-in-3094299}{Mukesh Ambani deepfake video: Mumbai doctor falls prey to frauds, loses Rs 7 lakh in... (DNA India, 2024-06-22)}
    \item \href{https://www.business-standard.com/companies/news/16-ril-shareholders-oppose-reappointment-of-aramco-chairman-as-director-124062200208_1.html}{16\% RIL shareholders oppose reappointment of Aramco chairman as director (Business Standard (2024-06-22)}
\end{itemize}

\paragraph{Reddit Posts} The Reddit posts collected for the Mukesh Ambani's running example are listed below:

\begin{itemize}
    \item \href{https://www.reddit.com/r/anantambani/comments/1dk5s8p/how_does_anant_ambanis_approach_to_business/}{How does Anant Ambani's approach to business differ from that of his father, Mukesh Ambani? (2024-06-20)}
    \item \href{https://www.reddit.com/r/IndianModerate/comments/1djes3w/nita_ambani_worked_after_her_marriage_to_mukesh/}{Nita Ambani worked after her marriage to Mukesh Ambani. She earned this much (2024-06-19)}
    \item \href{https://i.redd.it/4fsu10ssuy5d1.png}{Meet Mukesh Ambani's niece, Nayantara Kothari (2024-06-11)}
    \item \href{https://www.reddit.com/r/Maharashtra/comments/1d99962/gujju_mukesh_ambani_to_invest_13400_crores_to_buy/}{Gujju Mukesh Ambani to invest 13,400 crores to buy land in Navi Mumbai... (2024-06-05)}
    \item \href{https://www.reddit.com/r/india/comments/1d6a42b/gautam_adani_replaces_mukesh_ambani_as_richest/}{[Business] - Gautam Adani beats Mukesh Ambani to become richest Asian | Times of India (2024-06-02)}
    \item \href{https://www.reddit.com/r/TIMESINDIAauto/comments/1d6p5l4/business_gautam_adani_beats_mukesh_ambani_to/}{[Business] - Gautam Adani beats Mukesh Ambani to become richest Asian (2024-06-02)}
    \item \href{https://www.reddit.com/r/indianews/comments/1d6a35f/gautam_adani_replaces_mukesh_ambani_as_richest/}{Gautam Adani replaces Mukesh Ambani as richest Asian (2024-06-02)}
    \item \href{https://www.reddit.com/r/updatenewdailyweekly/comments/1d1ja2i/rils_mukesh_ambani_set_for_african_safari_with_5g/}{RIL's Mukesh Ambani set for African safari with 5G tech solutions (2024-05-26)}
\end{itemize}

% If your supplement is very short you might need to uncomment the following line to avoid
% layout problems with the figures and tables.
\newpage

%%%%%%%%%%%%%%%% SUPPLEMENTARY FIGURES %%%%%%%%%%%%%%%

\begin{figure}[h!]
    \centering
    \subfigure[CADD vs. State-of-the-art (SOTA) baselines]{
        \includegraphics[width=\linewidth]{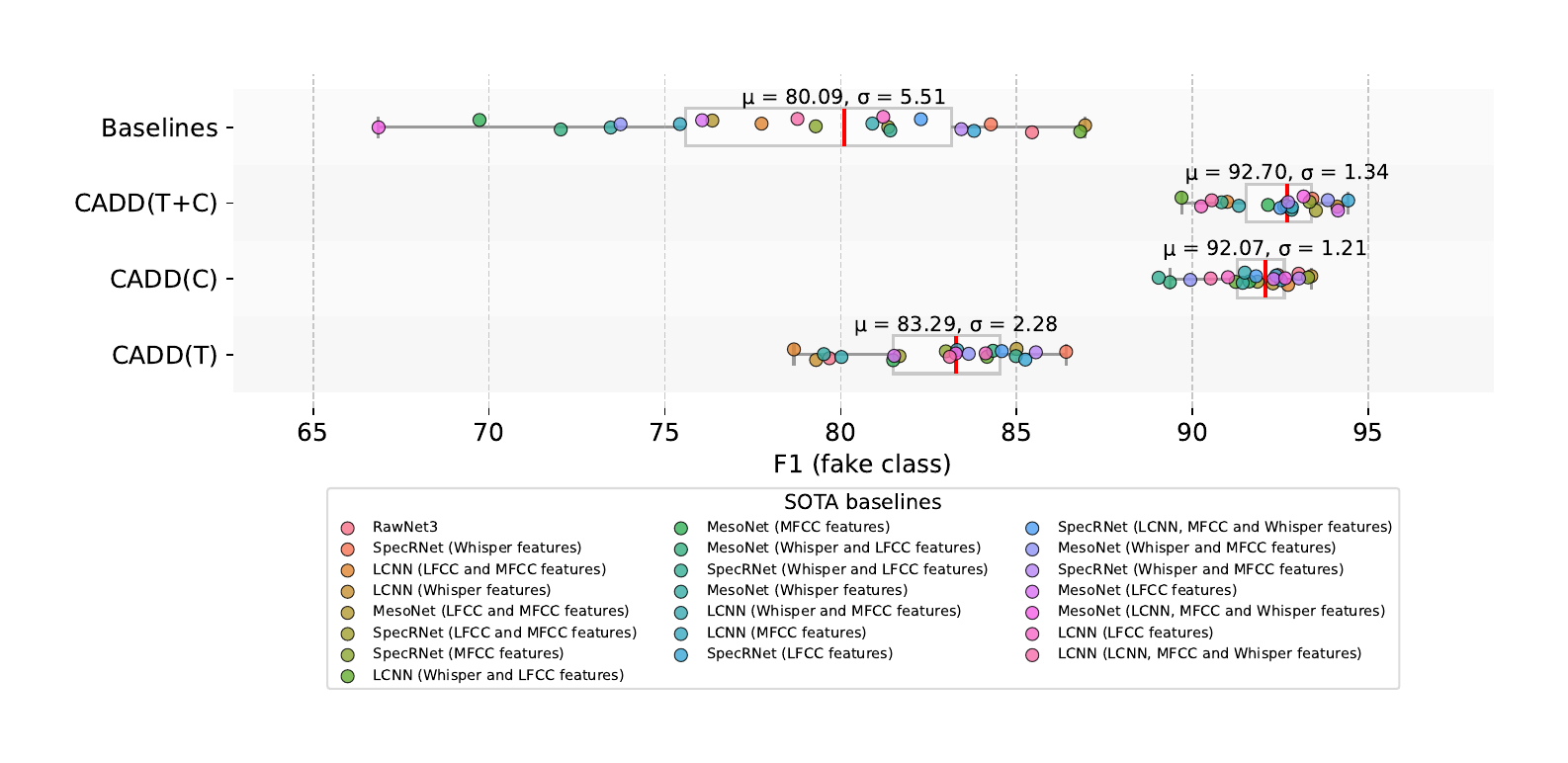}
        \label{fig:sota-baselines-F1_JDD}
    }
    \subfigure[CADD vs. Traditional Machine Learning (TML) baselines]{
        \includegraphics[width=\linewidth]{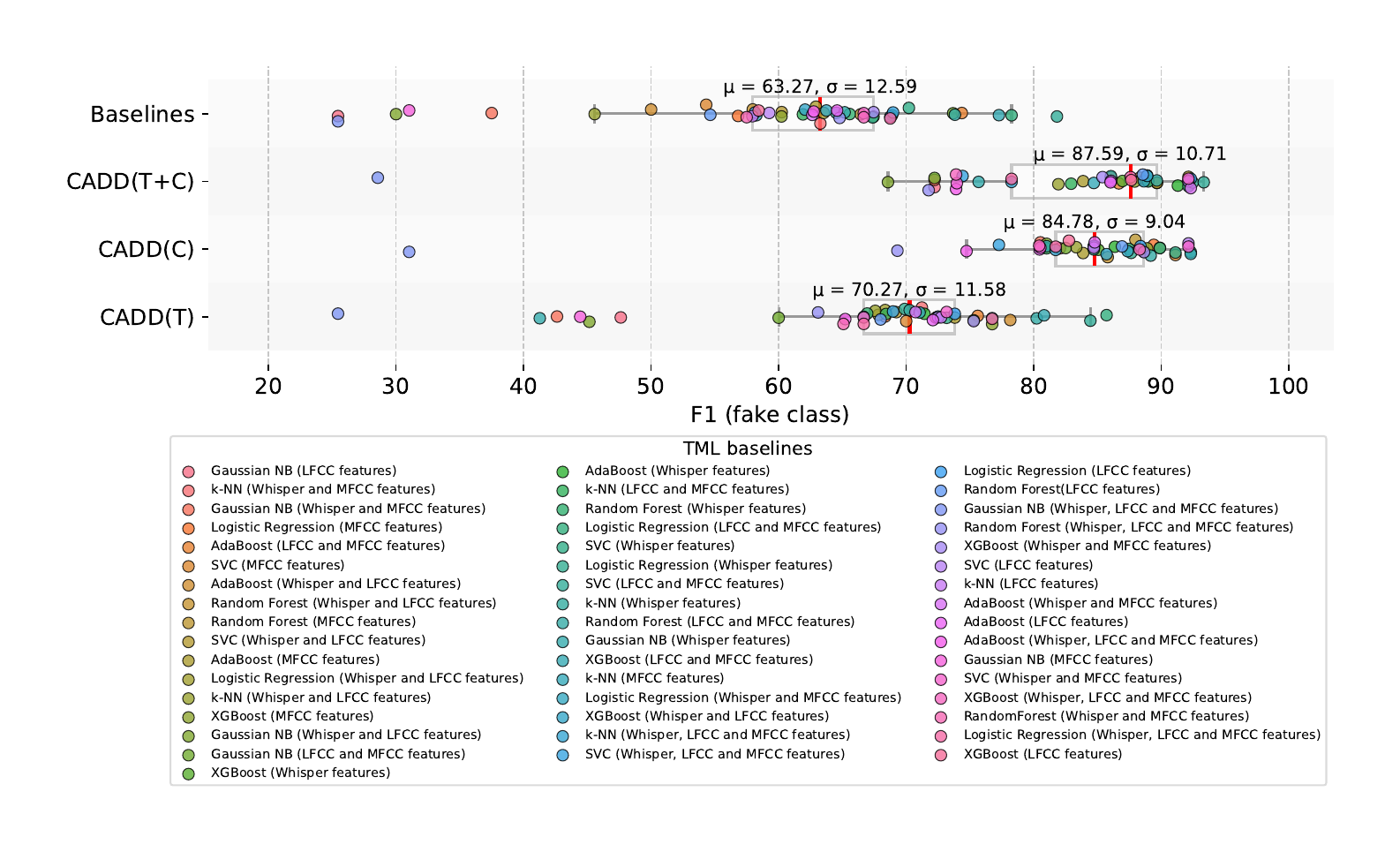}
        \label{fig:tml-baselines-F1_JDD}
    }
    \caption{\textbf{Performance comparison of state-of-the-art (a) and traditional machine learning (b) baselines and our CADD configurations (CADD(T), CADD(C), and CADD(T+C)).} Each point represents a model's F1 score computed on our Journalist-provided Deepfake Dataset (JDD), with the same color denoting the same baseline across different configurations.}
    \label{fig:combined-figures-F1_JDD}
\end{figure}

\begin{figure}[h!]
    \centering
    \subfigure[CADD vs. State-of-the-art (SOTA) baselines]{
        \includegraphics[width=\linewidth]{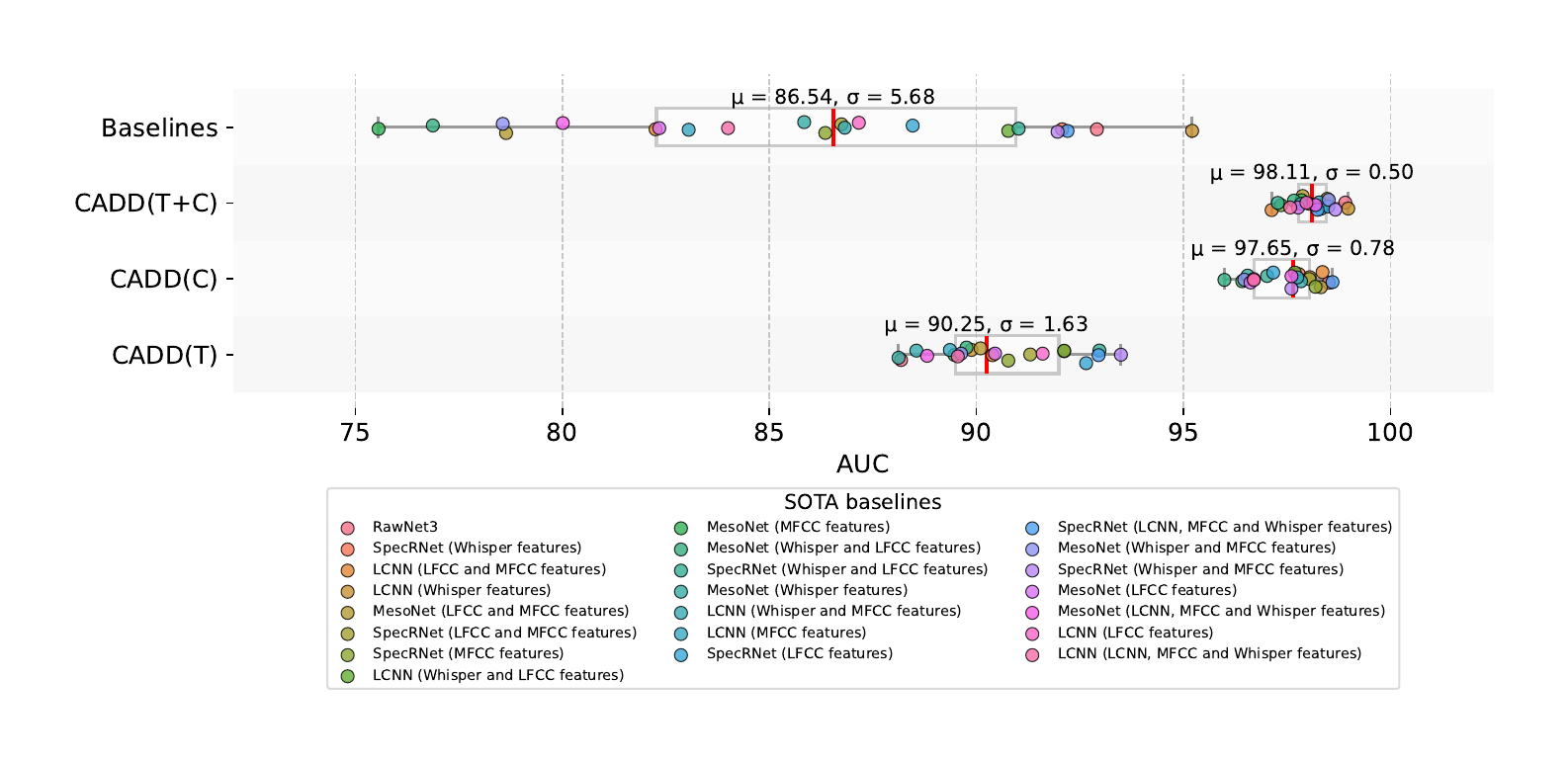}
        \label{fig:sota-baselines-AUC_JDD}
    }
    \subfigure[CADD vs. Traditional Machine Learning (TML) baselines]{
        \includegraphics[width=\linewidth]{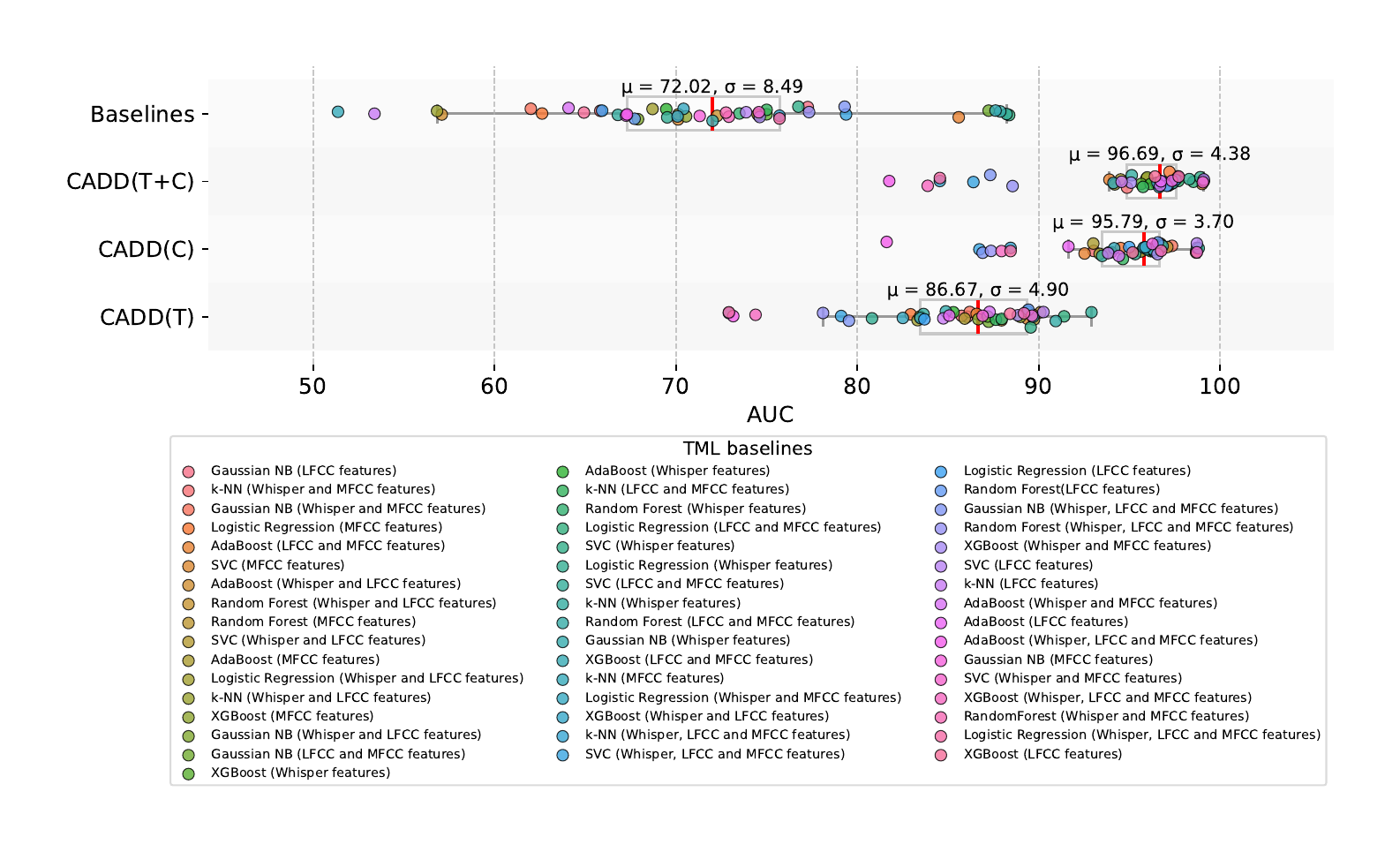}
        \label{fig:tml-baselines-AUC_JDD}
    }
    \caption{\textbf{Performance comparison of state-of-the-art (a) and traditional machine learning (b) baselines and our CADD configurations (CADD(T), CADD(C), and CADD(T+C)).} Each point represents a model's AUC score computed on our Journalist-provided Deepfake Dataset (JDD), with the same color denoting the same baseline across different configurations.}
    \label{fig:combined-figures-AUC_JDD}
\end{figure}

\begin{figure}[h!]
    \centering
    \subfigure[CADD vs. State-of-the-art (SOTA) baselines]{
        \includegraphics[width=\linewidth]{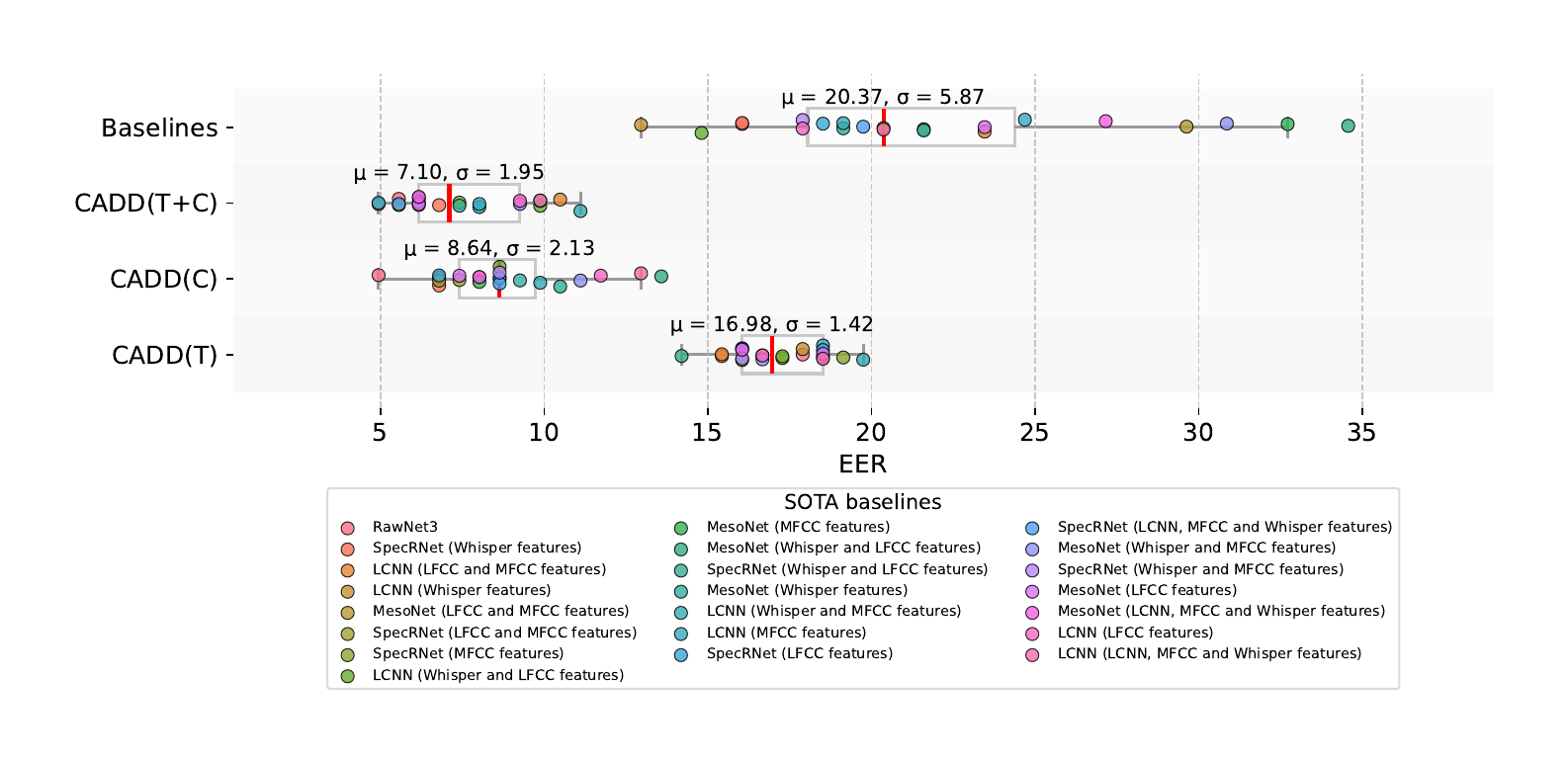}
        \label{fig:sota-baselines-EER_JDD}
    }
    \subfigure[CADD vs. Traditional Machine Learning (TML) baselines]{
        \includegraphics[width=\linewidth]{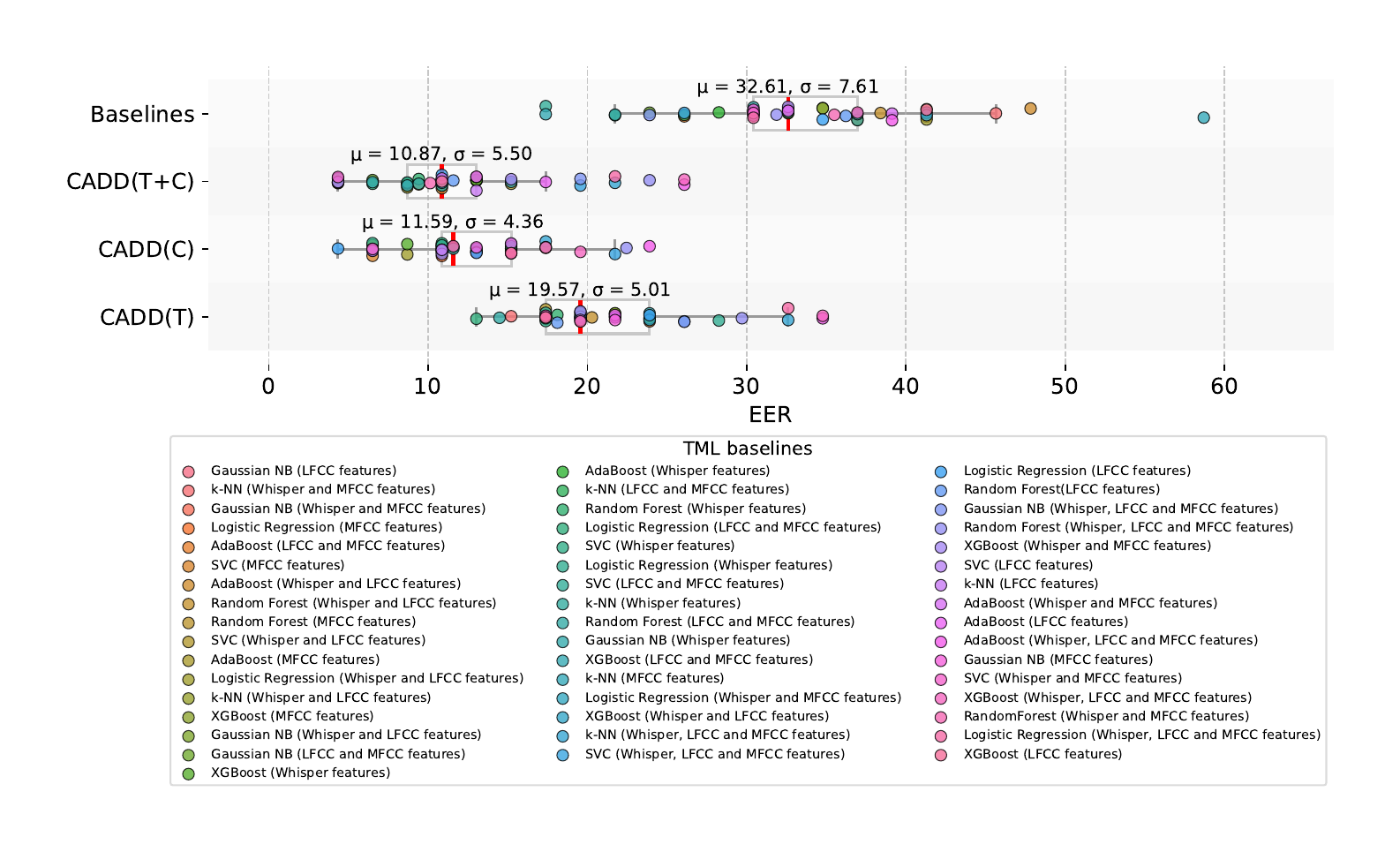}
        \label{fig:tml-baselines-EER_JDD}
    }
    \caption{\textbf{Performance comparison of state-of-the-art (a) and traditional machine learning (b) baselines and our CADD configurations (CADD(T), CADD(C), and CADD(T+C)).} Each point represents a model's EER score computed on our Journalist-provided Deepfake Dataset (JDD), with the same color denoting the same baseline across different configurations.}
    \label{fig:combined-figuress-EER_JDD}
\end{figure}

\begin{figure}[h!]
    \centering
    \subfigure[CADD vs. State-of-the-art (SOTA) baselines]{
        \includegraphics[width=\linewidth]{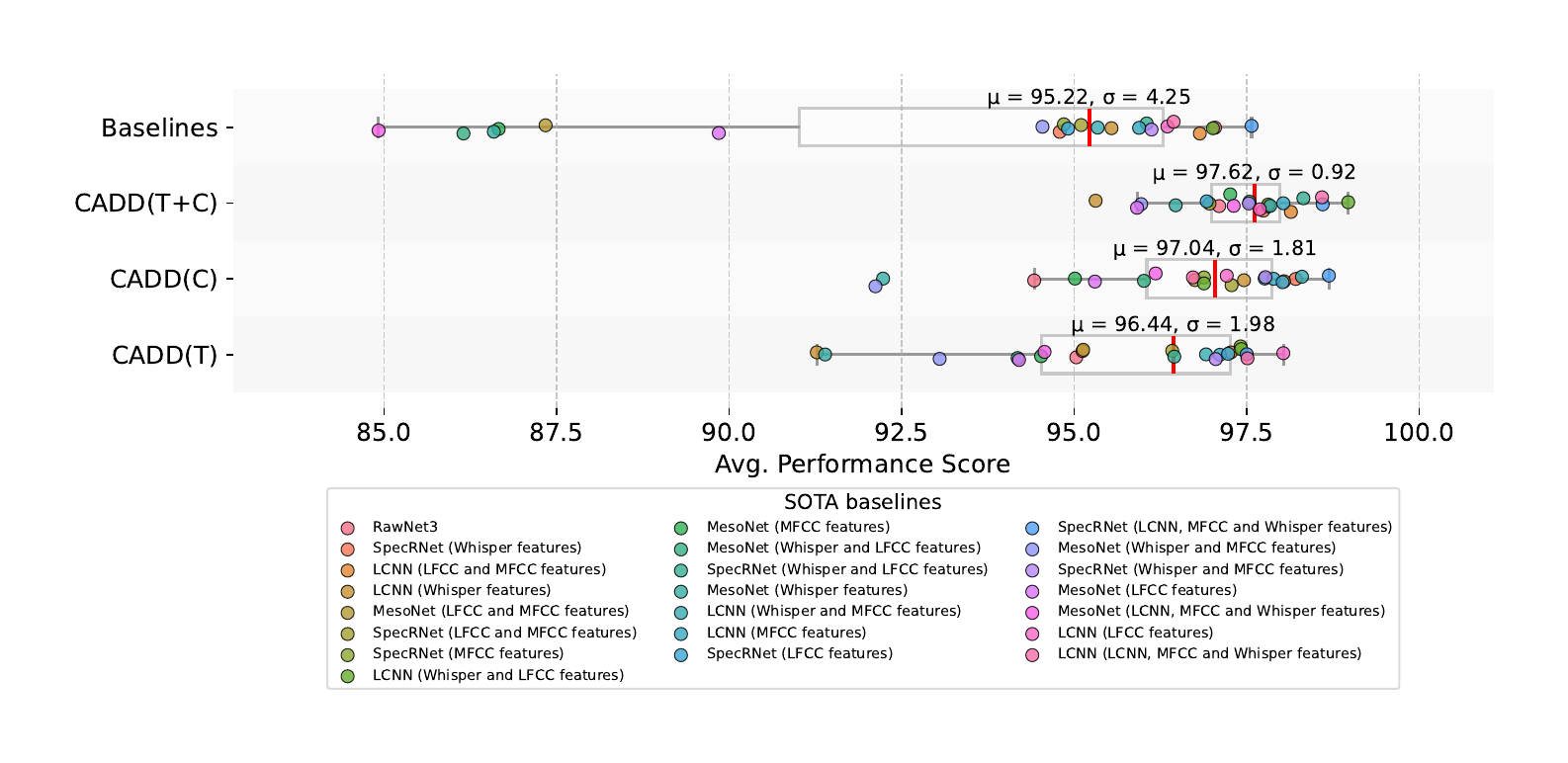}
        \label{fig:sota-baselines-Avg_SYN}
    }
    \subfigure[CADD vs. Traditional Machine Learning (TML) baselines]{
        \includegraphics[width=\linewidth]{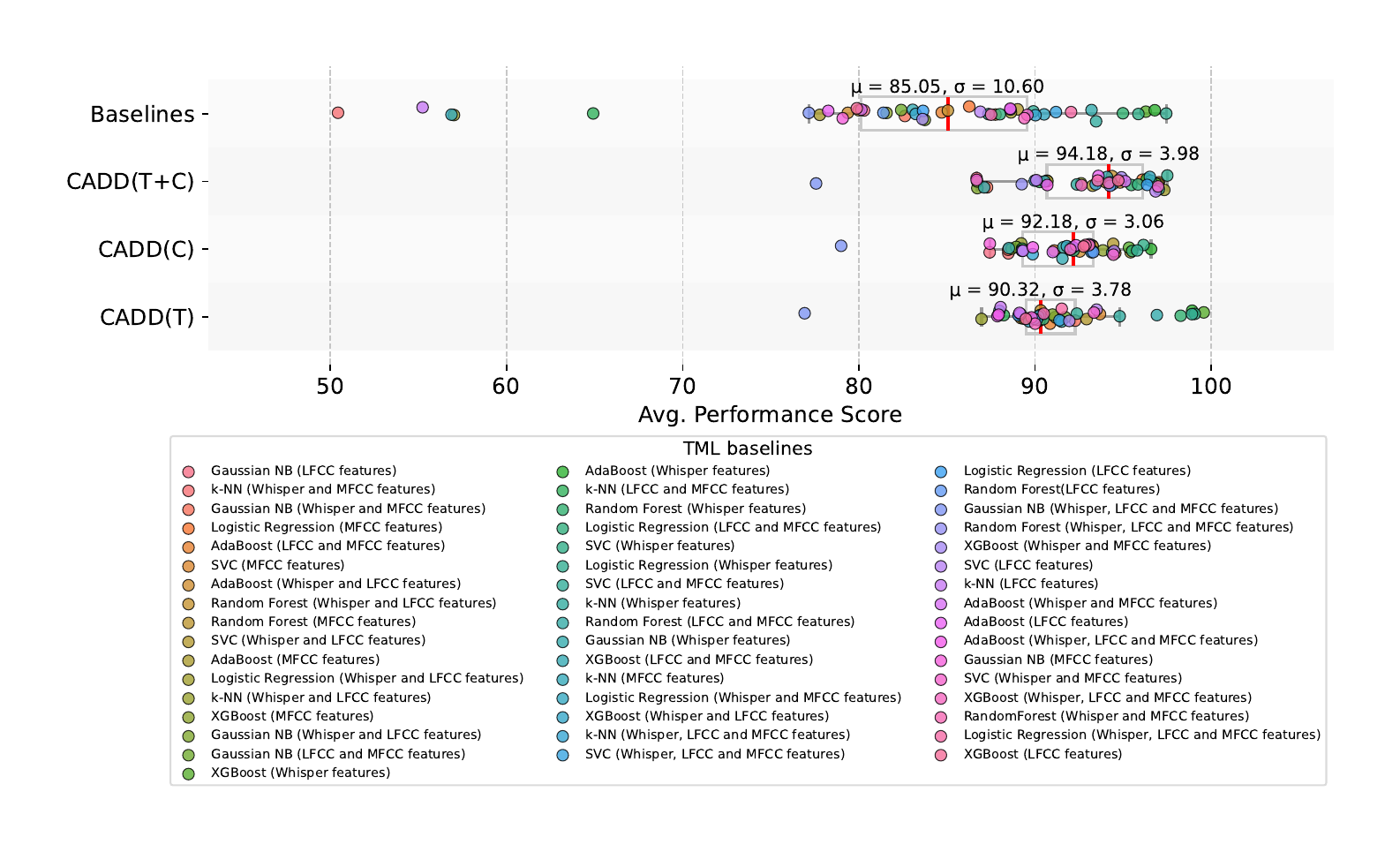}
        \label{fig:tml-baselines-Avg_SYN}
    }
    \caption{\textbf{Performance comparison of state-of-the-art (a) and traditional machine learning (b) baselines and our CADD configurations (CADD(T), CADD(C), and CADD(T+C)).} Each point represents a model's Avg score computed on our synthetic dataset (SYN), with the same color denoting the same baseline across different configurations.}
    \label{fig:combined-figures-Avg_SYN}
\end{figure}

\begin{figure}[h!]
    \centering
    \subfigure[CADD vs. State-of-the-art (SOTA) baselines]{
        \includegraphics[width=\linewidth]{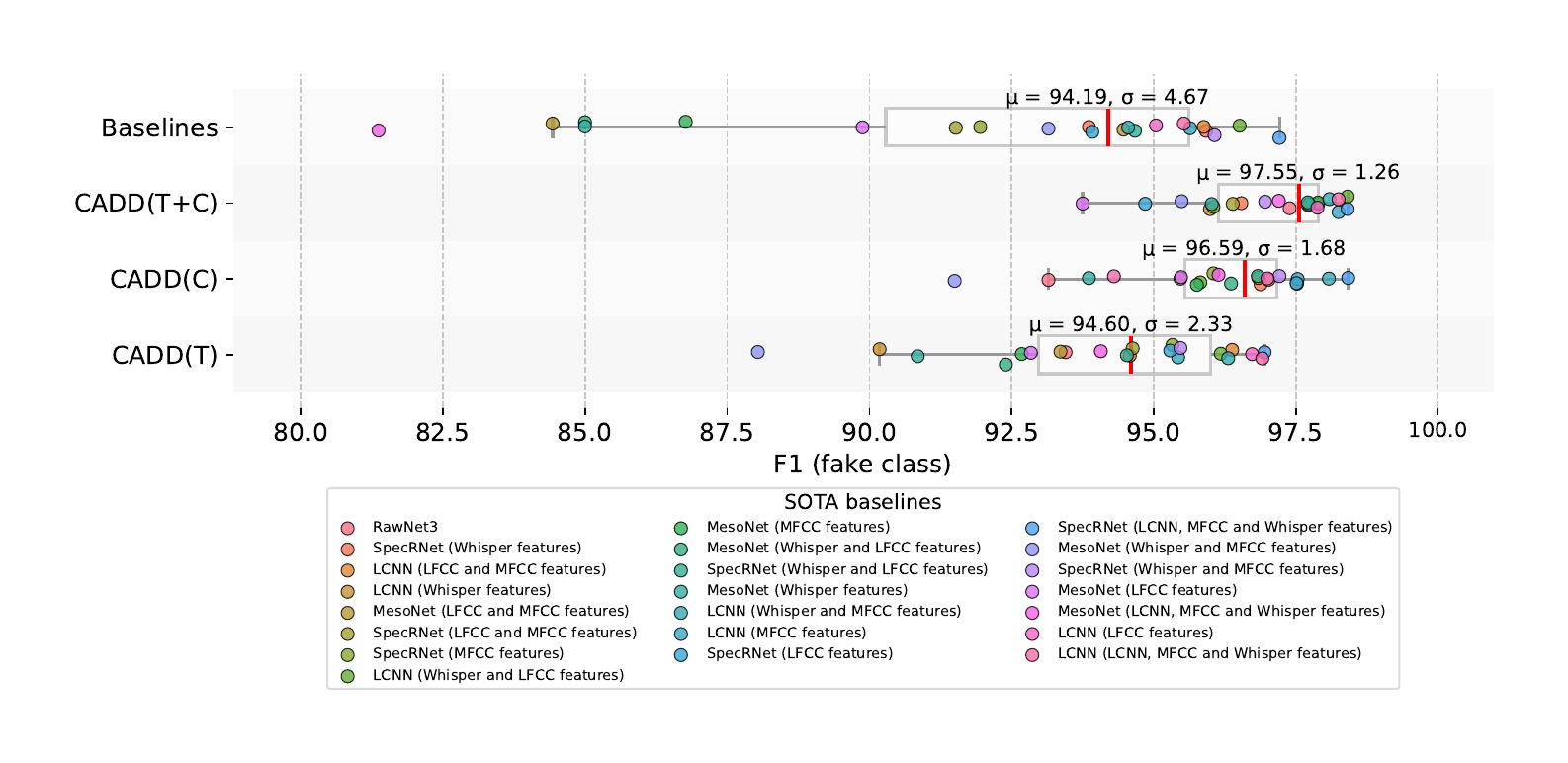}
        \label{fig:sota-baselines-F1_SYN}
    }
    \subfigure[CADD vs. Traditional Machine Learning (TML) baselines]{
        \includegraphics[width=\linewidth]{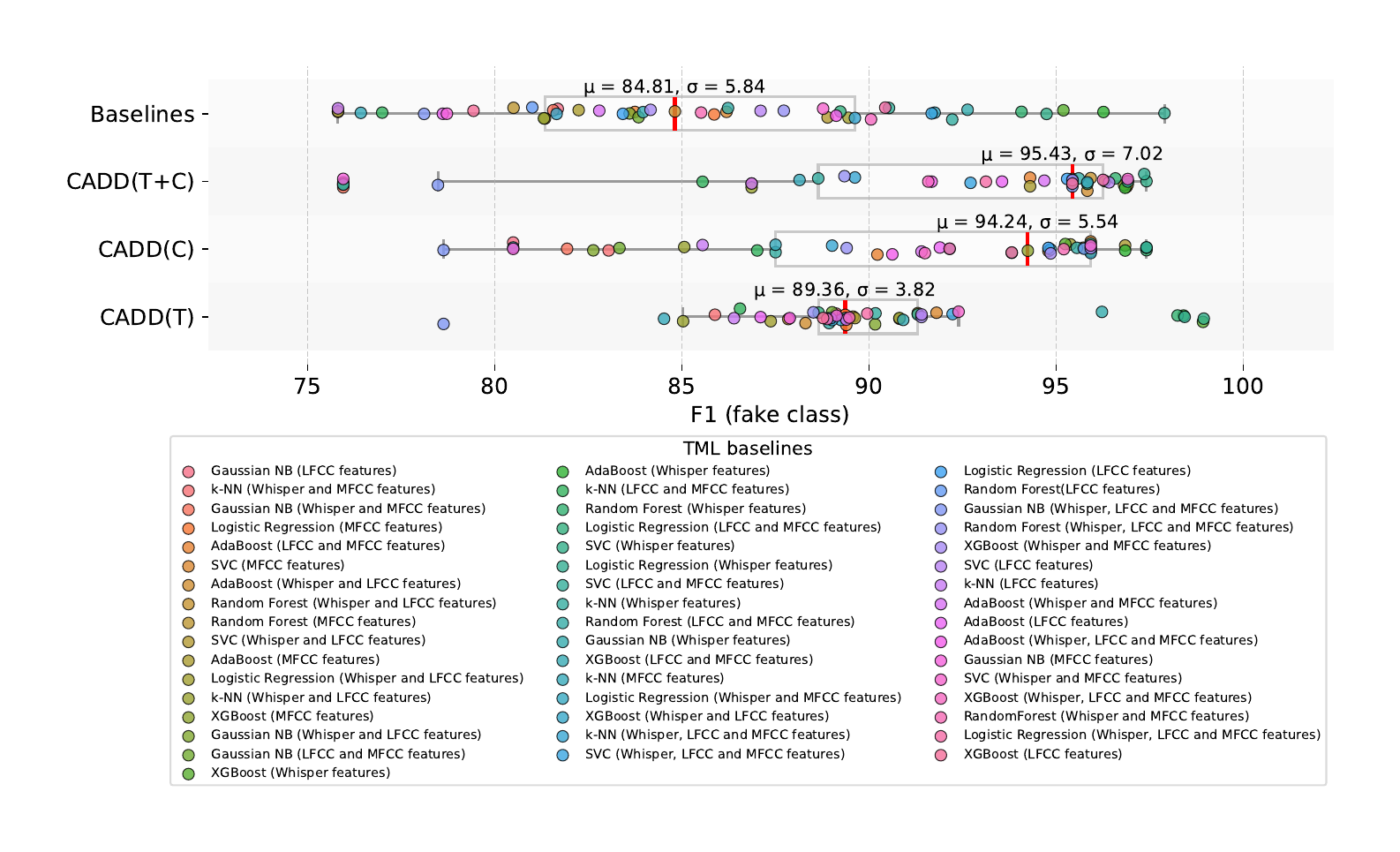}
        \label{fig:tml-baselines-F1_SYN}
    }
    \caption{\textbf{Performance comparison of state-of-the-art (a) and traditional machine learning (b) baselines and our CADD configurations (CADD(T), CADD(C), and CADD(T+C)).} Each point represents a model's F1 score computed on our synthetic dataset (SYN), with the same color denoting the same baseline across different configurations. }
    \label{fig:combined-figures-F1_SYN}
\end{figure}

\begin{figure}[h!]
    \centering
    \subfigure[CADD vs. State-of-the-art (SOTA) baselines]{
        \includegraphics[width=\linewidth]{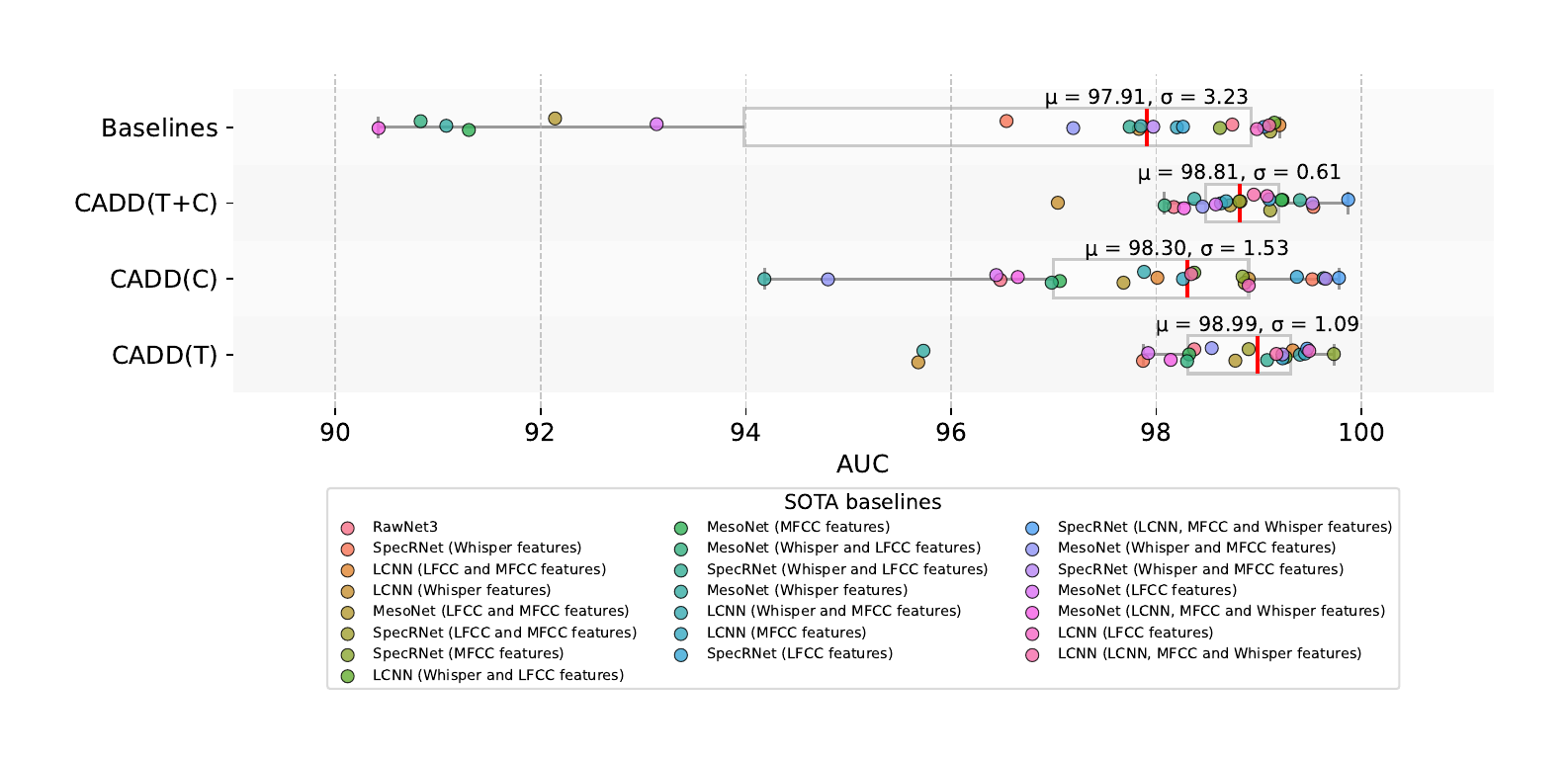}
        \label{fig:sota-baselines-AUC_SYN}
    }
    \subfigure[CADD vs. Traditional Machine Learning (TML) baselines]{
        \includegraphics[width=\linewidth]{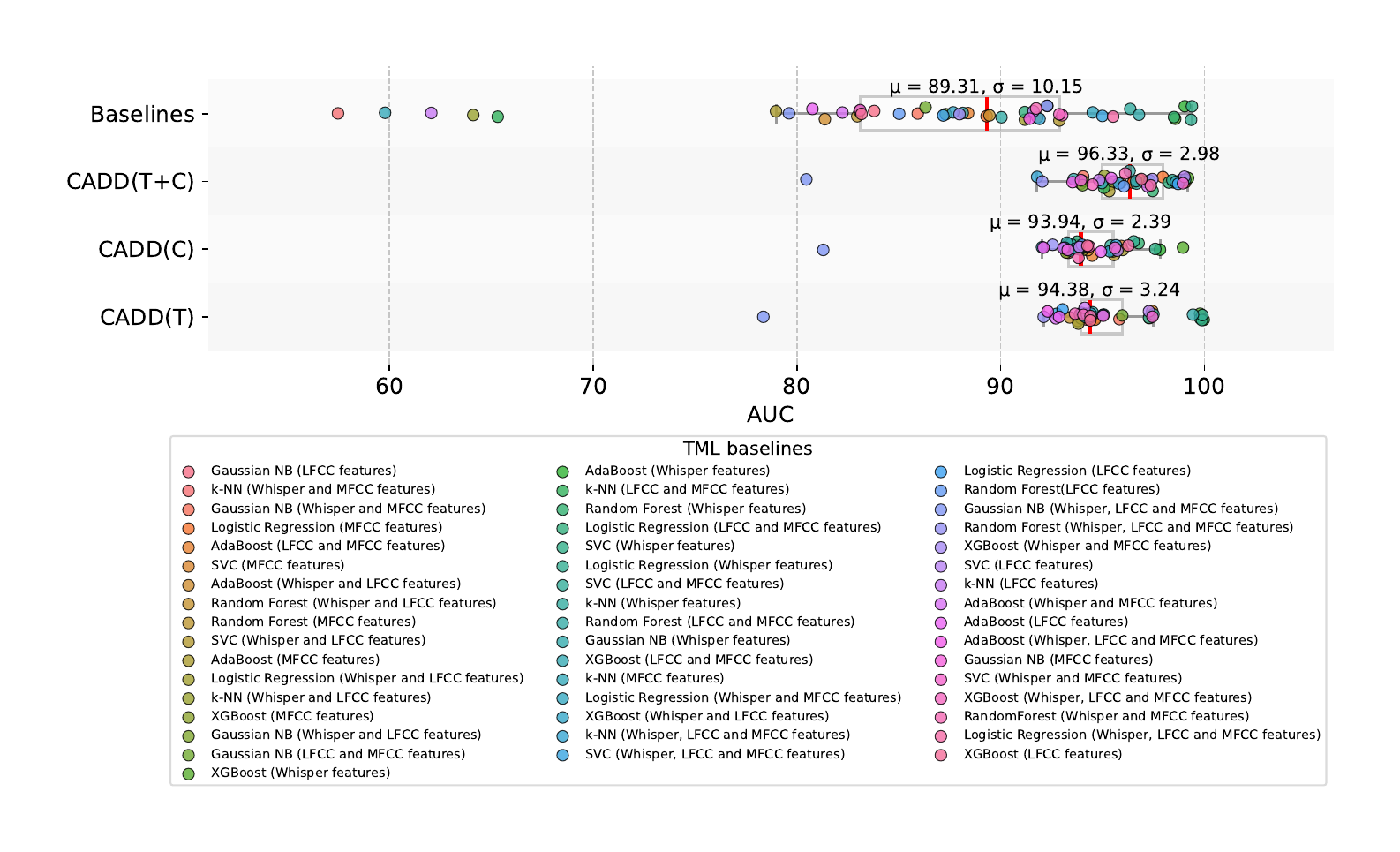}
        \label{fig:tml-baselines-AUC_SYN}
    }
    \caption{\textbf{Performance comparison of state-of-the-art (a) and traditional machine learning (b) baselines and our CADD configurations (CADD(T), CADD(C), and CADD(T+C)).} Each point represents a model's AUC score computed on our synthetic dataset (SYN), with the same color denoting the same baseline across different configurations.}
    \label{fig:combined-figures-AUC_SYN}
\end{figure}

\begin{figure}[h!]
    \centering
    \subfigure[CADD vs. State-of-the-art (SOTA) baselines]{
        \includegraphics[width=\linewidth]{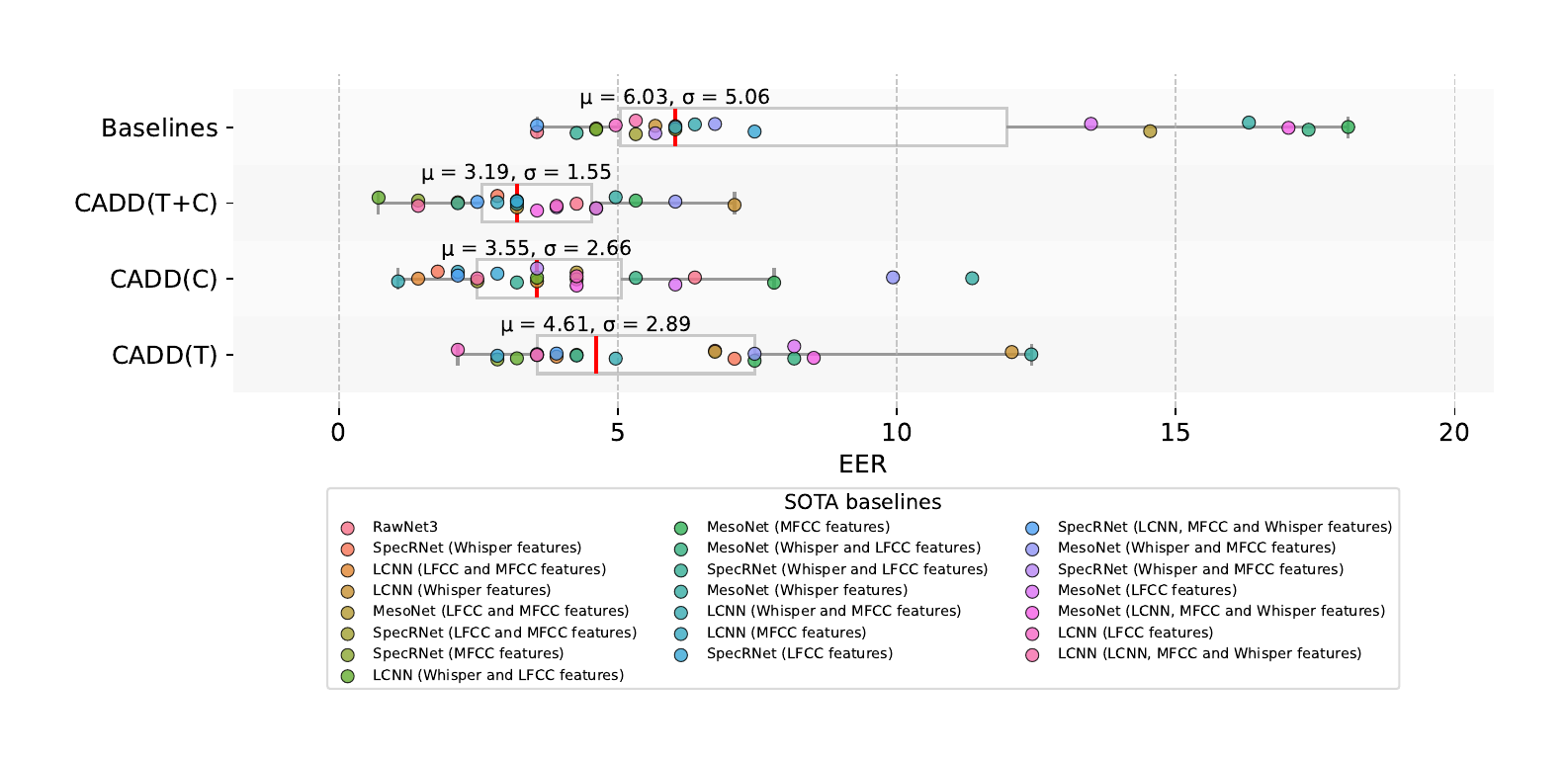}
        \label{fig:sota-baselines-EER_SYN}
    }
    \subfigure[CADD vs. Traditional Machine Learning (TML) baselines]{
        \includegraphics[width=\linewidth]{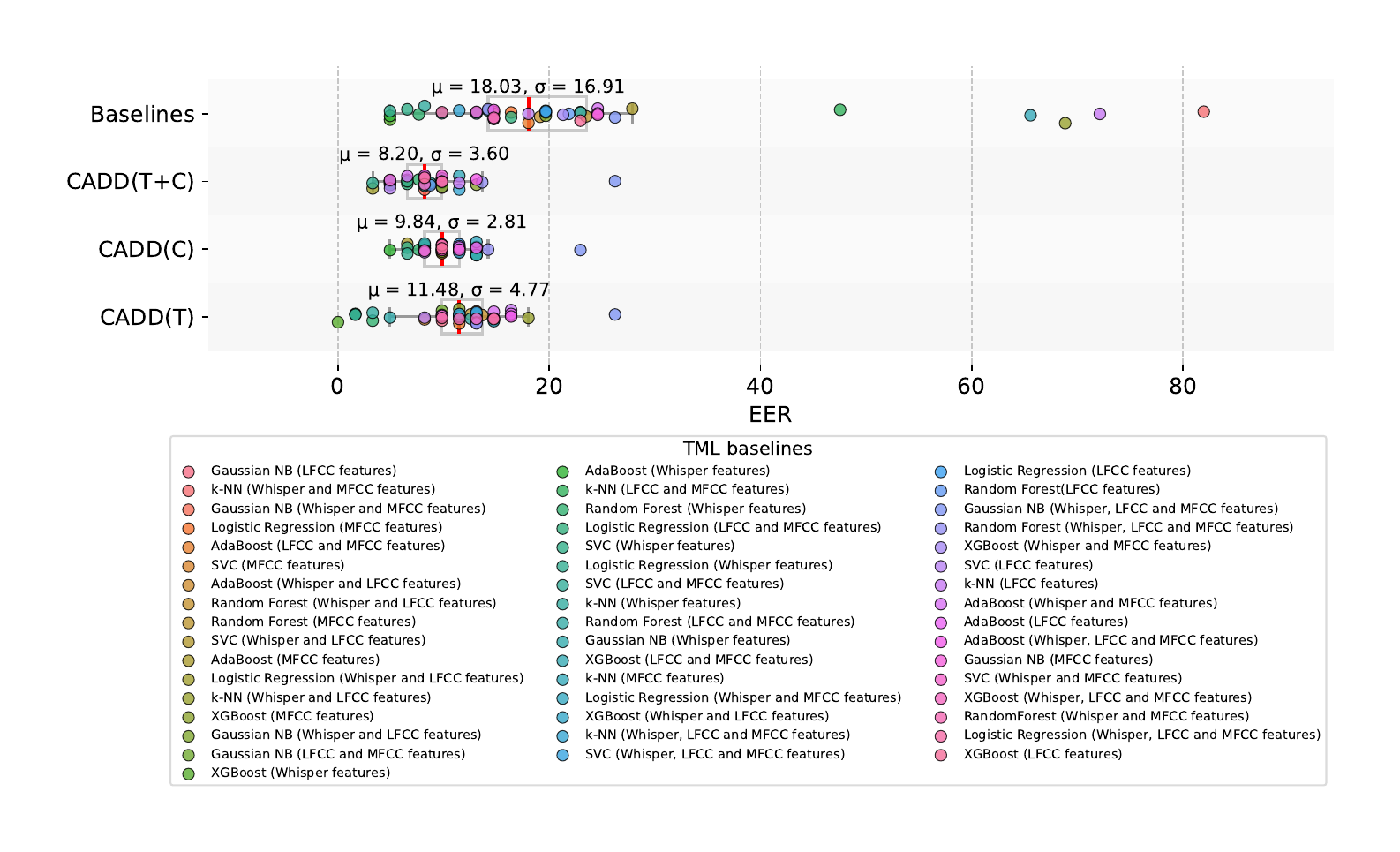}
        \label{fig:tml-baselines-EER_SYN}
    }
    \caption{\textbf{Performance comparison of state-of-the-art (a) and traditional machine learning (b) baselines and our CADD configurations (CADD(T), CADD(C), and CADD(T+C)).} Each point represents a model's EER score computed on our synthetic dataset (SYN), with the same color denoting the same baseline across different configurations.}
    \label{fig:combined-figures-EER_SYN}
\end{figure}

\begin{figure}[h!]
    \centering
    \subfigure[CADD vs. State-of-the-art (SOTA) baselines]{
        \includegraphics[width=\linewidth]{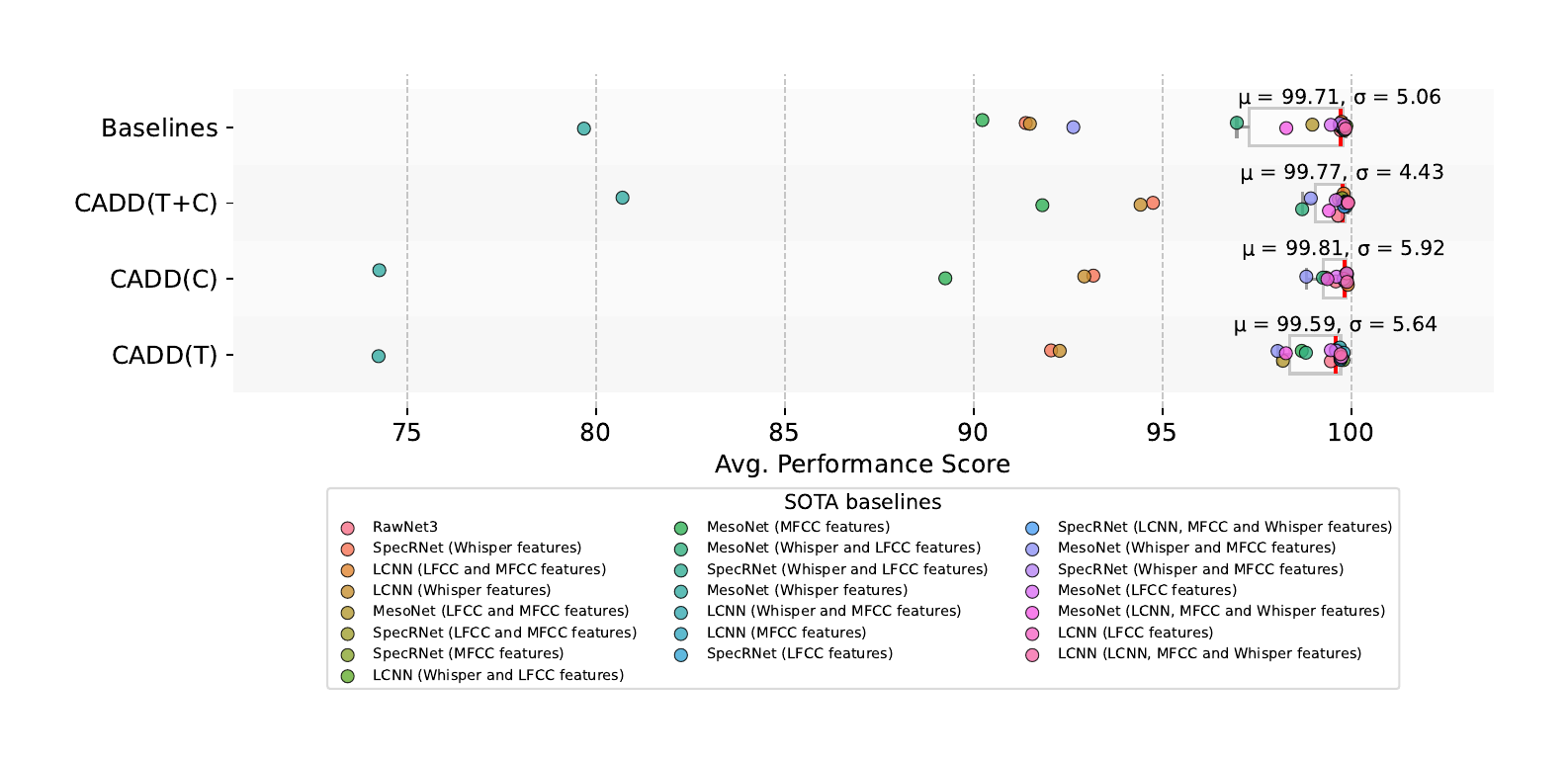}
        \label{fig:sota-baselines-Avg_ITW}
    }
    \subfigure[CADD vs. Traditional Machine Learning (TML) baselines]{
        \includegraphics[width=\linewidth]{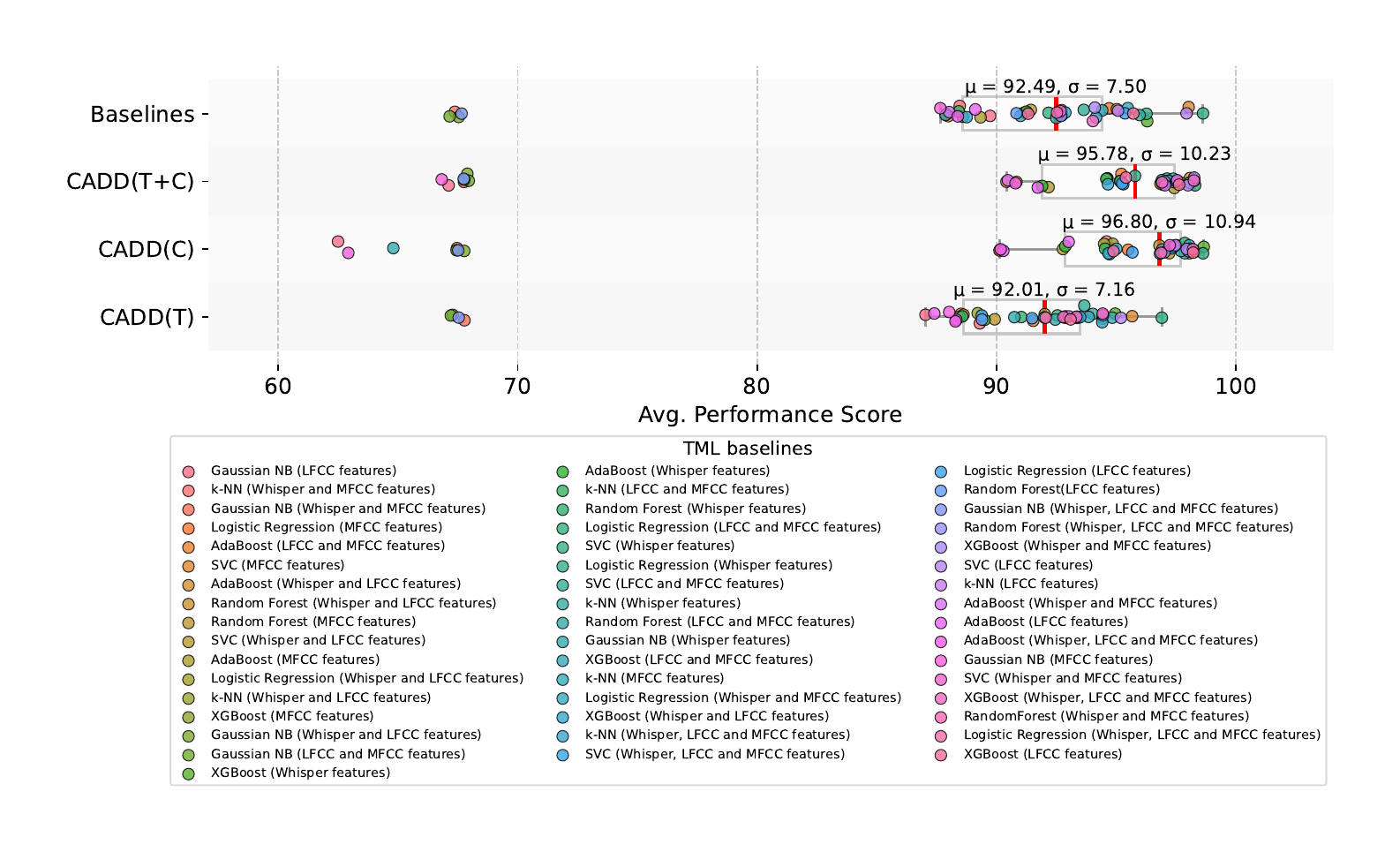}
        \label{fig:tml-baselines-Avg_ITW}
    }
    \caption{\textbf{Performance comparison of state-of-the-art (a) and traditional machine learning (b) baselines and our CADD configurations (CADD(T), CADD(C), and CADD(T+C)).} Each point represents a model's Avg score computed on In-The-Wild (ITW), with the same color denoting the same baseline across different configurations. }
    \label{fig:combined-figures-Avg_ITW}
\end{figure}

\begin{figure}[h!]
    \centering
    \subfigure[CADD vs. State-of-the-art (SOTA) baselines]{
        \includegraphics[width=\linewidth]{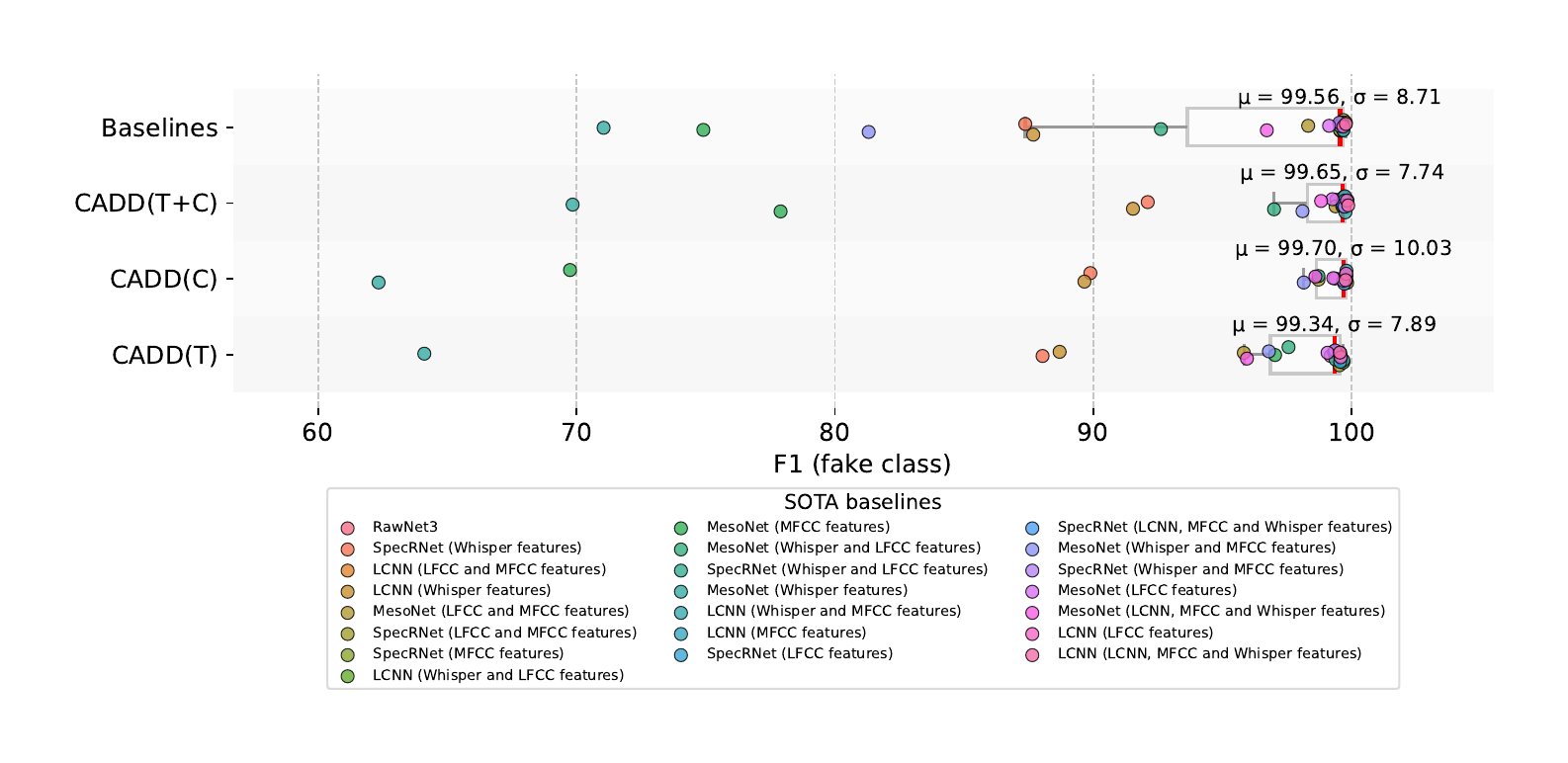}
        \label{fig:sota-baselines-F1_ITW}
    }
    \subfigure[CADD vs. Traditional Machine Learning (TML) baselines]{
        \includegraphics[width=\linewidth]{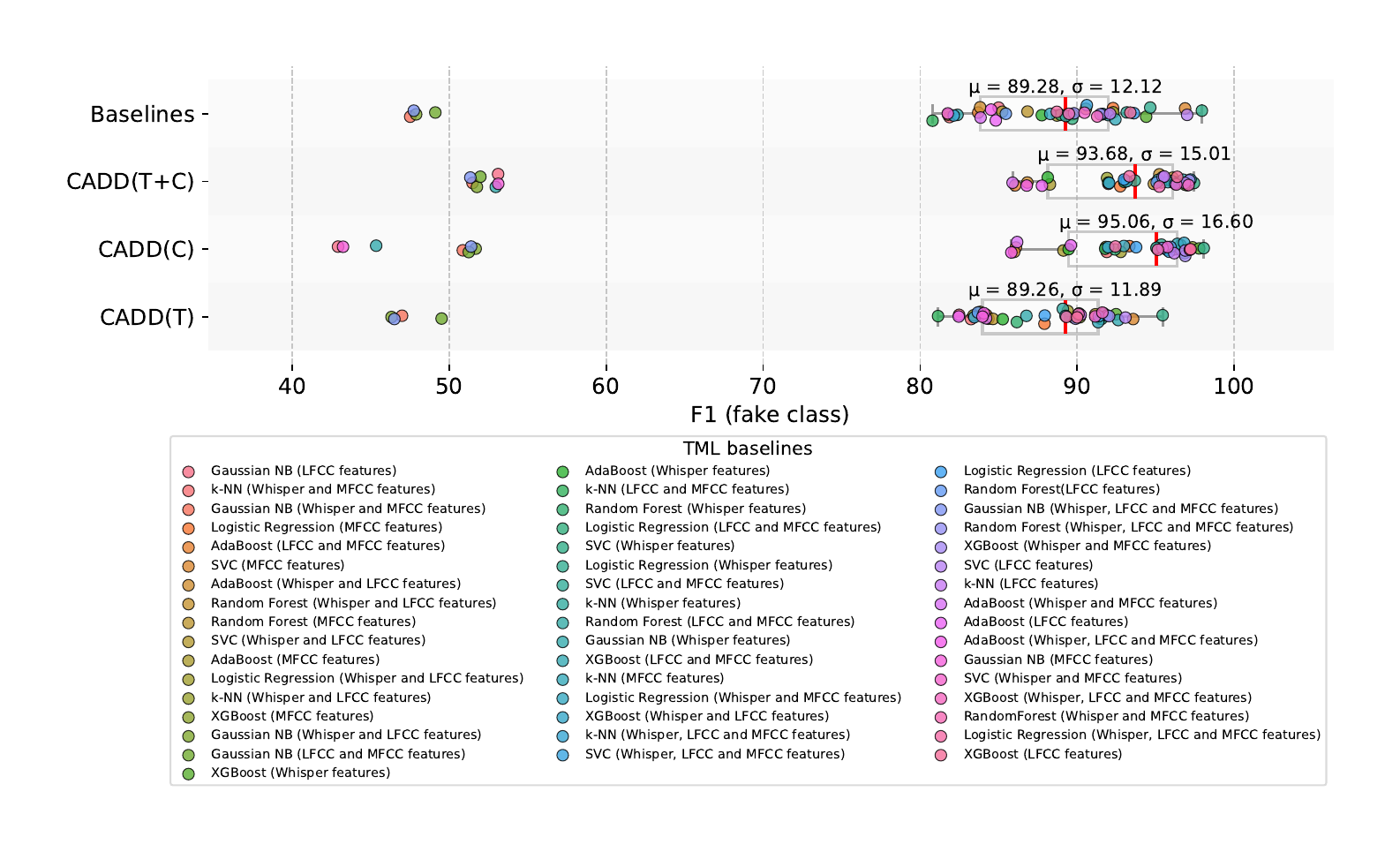}
        \label{fig:tml-baselines-F1_ITW}
    }
    \caption{\textbf{Performance comparison of state-of-the-art (a) and traditional machine learning (b) baselines and our CADD configurations (CADD(T), CADD(C), and CADD(T+C)).} Each point represents a model's F1 score computed on In-The-Wild (ITW), with the same color denoting the same baseline across different configurations.}
    \label{fig:combined-figures-F1_ITW}
\end{figure}

\begin{figure}[h!]
    \centering
    \subfigure[CADD vs. State-of-the-art (SOTA) baselines]{
        \includegraphics[width=\linewidth]{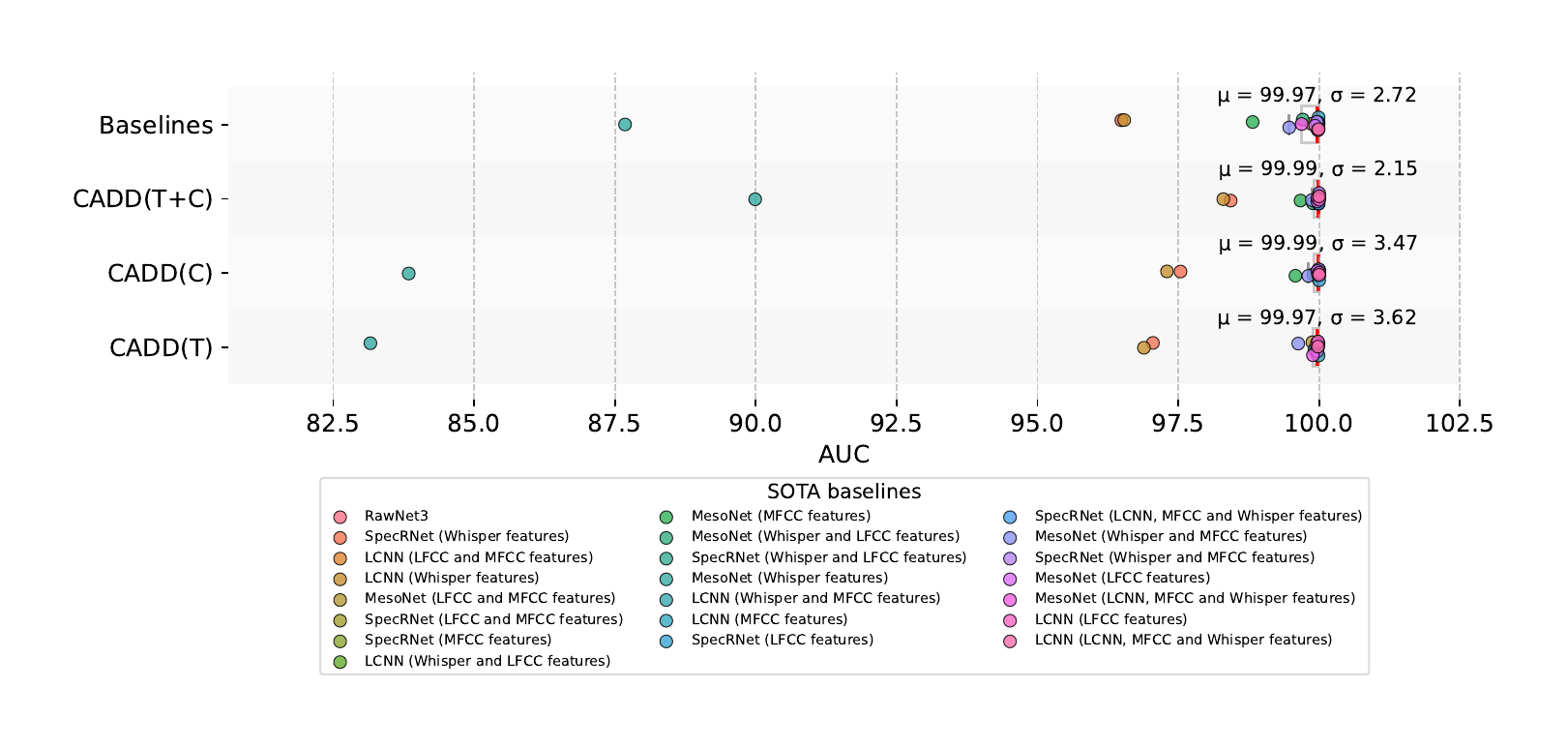}
        \label{fig:sota-baselines-AUC_ITW}
    }
    \subfigure[CADD vs. Traditional Machine Learning (TML) baselines]{
        \includegraphics[width=\linewidth]{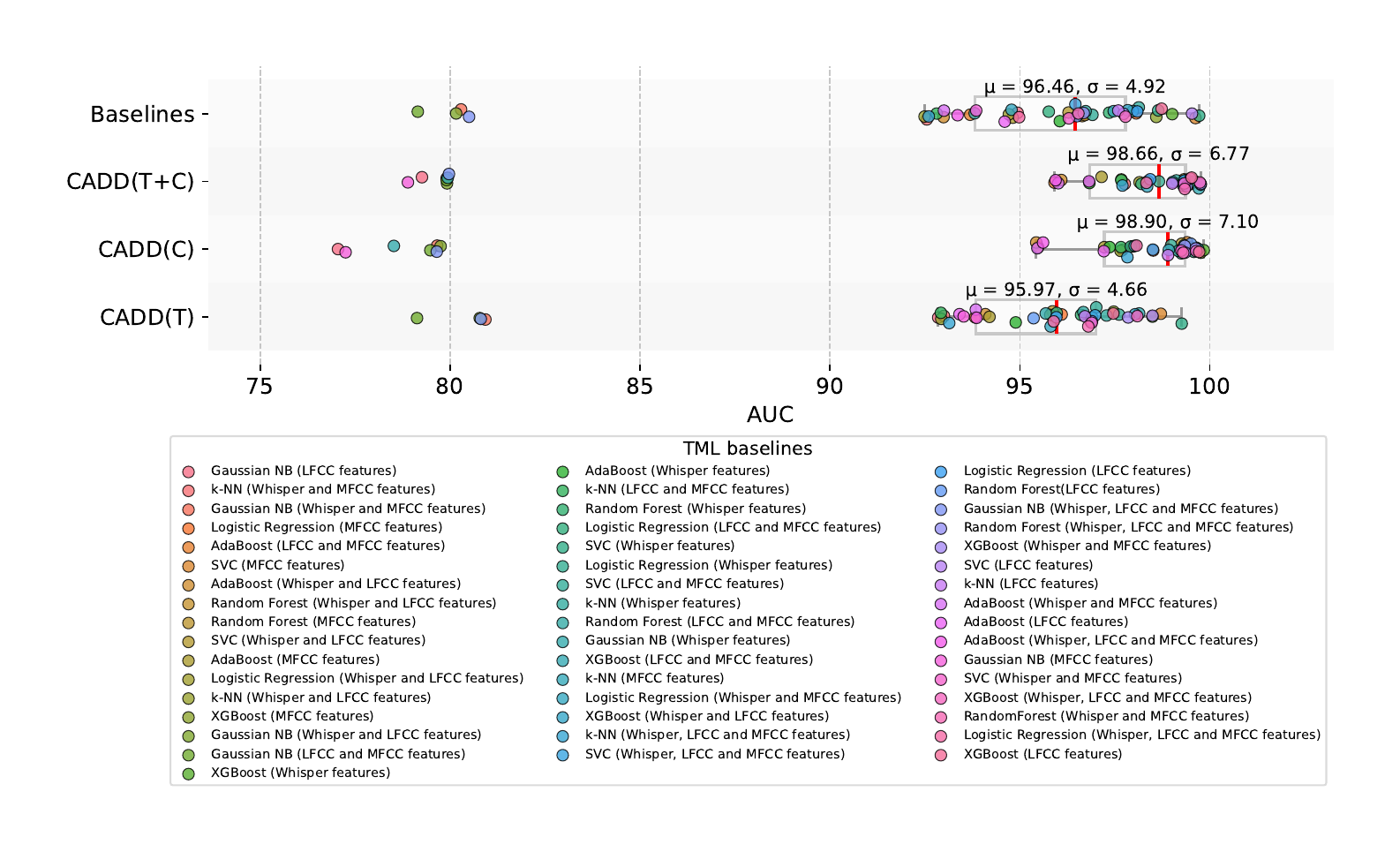}
        \label{fig:tml-baselines-AUC_ITW}
    }
    \caption{\textbf{Performance comparison of state-of-the-art (a) and traditional machine learning (b) baselines and our CADD configurations (CADD(T), CADD(C), and CADD(T+C)).} Each point represents a model's AUC score computed on In-The-Wild (ITW), with the same color denoting the same baseline across different configurations. }
    \label{fig:combined-figures-AUC_ITW}
\end{figure}

\begin{figure}[h!]
    \centering
    \subfigure[CADD vs. State-of-the-art (SOTA) baselines]{
        \includegraphics[width=\linewidth]{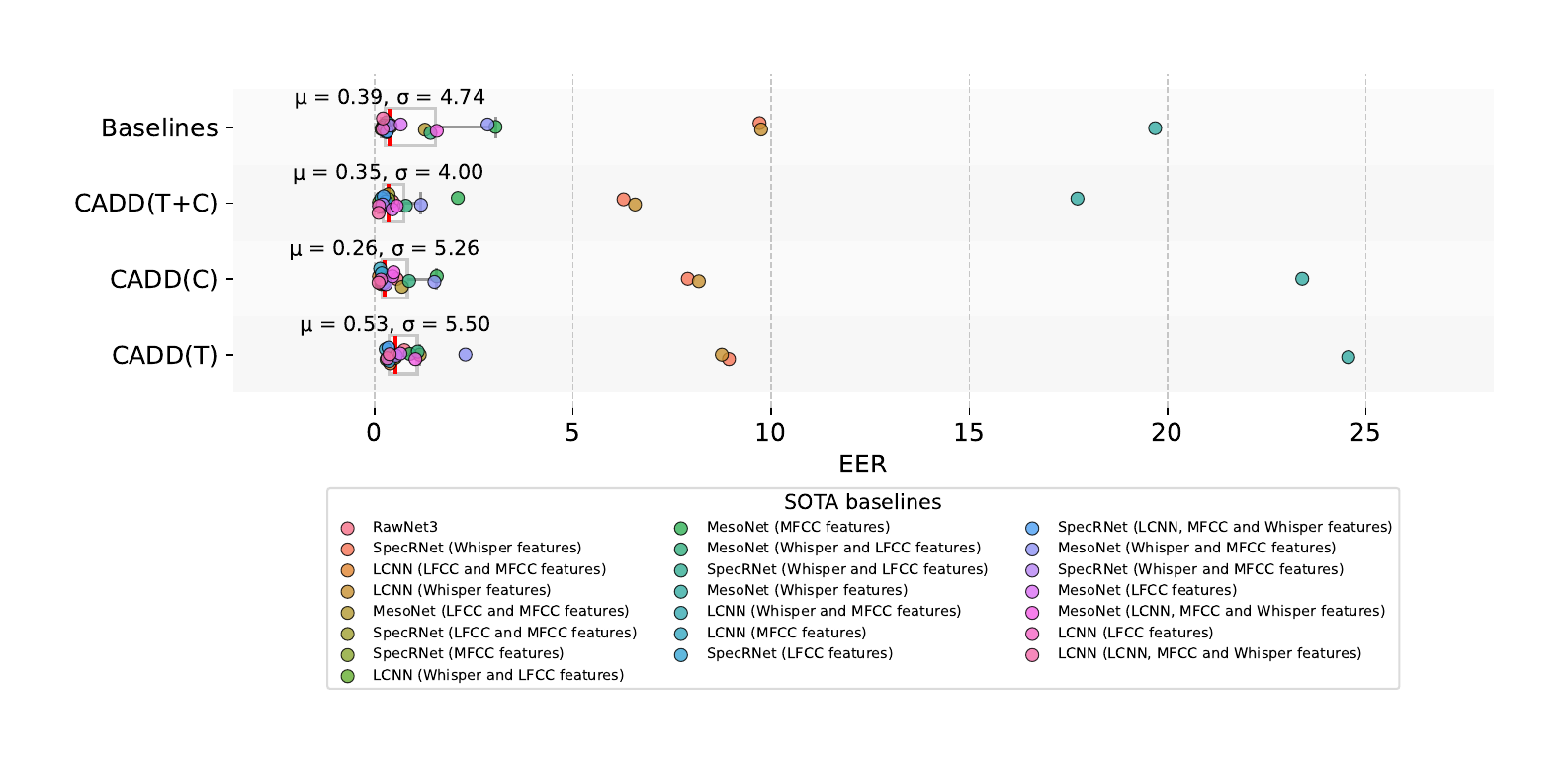}
        \label{fig:sota-baselines-EER_ITW}
    }
    \subfigure[CADD vs. Traditional Machine Learning (TML) baselines]{
        \includegraphics[width=\linewidth]{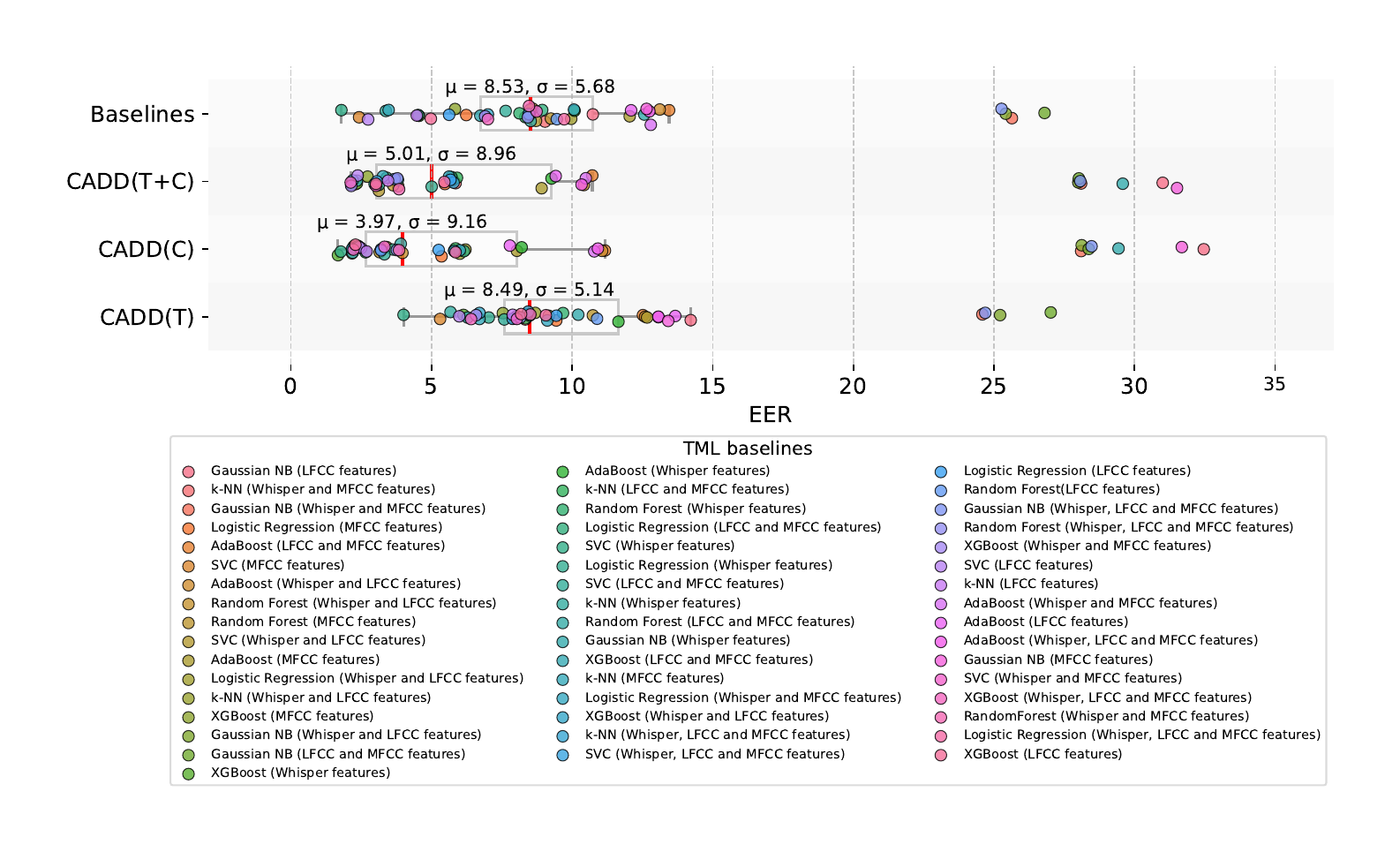}
        \label{fig:tml-baselines-EER_ITW}
    }
    \caption{\textbf{Performance comparison of state-of-the-art (a) and traditional machine learning (b) baselines and our CADD configurations (CADD(T), CADD(C), and CADD(T+C)).} Each point represents a model's EER score computed on In-The-Wild (ITW), with the same color denoting the same baseline across different configurations.}
    \label{fig:combined-figures-EER_ITW}
\end{figure}

\begin{figure}[h!]
    \centering
    \includegraphics[width=.9\linewidth]{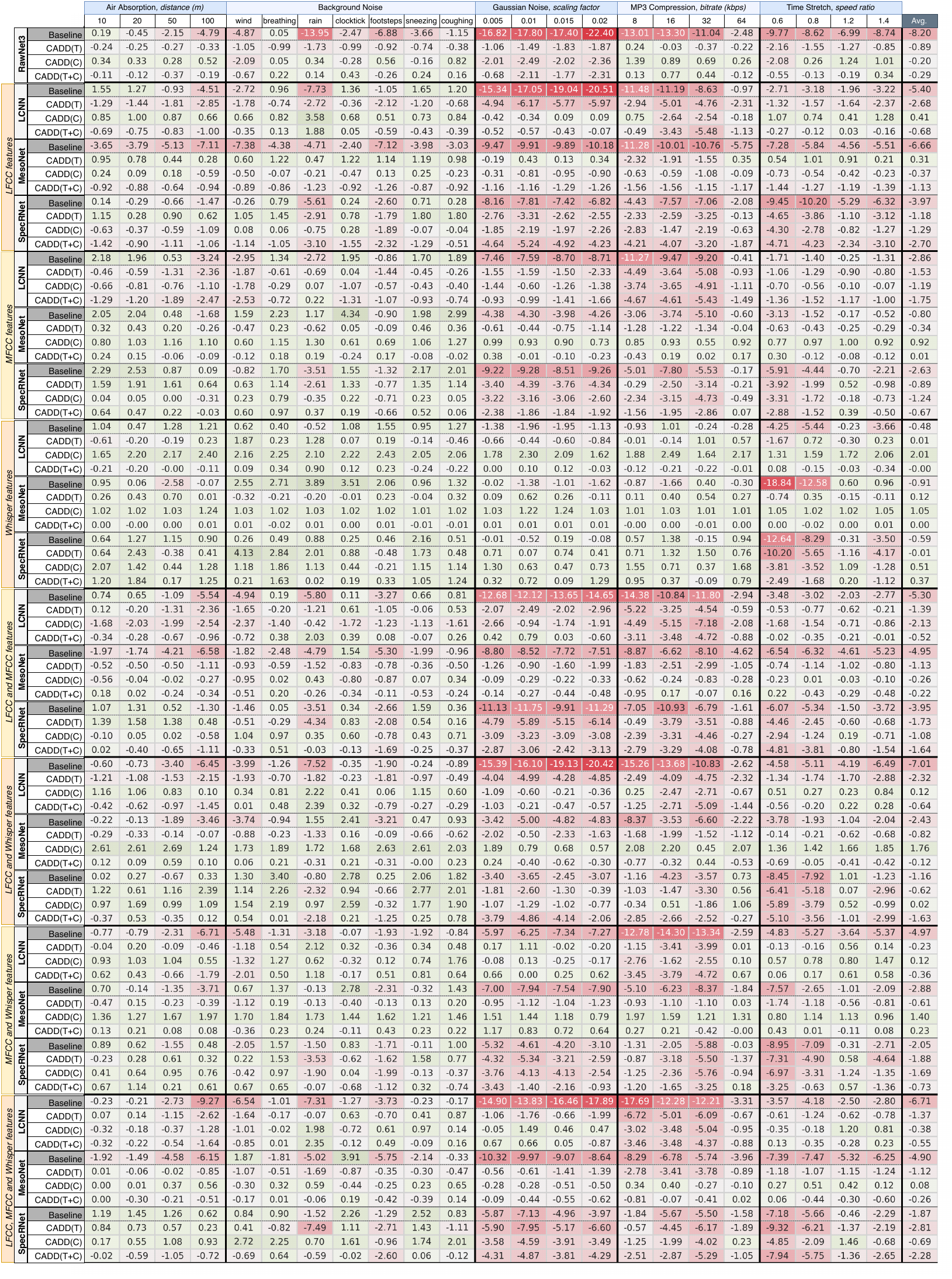}
    \caption{\textbf{Performance Robustness Under Audio Perturbations (JDD dataset).} This heatmap shows the degradation (i.e. absolute difference of performance) in Avg scores for state-of-the-art (SOTA) baselines and CADD configurations under various noise perturbations applied to the JDD test set. Red indicates performance degradation, while green indicates improvement compared to unperturbed conditions.}
    \label{fig:robustness_jdd}
\end{figure}

\begin{figure}[h!]
    \centering
    \includegraphics[width=.9\linewidth]{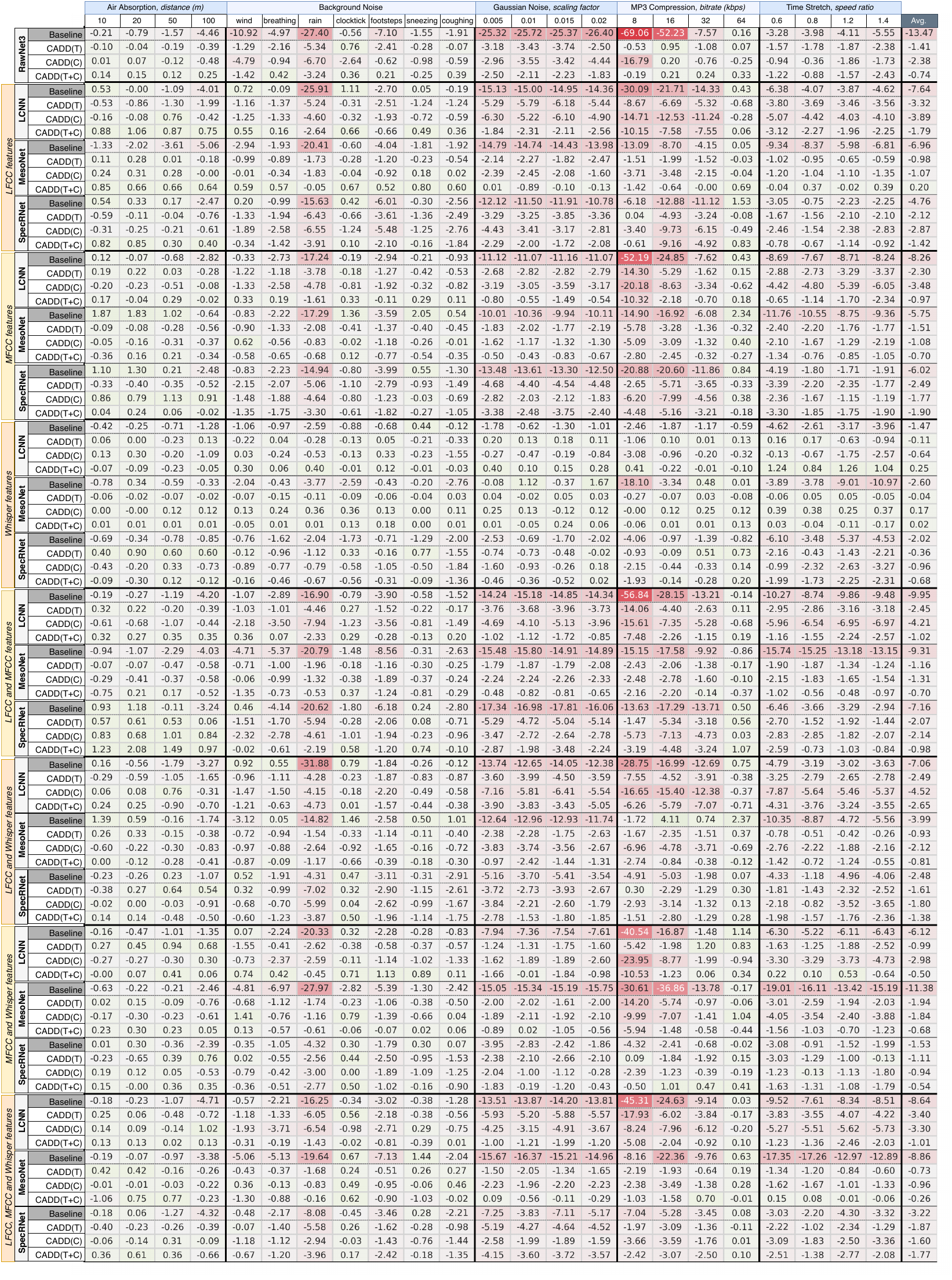}
    \caption{\textbf{Performance Robustness Under Audio Perturbations (SYN dataset).} This heatmap shows the degradation (i.e. absolute difference of performance) in Avg scores for state-of-the-art (SOTA) baselines and CADD configurations under various noise perturbations applied to the SYN test set. Red indicates performance degradation, while green indicates improvement compared to unperturbed conditions.}
    \label{fig:robustness_syn}
\end{figure}

\begin{figure}[h!]
    \centering
    \includegraphics[width=.9\linewidth]{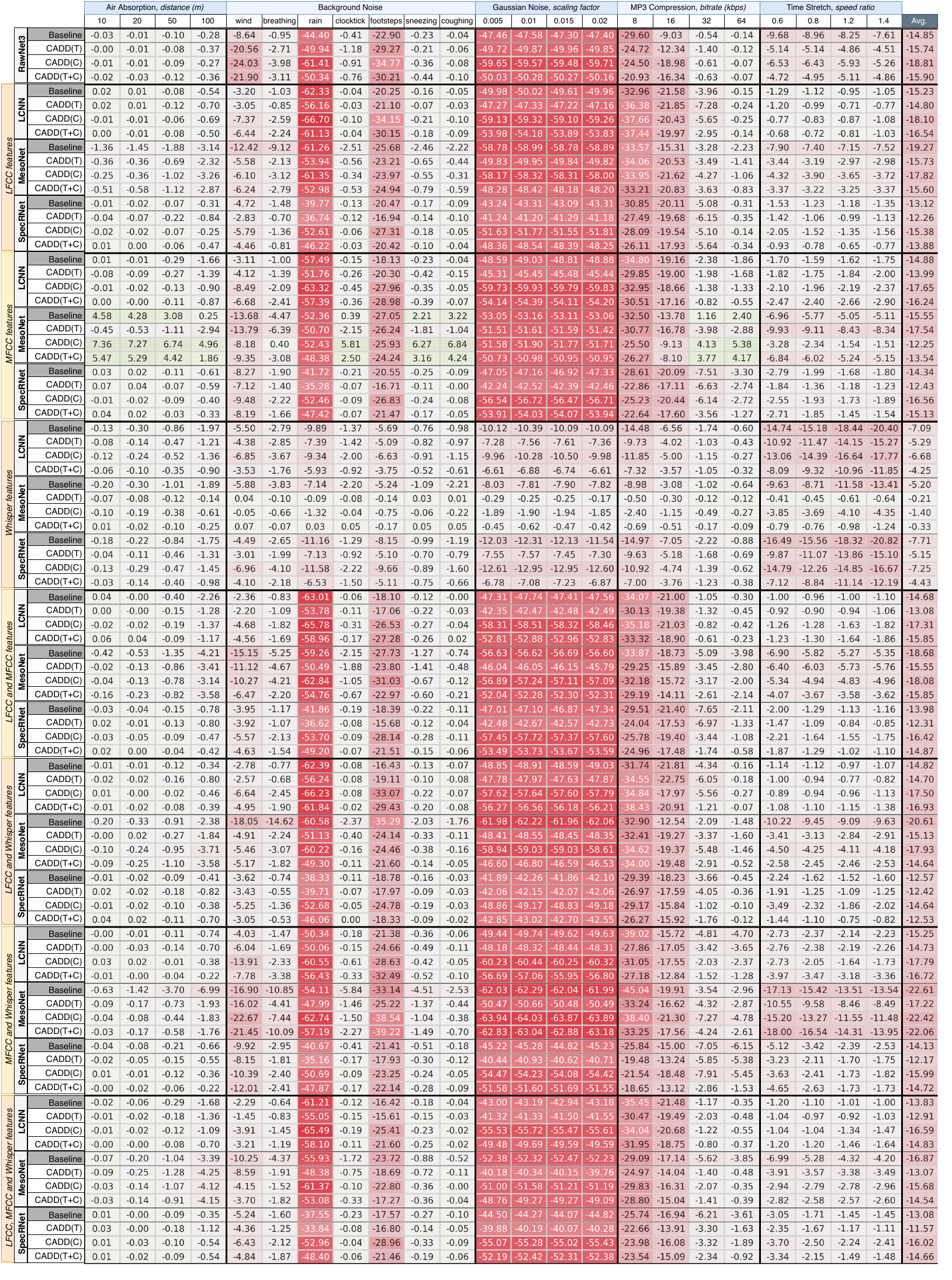}
    \caption{\textbf{Performance Robustness Under Audio Perturbations (ITW dataset).} This heatmap shows the degradation (i.e. absolute difference of performance) in Avg scores for state-of-the-art (SOTA) baselines and CADD configurations under various noise perturbations applied to the ITW test set. Red indicates performance degradation, while green indicates improvement compared to unperturbed conditions.}
    \label{fig:robustness_itw}
\end{figure}

\begin{figure}[h!]
    \centering
    \includegraphics[width=\linewidth]{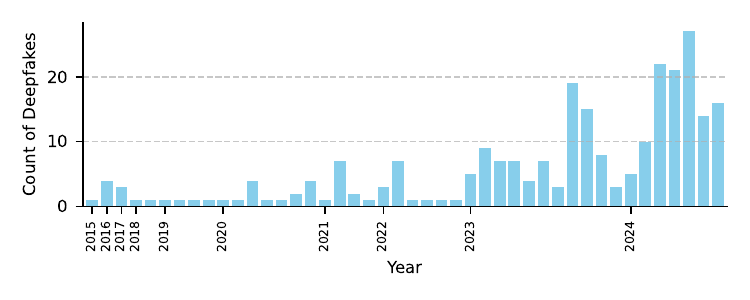}
    \caption{Temporal distribution of deepfake audio samples in the JDD dataset. Each bar represents the number of samples collected in a given month, with the leftmost bar of each year indicating the first month of collection for that year.}
    \label{fig:godds-rw-timetrend}
\end{figure}

\begin{figure}[h!]
    \centering
    \includegraphics[width=\linewidth]{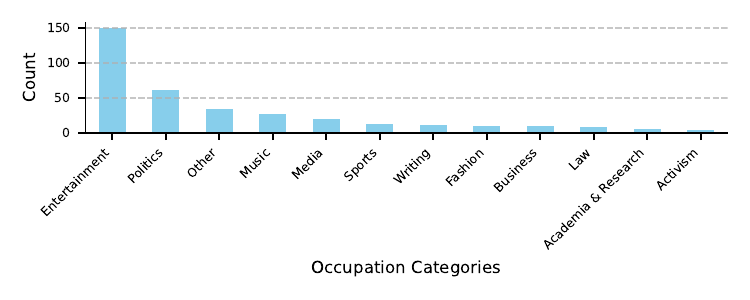}
    \caption{Distribution of occupation categories among the well-known figures in the JDD dataset}
    \label{fig:godds-rw-subjectcategories}
\end{figure}

%%%%%%%%%%%%%%%% SUPPLEMENTARY TABLES %%%%%%%%%%%%%%%

\input{nature_tables}

%%%%%%%%%%% CAPTIONS FOR OTHER SUPPLEMENTARY FILES %%%%%%%%%%

% \clearpage % Clear all remaining figures and tables then start a new page

% \paragraph{Caption for Movie S1.}
% \textbf{All captions must start with a short bold sentence, acting as a title.}
% Then explain what is shown in the supplementary video file.
% Give as much detail as you would for a figure \eg explain axes, color maps etc.
% If the video is an animated equivalent of one of the static figures, state \eg
% `Animated version of Figure~\ref{fig:example}.'

% \paragraph{Caption for Data S1.}
% \textbf{All captions must start with a short bold sentence, acting as a title.}
% Then explain what is included in the supplementary data file.
% Give as much detail as you would for a table \eg explain the meaning of every column,
% units used, any special notation etc.

%%%%%%%%%%%%%%%% SUPPLEMENTARY REFERENCES %%%%%%%%%%%%%%%

% Do NOT include a reference list in the supplement.
% All references must be in a single list at the end of the main text.
% The copyeditors will ensure that the correct reference list appears with each version of the paper
% (print, HTML, PDF, mobile app, metadata for bibliographic databases etc.)

\end{document}